%% file: main.tex
\documentclass{article} 
\usepackage{iclr2027_conference,times}

\input{math_commands.tex}

\usepackage{xcolor}
\usepackage{hyperref}
\usepackage{url}
\usepackage{graphicx}
\usepackage{multirow}
\usepackage{amsmath,amssymb,amsfonts}
\usepackage{mathtools}
\usepackage{amsthm}
\usepackage{mathrsfs}
\usepackage[title]{appendix}
\usepackage{textcomp}
\usepackage{manyfoot}
\usepackage{booktabs}
\usepackage{algorithm}
\usepackage{algorithmicx}
\usepackage{algpseudocode}
\usepackage{caption}
\usepackage{longtable}
\usepackage{rotating}
\usepackage{subcaption}
\usepackage{placeins}
\usepackage{fontawesome5}

\theoremstyle{plain}
\newtheorem{theorem}{Theorem}[section]
\newtheorem{proposition}[theorem]{Proposition}

\theoremstyle{definition}
\newtheorem{definition}[theorem]{Definition}

\theoremstyle{remark}

\title{Equivariant Deep Learning in Direct-Product Space for Electronic-Structure Hamiltonian Prediction}

\author{ 
	Shi Yin\textsuperscript{1,\faEnvelope[regular]}\hspace{0.1em}, 
	\hspace{0.5em} Zujian Dai\textsuperscript{1,\faEnvelope[regular]}, 
	\hspace{0.5em} Xinyang Pan\textsuperscript{2}, 
	\hspace{0.5em} Lixin He\textsuperscript{2,1,\faEnvelope[regular]}
	\\ 
	\textsuperscript{1}\;Institute of Artificial Intelligence, Hefei Comprehensive National Science Center \\ 
	\textsuperscript{2}\;Laboratory of Quantum Information, University of Science and Technology of China \\ 
	\textsuperscript{\faEnvelope[regular]}\;
	Corresponding authors.
}

\iclrfinalcopy
\begin{document}

\maketitle

\begin{abstract}
Deep learning provides a promising route to efficient prediction of electronic-structure Hamiltonians, enabling high-throughput screening across large collections of materials. However, most existing equivariant architectures propagate features in direct-sum spaces of irreducible spherical tensors, whereas the target Hamiltonian is naturally an operator block in a direct-product representation over local orbital and spin bases. This representation mismatch fragments contiguous Hamiltonian blocks, obscures direct manipulation of orbital couplings, and organizes features into irregular angular-momentum components that require numerous transformations and are poorly aligned with efficient hardware execution and memory management. To address both the representation mismatch and computational bottleneck, we propose DP-Learner, an equivariant learning framework that performs message passing directly on Hamiltonian-like operator blocks in the direct-product space. Theoretically, we propose Lie-generator-based directional embedding and invariantly weighted left--right products and contractions that preserve equivariance to $SO(3)$, $SU(2)$, and  spatial inversions, without explicit Clebsch--Gordan decomposition or inverse reconstruction. Methodologically, we instantiate this principle as a Transformer architecture, in which Hamiltonian-like feature matrices or tensors evolve through equivariant operators in the direct-product space, while non-linear invariant scalar routing controls structured gating over matrix units, directional propagation, attention, and environment-embedded context interactions. We further construct MP-HAM-110K, a large-scale Hamiltonian dataset comprising more than 110,000 material structures covering 68 elements with both Hamiltonian and wavefunction data. Experiments on MP-HAM-110K and Materials-HAM-SOC demonstrate that
DP-Learner dramatically improves Hamiltonian prediction accuracy
and inference efficiency over the state-of-the-art method,
with up to a $40.6\%$ reduction in prediction error, an $11.2\times$
speedup in neural inference, and a $95.2\%$ reduction in peak GPU
memory consumption.
\end{abstract}

\section{Introduction}
Accurately computing electronic-structure Hamiltonians, from which energy spectra, band topology, and electronic wavefunctions can be obtained, enables the prediction of electrical, optical, transport, and dynamical properties and ultimately connects microscopic electronic structure to macroscopic condensed-matter behavior. Such capabilities are essential for overcoming the efficiency bottleneck of traditional trial-and-error materials development and enabling low-cost, high-throughput materials screening for accelerated discovery. For decades, Hamiltonians have been primarily computed using density functional theory (DFT) \citep{hohenberg1964inhomogeneous,kohn1965self}. However, DFT requires a self-consistent (SC) iteration loop involving repeated matrix diagonalizations, leading to a prohibitive computational cost of $\mathcal{O}(T N^3)$ for a system of size $N$ and $T$ SC iterations. This cubic scaling severely limits the simulation of large-scale or structurally complex materials.

The landscape of computational physics has been substantially reshaped by the recent integration of artificial intelligence \citep{zhang2025artificial}. State-of-the-art deep geometric networks \citep{gong2023general,DBLP:conf/icml/YuXQQJ23,zhang2024self,zhong2024universal,wang2024universal,tracegrad,DBLP:conf/iclr/LiXHWHYLWZLSG25} have demonstrated the capability to directly map spatial atomic configurations to the corresponding electronic-structure Hamiltonian matrices. By bypassing the expensive self-consistent iterations, these data-driven approaches achieve orders-of-magnitude computational acceleration in acquiring core electronic-structure quantities, enabling high-fidelity exploration of novel materials. For more details, readers are directed to Appendix~\ref{eed}.

\begin{figure}
	\centering    
	\includegraphics[scale=0.35]{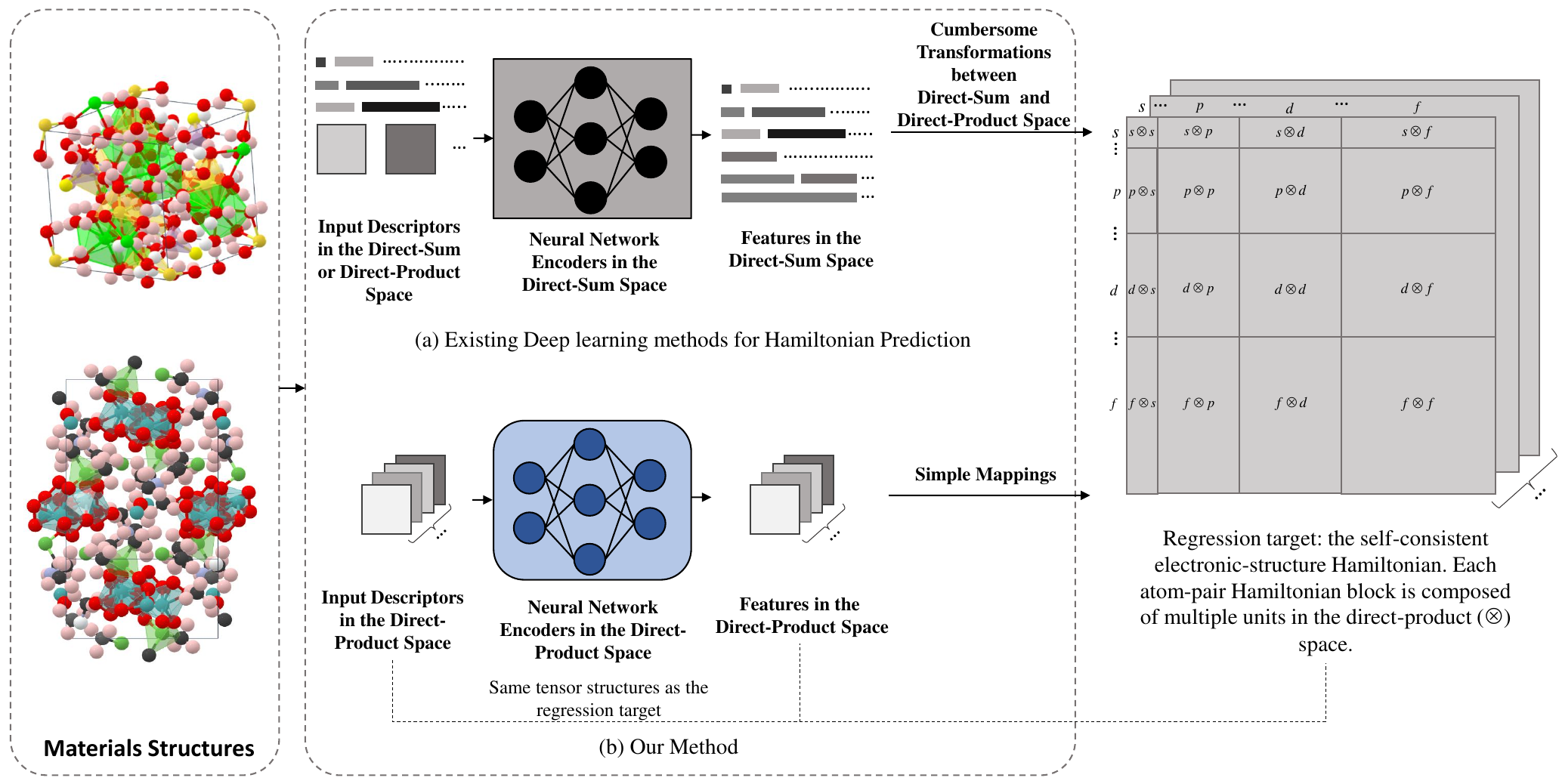}
	\caption{Existing approaches often encode features in the direct-sum representations of irreducible spherical tensors, creating a mismatch with the Hamiltonian target and requiring irregular, transformation-heavy computations that are difficult to execute efficiently. In contrast, DP-Learner operates directly in the Hamiltonian's native direct-product representation, preserving the target tensor structure, orbital and spin couplings, as well as the symmetry constraints throughout the network for accurate and efficient equivariant Hamiltonian prediction.}
	\label{main_framework}
\end{figure}

However, current deep learning methods for electronic-structure Hamiltonian prediction still face central limitations in both representation quality and computational efficiency. Physically, the Hamiltonian is naturally represented as operator blocks in the direct-product (or tensor-product) space of local orbital and spin bases, where orbital and spin degrees of freedom are coupled with the underlying spatial geometry within a unified matrix- or tensor-valued observable.  In contrast, existing equivariant architectures for Hamiltonian prediction commonly re-express these observables as direct sums of irreducible spherical tensors. This indirect organization fragments contiguous Hamiltonian blocks and obscures explicit orbital couplings. In $SO(3)$ tensor-product layers, feature products repeatedly undergo Clebsch--Gordan decomposition into irreducible components. Moreover, the heterogeneous channel sizes and angular-momentum-dependent tensor product, decomposition and reconstruction paths commonly involve padding and masking, sparse Clebsch--Gordan coefficient accesses, frequent indexing and coefficient lookups, and many small tensor kernels. These irregular operations complicate memory management, reduce data reuse, and are poorly matched to dense computation on modern hardware accelerators, particularly for rich basis sets containing $d$- and $f$-orbitals or explicit spin degrees of freedom.

A recent advance in Hamiltonian prediction is NextHAM \citep{yin2026advancing}, which introduces the zeroth-step Hamiltonian $H^{(0)}$ as a physically informed input descriptor that is naturally represented in the same direct-product operator space as the target Hamiltonian and can be efficiently constructed without matrix diagonalization. Because $H^{(0)}$ already encodes substantial electronic-structure priors while sharing the matrix topology, orbital basis, and symmetry constraints of the final self-consistent Hamiltonian, it alleviates the information sparsity of conventional atomic or bond embeddings and provides a unified, physically informed representation across diverse materials. This advance substantially improves cross-material transferability of the neural network, moving Hamiltonian learning beyond individual material systems with limited variation toward generalizable electronic-structure prediction over a chemically diverse materials space of more than $60$ elements. However, NextHAM still converts $H^{(0)}$ into irreducible spherical tensors for feature propagation. The mismatch between fragmented features in the direct-sum space and native Hamiltonian blocks, together with the overhead of sparse and heterogeneous tensor operations and the heavy transformation between direct-sum and direct-product states, still remains. A detailed discussion of these related works is provided in Appendix~\ref{related_work}.

To address the representation mismatch and computational burdens, we propose $DP$-$Learner$, an equivariant  Hamiltonian-prediction framework that performs message passing directly on complete atom-pair Hamiltonian blocks in the $D$irect-$P$roduct space. Starting from the zeroth-step Hamiltonian $H^{(0)}$, DP-Learner preserves the matrix or tensor layout of the target self-consistent Hamiltonian $H^{(T)}$ throughout the network and learns the  correction from $H^{(0)}$ to $H^{(T)}$. This design  naturally captures the intrinsic coupling structure of Hamiltonians in the direct-product space, while avoiding Clebsch--Gordan decompositions of feature tensor products and conversions between Hamiltonian matrices and irreducible features, providing substantial advantages in computational efficiency and hardware utilization, as analyzed in Appendix~\ref{app:complexity_efficiency}.

Theoretically, we establish a Hamiltonian-native equivariant deep learning framework based on two complementary ingredients. First, we develop a Lie-generator-based directional embedding mechanism that maps each local geometric direction $\hat{\mathbf r}_{ab}$ to an equivariant operator acting on the local orbital or orbital--spin spaces associated with the Hamiltonian block in the direct-product space. Second, we construct invariantly weighted left--right product and contraction mechanisms to update Hamiltonian-like matrices and spinor tensors while preserving their transformation laws under spatial rotations and inversions. These formulations encode geometric information and electronic couplings in a unified direct-product space instead of decomposing and reconstructing them through irreducible channels.

Methodologically, we develop our learning framework as an expressive and equivariant Transformer architecture in the direct-product space. Each  layer encodes both equivariant matrix or tensor features and invariant scalar features. Normalized Frobenius summaries are computed for the direct-product units within each Hamiltonian block, where each unit couples one row orbital set to one column orbital set. These summaries, together with radial features and scalar node and edge features, are processed by non-linear scalar networks to produce per-unit gates, attention weights, directional weights, and contraction coefficients. Multi-head attention updates node features and conditions the edge-feature update, while constructing equivariant atom-wise context representations. The directional operator and the context then update each edge matrix or tensor through left, right, and bilateral products and contractions. Dimension-normalized products, LayerNorm, LayerScale, and zero-initialized update heads stabilize deep residual learning around $H^{(0)}$.

Broad-coverage benchmarks for deep learning Hamiltonian prediction remain limited. To address this gap, we construct
and release MP-HAM-110K, a benchmark comprising more than
$110{,}000$ diverse material structures sampled from the
Materials Project~\citep{jain2013commentary} and covering
$68$ elements. It spans the first six rows of the periodic
table and employs a rich atomic basis up to $4s2p2d1f$.
MP-HAM-110K is approximately six times larger than the existing
Materials-HAM-SOC dataset and includes structures containing
more than $150$ atoms, whereas Materials-HAM-SOC contains at
most $80$ atoms per structure. It thus provides a valuable
benchmark for Hamiltonian prediction across chemically diverse
materials and larger structures.

Comprehensive experiments on the proposed MP-HAM-110K benchmark and the
existing Materials-HAM-SOC dataset demonstrate that DP-Learner achieves
state-of-the-art Hamiltonian prediction accuracy together with
dramatically improved computational efficiency across chemically
diverse materials spanning more than 60 elements, substantially
outperforming NextHAM, the pioneering universal Hamiltonian baseline published very recently. The predicted
Hamiltonians also yield accurate band structures. These results
demonstrate the value of representation learning in the
direct-product space for accurate and efficient electronic-structure
prediction, materials screening, and discovery.

\section{Preliminary}
\label{sec:preliminary}

Let $H_{ab}$ denote the atom-pair Hamiltonian block whose row basis belongs to atom $a$ and whose column basis belongs to atom $b$. Throughout the paper, \emph{Hamiltonian block} refers to this complete matrix or tensor for one atomic pair, while \emph{direct-product unit} refers to the part of the block coupling one row orbital set to one column orbital set, with the spin degree of freedom included in the spin-orbit-coupled (SOC) setting. The notation covers both on-site blocks $(a=b)$ and off-site blocks $(a\neq b)$. For a non-SOC system represented in a real orbital basis, a spatial rotation $R\in SO(3)$ acts as
\begin{equation}
	\mathcal{T}_R(H_{ab})
	=
	D_a(R)H_{ab}D_b(R)^{\top},
	\label{eq:so3_full_block_equivariance}
\end{equation}
where $D_a(R)$ and $D_b(R)$ are the $SO(3)$ representations aligned with the local basis spaces of $H_{ab}$. They are constructed by block-diagonal concatenation of the corresponding irreducible Wigner-$D$ matrices for the local orbital sets.

For an SOC system, each basis function carries an orbital index and a spin index. The corresponding complex Hamiltonian block transforms as
\begin{equation}
	\mathcal{T}_R
	\left(
	(H_{ab})_{i\alpha,j\beta}
	\right)
	=
	\sum_{k\gamma,l\delta}
	(\widetilde{D}_a)_{i\alpha,k\gamma}(R)
	(H_{ab})_{k\gamma,l\delta}
	(\widetilde{D}_b^{\dagger})_{l\delta,j\beta}(R),
	\label{eq:su2_full_block_equivariance}
\end{equation}
where $\widetilde{D}_{a}(R)$ and $\widetilde{D}_{b}(R)$ are the
orbital--spin representations associated with the local basis spaces.
Here $R\in SO(3)$ denotes a proper active spatial rotation, and the spin
degree of freedom transforms through a corresponding lift of $R$ to
$SU(2)$. In
addition to rotational symmetry, Hamiltonians also exhibit symmetry
under spatial inversion. For a comprehensive symmetry analysis of
Hamiltonians, we refer the reader to
Appendix~\ref{app:symmetry_background}.

\section{Theory: Equivariance in Direct-Product Space}
\label{sec:theory}

Conventional equivariant networks for Hamiltonian prediction usually learn features in direct sums of irreducible tensors, whereas a Hamiltonian block is naturally represented as a matrix or tensor in the direct-product space of local orbital bases, with spin degrees of freedom additionally included when SOC is considered. Formal definitions of the direct-product and direct-sum spaces and their associated representations are provided in Definitions~\ref{def:direct_product_space_representation} and~\ref{def:direct_sum_space_representation}, respectively; the direct-product units of Hamiltonian blocks are defined in Definition~\ref{def:direct_product_unit}. Here, we aim to establish a rigorous theoretical foundation for symmetry-preserving operations in the Hamiltonian direct-product space, with exact $SO(3)$ equivariance, the corresponding $SU(2)$ spinor equivariance, and spatial-inversion equivariance, providing the theoretical basis for Section~\ref{sec:method}. 

\subsection{Equivariant Directional Embedding}
The direction vector $\hat{\mathbf r}_{ab}$ specifies an orientation in three-dimensional space, but cannot directly act on the row or column basis of a Hamiltonian block. To bridge this gap, we develop a Lie-generator-based \emph{directional embedding} formulation that maps the local geometric direction into equivariant operators acting directly on Hamiltonian-like matrices or tensors in the direct-product space. Specifically, we contract the Cartesian components of $\hat{\mathbf r}_{ab}$ with orbital- and spin-resolved Lie generators to obtain the atom-wise directional operators $V_{\xi}^{\mathrm{SO(3)}}$ and $V_{\xi}^{\mathrm{SU(2)}}$ for non-SOC and SOC systems, respectively, as detailed in Appendix~\ref{app:generator}. Under proper spatial rotations and spatial inversion, these directional operators obey the corresponding equivariance laws in the orbital or orbital--spin representations, as formalized in the following theorem:
\begin{theorem}[Equivariance of Lie-algebraic directional embeddings]
	\label{theorem:covariance}
	The directional embeddings $V_{\xi}^{\mathrm{SO(3)}}(\hat{\mathbf r})$ and $V_{\xi}^{\mathrm{SU(2)}}(\hat{\mathbf r})$ defined in Appendix~\ref{app:generator} transform equivariantly under  spatial rotations and are spatial-inversion equivariant with operator-level intrinsic parity $\eta_V=-1$.
	For non-SOC systems, $V_{\xi}^{\mathrm{SO(3)}}$ is a real orbital matrix and satisfies
	\begin{equation}
		D_{\xi}(R)
		V_{\xi}^{\mathrm{SO(3)}}(\hat{\mathbf r})
		D_{\xi}(R)^{\top}
		=
		V_{\xi}^{\mathrm{SO(3)}}(R\hat{\mathbf r}).
		\label{eq:directional_operator_so3}
	\end{equation}
	For SOC systems, $V_{\xi}^{\mathrm{SU(2)}}$ is a complex spinor tensor with components
	$(V_{\xi}^{\mathrm{SU(2)}})_{i\alpha,p\mu}$ and satisfies
	\begin{equation}
		\sum_{r\gamma,s\delta}
		(\widetilde{D}_{\xi})_{i\alpha,r\gamma}(R)
		(V_{\xi}^{\mathrm{SU(2)}}(\hat{\mathbf r}))_{r\gamma,s\delta}
		(\widetilde{D}_{\xi}^{\dagger})_{s\delta,p\mu}(R)
		=
		(V_{\xi}^{\mathrm{SU(2)}}(R\hat{\mathbf r}))_{i\alpha,p\mu}.
		\label{eq:directional_operator_su2}
	\end{equation}
	Under spatial inversion $\mathcal I:\hat{\mathbf r}\mapsto-\hat{\mathbf r}$, with $\mathsf P_{\xi}$ denoting the local parity operator defined in Eq.~(\ref{eq:inversion_parity_operator}), the two embeddings further satisfy
	\begin{equation}
			V_{\xi}^{\mathrm{SO(3)}}(-\hat{\mathbf r})
			=-\mathsf P_{\xi}V_{\xi}^{\mathrm{SO(3)}}(\hat{\mathbf r})\mathsf P_{\xi}^{\top}, \,\,\, V_{\xi}^{\mathrm{SU(2)}}(-\hat{\mathbf r})
		=-\mathsf P_{\xi}V_{\xi}^{\mathrm{SU(2)}}(\hat{\mathbf r})\mathsf P_{\xi}^{\dagger}.
		\label{eq:directional_operator_inversion_odd}
	\end{equation}
\end{theorem}

\subsection{SO(3)-Equivariant Operators}
\label{subsec:so3_theory}

For non-SOC systems, consider three feature matrices in the direct-product space:
\begin{equation}
	F_{ab}\in\mathbb{R}^{d_a\times d_b},
	\qquad
	A_a^{(m)}\in\mathbb{R}^{d_a\times d_a},
	\qquad
	B_b^{(m)}\in\mathbb{R}^{d_b\times d_b}.
\end{equation}
Here $F_{ab}$ is the feature associated with atom pair $(a,b)$, while $A_a^{(m)}$ and $B_b^{(m)}$ are atom-wise features acting on the orbital bases of atoms $a$ and $b$, respectively. The dimensions $d_a$ and $d_b$ denote the numbers of orbital basis functions on the two atoms. The superscript $m=1,\ldots,M$ indexes different feature heads, and $c_m$ denotes an invariant scalar coefficient.

\begin{theorem}[SO(3) equivariance of operators in the direct-product space]
	\label{thm:so3_equivariance}
	Assume that $F_{ab}$ satisfies the same SO(3) transformation rule as the Hamiltonian block in Eq.~(\ref{eq:so3_full_block_equivariance}). Assume also that $A_a^{(m)}$ and $B_b^{(m)}$ satisfy the same transformation rule as Eq.~(\ref{eq:so3_full_block_equivariance}) with  $(a,b)$ replaced by $(a,a)$, and $(a,b)$ replaced by  $(b,b)$, respectively. Define
	\begin{equation}
		\left[\mathcal{M}_{SO(3)}(F_{ab})\right]_{ij}
		=
		\sum_{m=1}^{M} c_m
		\sum_{p=1}^{d_a}
		\sum_{q=1}^{d_b}
		\left(A_a^{(m)}\right)_{ip}
		(F_{ab})_{pq}
		\left(B_b^{(m)}\right)_{qj},
		\label{eq:so3_primitive_contraction}
	\end{equation}
	where $i,p$ index the orbital basis of atom $a$ and $j,q$ index the orbital basis of atom $b$. Then $\mathcal{M}_{SO(3)}(F_{ab})$ satisfies the same SO(3) transformation rule as Eq.~(\ref{eq:so3_full_block_equivariance}).
\end{theorem}

In Eq.~(\ref{eq:so3_primitive_contraction}),
$A_a^{(m)}$ can be instantiated by either the directional matrix
$V_a^{\mathrm{SO(3)}}$ or the equivariant context matrix
$N_a^{\mathrm{SO(3)}}$ defined in
Appendix~\ref{subsec:app_operator_attention}. $B_b^{(m)}$ can be instantiated by either
$(V_b^{\mathrm{SO(3)}})^{\top}$ or
$(N_b^{\mathrm{SO(3)}})^{\top}$.

\subsection{SU(2)-Equivariant Operators}
\label{subsec:su2_theory}

For SOC systems, consider three feature tensors in the direct-product space:
\begin{equation}
	F_{ab}\in\mathbb{C}^{L_a\times 2\times L_b\times 2},
	\qquad
	A_a^{(m)}\in\mathbb{C}^{L_a\times 2\times L_a\times 2},
	\qquad
	B_b^{(m)}\in\mathbb{C}^{L_b\times 2\times L_b\times 2}.
\end{equation}
The components of $F_{ab}$ are written as $F_{i\alpha,j\beta}$, where $i,j$ index orbital basis functions on atoms $a,b$, and $\alpha,\beta\in\{1,2\}$ index spin. The tensors $A_a^{(m)}$ and $B_b^{(m)}$ act on the orbital-spin bases of atoms $a$ and $b$, respectively.

\begin{theorem}[SU(2) equivariance of operators in the direct-product space]
	\label{thm:su2_equivariance}
	Assume that $F_{ab}$ satisfies the same SU(2) spinor transformation rule as the Hamiltonian block in Eq.~(\ref{eq:su2_full_block_equivariance}), with $H_{ab}$ replaced by $F_{ab}$. Assume also that $A_a^{(m)}$ and $B_b^{(m)}$ satisfy the same transformation rule as Eq.~(\ref{eq:su2_full_block_equivariance}) with  $(a,b)$ replaced by $(a,a)$, and $(a,b)$ replaced by  $(b,b)$, respectively. Define
	\begin{equation}
		\left[\mathcal{M}_{SU(2)}(F_{ab})\right]_{i\alpha,j\beta}
		=
		\sum_{m=1}^{M} c_m
		\sum_{p=1}^{L_a}\sum_{\mu=1}^{2}
		\sum_{q=1}^{L_b}\sum_{\nu=1}^{2}
		\left(A_a^{(m)}\right)_{i\alpha,p\mu}
		F_{p\mu,q\nu}
		\left(B_b^{(m)}\right)_{q\nu,j\beta},
		\label{eq:su2_primitive_contraction}
	\end{equation}
	where $c_m$ is an invariant scalar. Then $\mathcal{M}_{SU(2)}(F_{ab})$ satisfies the same SU(2) spinor transformation rule as Eq.~(\ref{eq:su2_full_block_equivariance}).
\end{theorem}

In Eq.~(\ref{eq:su2_primitive_contraction}),
$A_a^{(m)}$ can be instantiated by either the directional spinor tensor
$V_a^{\mathrm{SU(2)}}$ or the equivariant context tensor
$N_a^{\mathrm{SU(2)}}$ defined in
Appendix~\ref{subsec:app_operator_attention}. $B_b^{(m)}$ can be instantiated by either
$(V_b^{\mathrm{SU(2)}})^{\dagger}$ or
$(N_b^{\mathrm{SU(2)}})^{\dagger}$.

\subsection{Equivariance to Spatial Inversions}

\begin{theorem}[Spatial-inversion equivariance of operators in the
	direct-product space]
	\label{thm:inversion_equivariance}
	Assume that the column basis of $A_a^{(m)}$ and the row basis
	of $F_{ab}$ are the same ordered local basis of atom $a$,
	while the row basis of $B_b^{(m)}$ and the column basis of
	$F_{ab}$ are the same ordered local basis of atom $b$,
	including spin in the SOC setting.
	For local-set indices $u,u'\in\mathcal B_a$ and
	$v,v'\in\mathcal B_b$, assume that the units of $F_{ab}$,
	$A_a^{(m)}$, and $B_b^{(m)}$ satisfy
	\begin{align}
		F_{ab}^{u'\otimes v'}(\mathcal I X)
		&=
		\eta_F(-1)^{l_{u'}+l_{v'}}
		F_{ab}^{u'\otimes v'}(X),\\
		\left[A_a^{(m)}\right]^{u\otimes u'}(\mathcal I X)
		&=
		\eta_A^{(m)}(-1)^{l_u+l_{u'}}
		\left[A_a^{(m)}\right]^{u\otimes u'}(X),\\
		\left[B_b^{(m)}\right]^{v'\otimes v}(\mathcal I X)
		&=
		\eta_B^{(m)}(-1)^{l_{v'}+l_v}
		\left[B_b^{(m)}\right]^{v'\otimes v}(X).
	\end{align}
	If $c_m$ is inversion invariant, then the $u\otimes v$ unit
	generated by the $m$-th head in
	Eqs.~(\ref{eq:so3_primitive_contraction}) and
	(\ref{eq:su2_primitive_contraction}) satisfies
	\begin{equation}
		\left[\mathcal M_m(F_{ab})\right]^{u\otimes v}(\mathcal I X)
		=
		\eta_A^{(m)}\eta_F\eta_B^{(m)}
		(-1)^{l_u+l_v}
		\left[\mathcal M_m(F_{ab})\right]^{u\otimes v}(X).
		\label{eq:inversion_parity_product_rule}
	\end{equation}
	Therefore, for a Hamiltonian-shaped input with $\eta_F=+1$,
	requiring $\eta_A^{(m)}\eta_B^{(m)}=+1$ for every retained
	head ensures that their sum preserves the Hamiltonian
	inversion parity $(-1)^{l_u+l_v}$ defined in
	Eq.~(\ref{eq:hamiltonian_unit_inversion_parity})
	for every output unit.
\end{theorem}

For the non-SOC model, this result together with the established $SO(3)$
equivariance gives full $O(3)$ equivariance. For SOC systems, it combines
the $SU(2)$ equivariance with the corresponding orbital-parity
action. 

\begin{proposition}[Equivariance of invariantly weighted operator aggregation]
	\label{prop:canonical_context_equivariance}
	If feature matrices or tensors $F_e$ are
	expressed in the same carrier space and common local basis, with matching
	direct-product-unit structure, dimensions, and equivariant transformation
	laws, and share the same operator-level intrinsic inversion parity
	$\eta_F$, then
	$
		N_v
		=
		\sum_{e:\,\mathrm{dst}(e)=v}\alpha_e F_e
	$
	obeys the same rotation and inversion laws as the aggregated features
	$F_e$, provided that the weights $\alpha_e$ are invariant scalars.
\end{proposition}
\paragraph{Implication.} Scalar features can flexibly modulate the equivariant features without changing their transformation laws under rotations and inversions, thus enhancing the model capacity.

Proofs of all theorems and propositions are provided in Appendix~\ref{app:theory_proofs}.

\section{Methodology}
\label{sec:method}

\begin{figure}
	\centering    
	\includegraphics[scale=0.395]{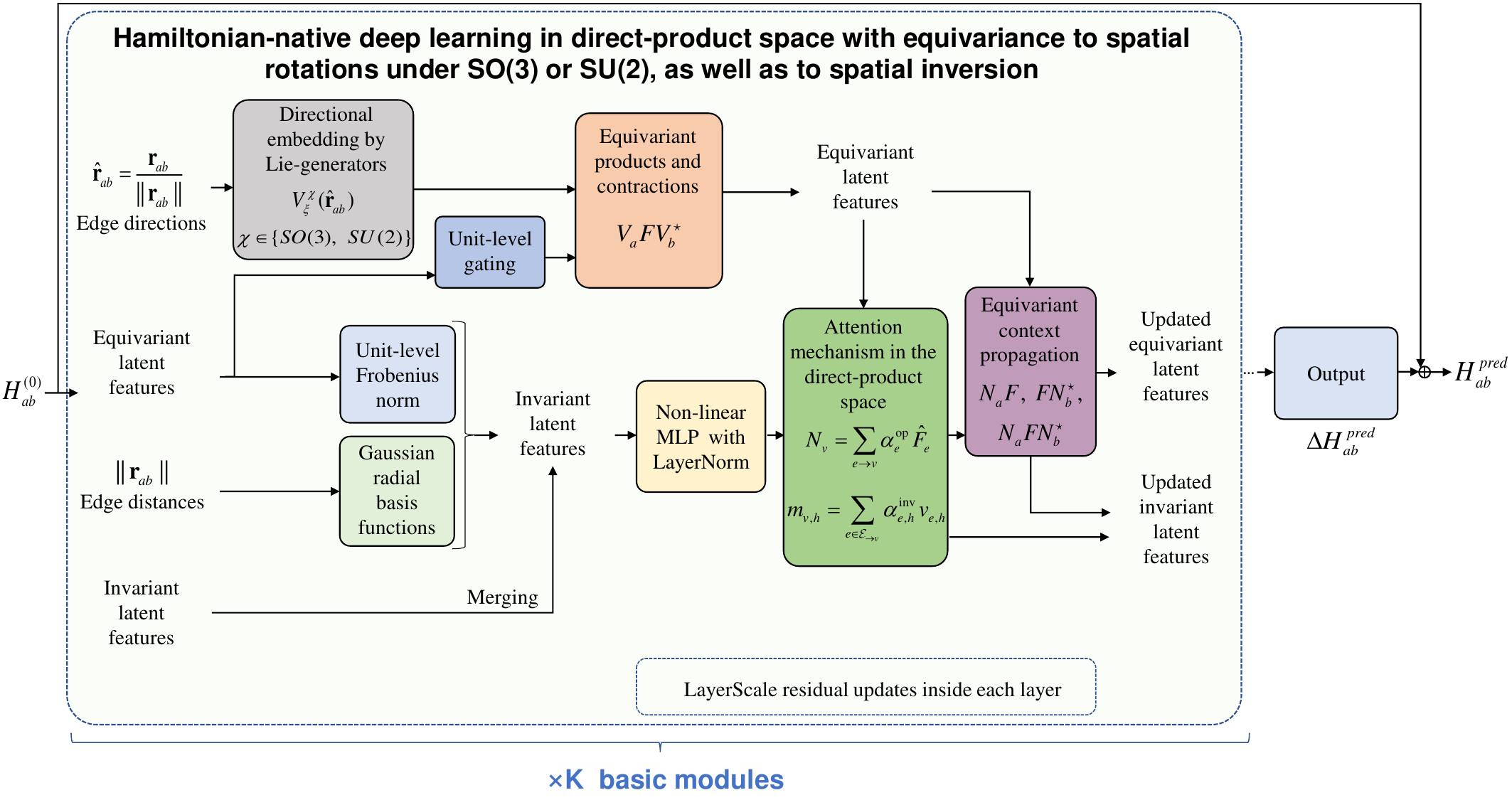}
	\caption{
		Overview of the neural network architecture of DP-Learner.
	}
	\label{network}
\end{figure}

We instantiate our theory as an equivariant and expressive Transformer architecture that jointly
learns equivariant features in the direct-product space and invariant
scalar features. Following the residual learning strategy of
NextHAM~\citep{yin2026advancing}, DP-Learner takes the zeroth-step
Hamiltonian $H^{(0)}$ as input and learns a
residual correction to predict the self-consistent Hamiltonian
$H^{(T)}$. We represent the equivariant atom-pair features propagated
between layers as $H^{(0)}$ plus accumulated corrections in the
target orbital or orbital--spin basis, as defined in
Appendix~\ref{subsec:app_residual_initialization}. Directional
embeddings, attention, and matrix or tensor contractions combine bond
directions with many-body environment information while preserving
$SO(3)$ or spinor $SU(2)$ equivariance and spatial-inversion
equivariance. Figure~\ref{network} summarizes the architecture.
The components described below are detailed in the corresponding
subsections of Appendix~\ref{app:architecture_details}, and the
training loss is given in Appendix~\ref{app:loss}.
Appendix~\ref{app:complexity_efficiency} analyzes the computational
efficiency relative to representation learning methods~\citep{e3nn,scmr} in the direct-sum space.

\paragraph{Dual conversion between equivariant and invariant features.}
DP-Learner gains non-linear expressive power from invariant scalar features while preserving equivariance in the direct-product space. Normalized Frobenius norms summarize each direct-product unit of the current matrix or tensor. Multi-layer perceptrons (MLPs) combine these summaries with invariant node, edge, and radial features to predict gates, directional weights, attention weights, and contraction coefficients. These invariant scalars then modulate the equivariant features without changing their transformation laws. Appendix~\ref{subsec:app_invariant_equivariant_routing} gives the corresponding equations.

\paragraph{Equivariant updates using directions.}
We encode the pair direction $\hat{\mathbf r}_{ab}$ using fixed orbital and spin Lie generators to construct directional operators acting on the local orbital or orbital--spin bases of atoms $a$ and $b$. These operators update the equivariant atom-pair feature through a bilateral contraction, injecting bond-orientation information into the direct-product space. An invariant scalar coefficient controls the strength of this update.  Details are given in Appendices~\ref{app:generator} and~\ref{subsec:app_directional_lifting}.

\paragraph{Attention-based updates of node, edge, and context features.}
The attention module updates invariant node and edge features and constructs equivariant atom-wise context features. Multi-head attention aggregates invariant edge features to update the destination-node features. In parallel, attention over incoming equivariant features assigns invariant weights and aggregates the matrices or tensors in a common local basis to form the context $N_v$. The updated node features, radial features, current edge features, and new Frobenius summaries then determine the invariant edge update. Appendix~\ref{subsec:app_attention_context} describes these operations.

\paragraph{Equivariant updates using context features.}
The context features at atoms $a$ and $b$ update the corresponding equivariant atom-pair feature through left, right, and bilateral products or contractions. Invariant coefficients predicted from the local environment control the three contributions, propagating many-body information from neighboring Hamiltonian blocks.  The resulting correction is accumulated for the next layer. Appendix~\ref{subsec:app_operator_update} gives the complete equations.

\paragraph{Stabilization techniques and $H^{(0)}$-centered residual correction.}
Dimension-aware normalization controls the growth of numerical scales of equivariant features during repeated contractions: per-unit Frobenius summaries use root-mean-square magnitudes, and one-sided and bilateral contractions are scaled by $D^{-1/2}$ and $D^{-1}$, respectively, where $D$ is the matrix dimension in the common basis layout. LayerNorm operates within invariant MLPs, while LayerScale regulates the equivariant, edge, and node updates. The output heads are zero initialized so that the initial prediction equals $H^{(0)}$, and directional weights use $w=1+\Delta w$ to retain the fixed generator contribution. After $K$ layers, the output module produces the Hamiltonian correction $\Delta H^{\mathrm{pred}}$, which is added to $H^{(0)}$ to obtain the final prediction. Appendix~\ref{subsec:app_norm_stability} describes the initialization and output.

\section{Dataset}
\label{data}
To enable Hamiltonian learning at a broader scale and with greater diversity, we construct and release \textbf{MP-HAM-110K}, comprising more than $110{,}000$ material structures sampled from the Materials Project~\citep{jain2013commentary}, for which self-consistent first-principles calculations are performed to obtain Hamiltonian annotations. MP-HAM-110K provides zeroth-step Hamiltonians $H^{(0)}$, self-consistent Hamiltonians $H^{(T)}$, overlap matrices, and the wavefunction information required for reciprocal-space supervision. As discussed in Appendix~\ref{benchmark}, the dataset is formulated in the non-SOC orbital $SO(3)$ setting to bridge the gap in existing data resources and provide a scalable complement to computationally demanding SOC datasets. This design enables efficient data generation, learning, and inference over a broad materials space. Further details are provided in Appendix~\ref{data_detail}.

\begin{table}[t]
	\centering
	\caption{
		Accuracy and computational efficiency comparison on the Materials-HAM-SOC
		benchmark \citep{yin2026advancing} and the proposed MP-HAM-110K benchmark.
		$N_{\theta}$ denotes the total number of network parameters.
		As detailed in Appendix~\ref{app:implementation}, $\overline{\mathrm{MAE}}_{\mathrm{G}}$ denotes the Gauge MAE metric;
		$\overline{t}_{\mathrm{NN}}$ denotes the mean per-sample runtime of the
		neural inference stage, including input
		feeding, neural-network execution, result format conversion and saving; 
		$\overline{M}_{\mathrm{GPU}}$ denotes the mean per-sample peak GPU memory
		reserved by PyTorch.
		Lower values indicate better performance for all metrics.
	}
	\label{tab:main_res}
	
	\small
	\setlength{\tabcolsep}{2.0pt}
	\renewcommand{\arraystretch}{1.10}
	
	\begin{tabular*}{\columnwidth}{
			@{\extracolsep{\fill}}
			l
			cccc
			cccc
			@{}
		}
		\toprule
		
		\multirow{2}{*}{Method}
		& \multicolumn{4}{c}{Materials-HAM-SOC }
		& \multicolumn{4}{c}{MP-HAM-110K }
		\\
		
		\cmidrule(lr){2-5}
		\cmidrule(lr){6-9}
		
		&
		$N_{\theta}$
		& $\overline{\mathrm{MAE}}_{\mathrm{G}}$
		& $\overline{t}_{\mathrm{NN}}$
		& $\overline{M}_{\mathrm{GPU}}$
		&
		$N_{\theta}$
		& $\overline{\mathrm{MAE}}_{\mathrm{G}}$
		& $\overline{t}_{\mathrm{NN}}$
		& $\overline{M}_{\mathrm{GPU}}$
		\\[-1pt]
		
		&
		{\scriptsize (M)}
		& {\scriptsize (meV)}
		& {\scriptsize (s)}
		& {\scriptsize (GiB)}
		&
		{\scriptsize (M)}
		& {\scriptsize (meV)}
		& {\scriptsize (s)}
		& {\scriptsize (GiB)}
		\\
		
		\midrule
		
		NextHAM
		& 50.285
		& 1.417
		& 3.014
		& 20.398
		& 50.263
		& 2.394
		& 2.032
		& 6.135
		\\
		
		DP-Learner
		& \textbf{26.445}
		& \textbf{0.841}
		& \textbf{0.269}
		& \textbf{0.989}
		& \textbf{26.094}
		& \textbf{1.595}
		& \textbf{0.203}
		& \textbf{0.531}
		\\
		
		\bottomrule
	\end{tabular*}
\end{table}

\section{Empirical Study}
Given the page limit, we present only the key experimental
results in the main text. Ablation studies, element-wise
error analyses, and additional band-structure comparisons
are comprehensively provided in Appendix~\ref{app:ablation} and are also
referenced in the analysis below. We compare the prediction accuracy and computational efficiency
of DP-Learner with NextHAM~\citep{yin2026advancing}, a recent
state-of-the-art method for Hamiltonian prediction across a broad
range of elements. Experiments are conducted on Materials-HAM-SOC
and MP-HAM-110K, both of which cover chemically diverse periodic
materials and provide the zeroth-step Hamiltonians required by
the two methods. Other Hamiltonian benchmarks often focus on
specific material systems, lack publicly accessible data, or
do not provide $H^{(0)}$, limiting their suitability for this
comparison, as discussed in Appendix~\ref{benchmark}.
The software and hardware environments, parameter settings, and
training and inference configurations for both methods are
detailed in Appendix~\ref{app:implementation}.

As shown in Table~\ref{tab:main_res}, DP-Learner improves both
accuracy and efficiency relative to NextHAM. On Materials-HAM-SOC,
it reduces Gauge MAE~\citep{wang2024universal} by $40.6\%$,
peak GPU memory consumption by $95.2\%$, and the parameter
count by $47.4\%$, while achieving an $11.2\times$ speedup
in neural inference. On MP-HAM-110K, it reduces Gauge MAE
by $33.4\%$, peak GPU memory consumption by $91.3\%$, and
the parameter count by $48.1\%$, with a $10.0\times$
inference speedup. Together, these results establish a new state of the art on both
benchmarks in Hamiltonian prediction accuracy and inference
efficiency. Preserving the Hamiltonian structure provides an
inductive bias for learning how the local environment affects
orbital or orbital--spin couplings, while equivariant matrix
contractions enable efficient batched computation. Table~\ref{tab:ablation_dp_learner} in
Appendix~\ref{app:ablation} further shows substantial reductions
in Gauge MAE relative to using $H^{(0)}$ directly on both datasets.
Figure~\ref{fig:band_h0_compare} illustrates the improvement in
band structures: DP-Learner closely reproduces the DFT reference
bands, whereas $H^{(0)}$ exhibits pronounced deviations.

The results on MP-HAM-110K further demonstrate the efficiency
of the non-SOC formulation. Although its testing structures contain
approximately twice as many atoms on average as those in
Materials-HAM-SOC, DP-Learner requires substantially less inference
time and GPU memory. It is worth noting that the higher reported MAE on MP-HAM-110K
does not by itself indicate poorer regression performance.
The reported errors also reflect the different block
structures of SOC and non-SOC Hamiltonians: the SOC error is
averaged over four spin blocks, whereas the non-SOC error is
evaluated over a single orbital block. In SOC systems, $H^{(0)}$
already closely approximates the spin-flip blocks, as shown by
the statistics in NextHAM~\citep{yin2026advancing}. The reported
SOC error is averaged over all four spin blocks, including the
two spin-flip blocks with negligible errors. Consequently, the
mean error over the two spin-conserving blocks is approximately
twice the overall error. After accounting for this averaging,
the errors are comparable to those on the single orbital block
in the non-SOC setting. 

\begin{figure}
	\centering    
	\includegraphics[scale=0.77]{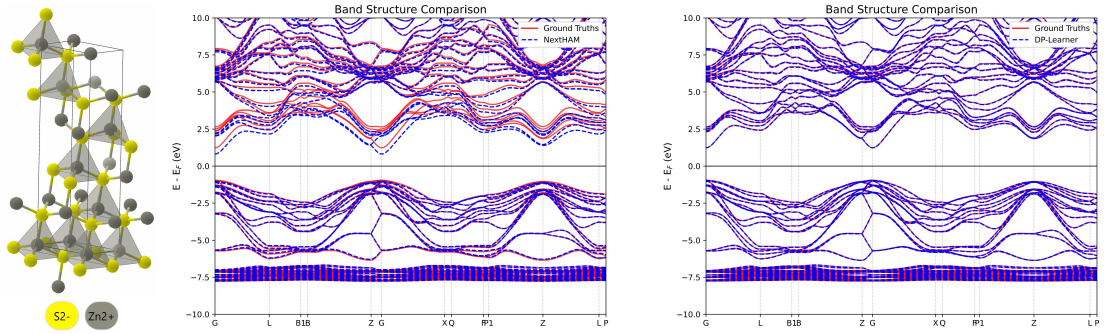}
	\caption{Band structure comparison for a representative  testing sample from  Materials-HAM-SOC. Red solid lines show the DFT reference bands used as ground truths in our dataset; blue dashed lines show the bands obtained from the Hamiltonians predicted by NextHAM (left) and DP-Learner (right). Additional DP-Learner results are shown in Figures~\ref{fig:band_res1} and~\ref{fig:band_res2}. The displayed energies are referenced to the Fermi energy $E_F$.
	}
	\label{band_compare}
\end{figure}

 Figures~\ref{fig:compare_table1} and~\ref{fig:compare_table2}
in Appendix~\ref{app:ablation} compare prediction accuracy across
element-defined subsets of the Materials-HAM-SOC and MP-HAM-110K
testing sets, respectively. For each element, we collect all of the testing
structures containing it and compute the Gauge MAE metric within each subset.  DP-Learner outperforms NextHAM in every subset shown, with
consistent accuracy improvements across all of the
elements represented in these testing sets. These results demonstrate
that the accuracy gains brought by DP-Learner cover chemically diverse materials,
rather than being confined to structures containing a few
specific elements. The gains in Hamiltonian prediction accuracy also translate into
more accurate band structures. For the ZnS example in
Figure~\ref{band_compare}, DP-Learner substantially reduces the
deviations present in the prediction results of NextHAM, particularly in
the band gap, and closely reproduces the DFT reference bands.
Figures~\ref{fig:band_res1} and~\ref{fig:band_res2} provide
further evidence of band structure agreement of our method with DFT calculations across
diverse materials, including structurally complex systems
with many closely spaced bands. These results demonstrate the practical value of DP-Learner
for electronic-structure analysis and highlight its potential
for industrial applications in which electrical conductivity
and optical properties are critical. The ablation studies on both datasets in
Table~\ref{tab:ablation_dp_learner} demonstrate the individual
contributions of five core components to prediction accuracy: updating equivariant features across
layers, incorporating bond directions, aggregating equivariant
context features, predicting invariant coefficients from the
local environment, and scaling residual feature updates with
LayerScale. The full model combines all five components and
achieves the lowest prediction error on both datasets. Taken together, these results support the central premise of
DP-Learner: learning equivariant features in the direct-product
space with preservation of the Hamiltonian structure supports both
accurate prediction and efficient computation.

\section{Conclusion}
We present an equivariant Hamiltonian prediction framework that operates directly on Hamiltonian-like operator blocks in the direct-product space. By preserving the native matrix or spinor-tensor structure of Hamiltonians throughout message passing, DP-Learner avoids the representation mismatch and irregular tensor transformations between direct-product and direct-sum spaces used by conventional methods. The framework combines Lie-generator-based directional embedding with invariantly weighted left--right contractions, ensuring exact rotation and inversion equivariance. These principles are instantiated as an equivariant Transformer architecture, where dense left--right operator products and contractions are guided by non-linear invariant scalar routing over gates, directional propagation, attention, and environmental context. We further present MP-HAM-110K, a comprehensive benchmark with over 110,000 periodic-material structures across 68 elements, supporting systematic evaluation of universal Hamiltonian prediction at substantially larger scale. Experiments on MP-HAM-110K and Materials-HAM-SOC demonstrate that DP-Learner achieves superior Hamiltonian prediction accuracy and dramatic gains in inference efficiency compared to the state-of-the-art NextHAM method.

\section*{AI use statement}
LLMs were used for manuscript language polishing to improve fluency and grammatical correctness.

\section*{Ethics Statement}

This work aims to advance deep-learning methods for materials research. Although no immediate adverse social or ethical consequences are apparent, it remains important to consider the broader implications of data-driven scientific modeling. DP-Learner provides a physically informed framework for learning Hamiltonians directly in their native direct-product space. This design improves the alignment between the model architecture and the underlying quantum-mechanical representation of the prediction target, while substantially improving computational efficiency. Nevertheless, the detailed roles of the learned invariant routing coefficients, attention weights, and environment-dependent operator interactions remain only partially understood. Further analysis of these learned quantities may clarify how physical information is encoded and utilized by the model, thereby strengthening the scientific interpretability, reliability, and applicability of its predictions across diverse materials systems.

\bibliography{sn-bibliography2}
\bibliographystyle{iclr2027_conference}

\newpage
\appendix

\section{Overview of Basic Concepts in Group Theory}
\label{concept}

This section presents the group-theoretic definitions used throughout the
paper. The presentation is intentionally brief and focuses on the concepts relevant to Hamiltonian blocks. For a more
comprehensive physical introduction, we refer the reader to standard
texts on group theory in solid-state physics
\citep{dresselhaus2007group}.

\begin{definition}[Group]
	A \emph{group} is a pair $(\mathcal{G},\cdot)$, where $\mathcal{G}$ is a nonempty set and
	\[
	\cdot:\mathcal{G}\times\mathcal{G}\rightarrow\mathcal{G}
	\]
	is a binary operation satisfying the following axioms:
	\begin{enumerate}
		\item \textbf{Closure.}
		For any $g_1,g_2\in\mathcal{G}$,
		\[
		g_1\cdot g_2\in\mathcal{G}.
		\]
		
		\item \textbf{Associativity.}
		For any $g_1,g_2,g_3\in\mathcal{G}$,
		\[
		(g_1\cdot g_2)\cdot g_3
		=
		g_1\cdot(g_2\cdot g_3).
		\]
		
		\item \textbf{Identity.}
		There exists an element $e\in\mathcal{G}$, called the identity element,
		such that for every $g\in\mathcal{G}$,
		\[
		e\cdot g=g\cdot e=g.
		\]
		
		\item \textbf{Inverse.}
		For every $g\in\mathcal{G}$, there exists an element
		$g^{-1}\in\mathcal{G}$, called the inverse of $g$, such that
		\[
		g\cdot g^{-1}
		=
		g^{-1}\cdot g
		=
		e.
		\]
	\end{enumerate}
\end{definition}

\begin{definition}[Linear representation]
	Let $V$ be a finite-dimensional real or complex vector space. A
	representation of $\mathcal{G}$ on $V$ is a group homomorphism
	\[
	\rho:\mathcal{G}\rightarrow GL(V),
	\qquad
	\rho(g_1g_2)=\rho(g_1)\rho(g_2),
	\quad
	\rho(e)=I.
	\]
	After choosing a basis of $V$, $\rho(g)$ is represented by an
	invertible matrix. In this paper, the relevant representations are
	the orbital representations of $SO(3)$ and their orbital--spin
	extensions induced by the $SU(2)$ lift of spatial rotations.
\end{definition}

\begin{definition}[Invariant subspace and irreducible representation]
	A subspace $U\subseteq V$ is invariant under $\rho$ if
	\[
	\rho(g)U\subseteq U,
	\qquad
	\forall g\in\mathcal{G}.
	\]
	Equivalently, every vector initially in $U$ remains in $U$ after the
	group action. The representation $\rho$ is irreducible if its only
	invariant subspaces are $\{0\}$ and $V$. Otherwise, $\rho$ is
	reducible and, for compact groups, can be decomposed after a suitable
	change of basis into a direct sum of irreducible representations.
\end{definition}

\begin{definition}[Equivariant and invariant maps]
	Let $\rho_V$ and $\rho_W$ be representations of $\mathcal{G}$ on
	vector spaces $V$ and $W$. A map $f:V\rightarrow W$ is equivariant if
	\[
	f(\rho_V(g)v)
	=
	\rho_W(g)f(v),
	\qquad
	\forall g\in\mathcal{G},\ v\in V.
	\]
	If $W$ carries the trivial representation, this condition reduces to
	invariance:
	\[
	f(\rho_V(g)v)=f(v).
	\]
	In DP-Learner, scalar networks act on invariant quantities, while
	matrix or spinor-tensor states carry nontrivial equivariant
	representations.
\end{definition}

\begin{definition}[$SO(3)$ and $SU(2)$]
	The special orthogonal group is
	\[
	SO(3)
	=
	\left\{
	R\in\mathbb{R}^{3\times3}:
	R^\top R=I,\ \det R=1
	\right\},
	\]
	the group of proper rotations in three-dimensional Euclidean space.
	The special unitary group is
	\[
	SU(2)
	=
	\left\{
	U\in\mathbb{C}^{2\times2}:
	U^\dagger U=I,\ \det U=1
	\right\}.
	\]
	The group $SU(2)$ is the double cover of $SO(3)$ and provides the
	spinor representation associated with spatial rotations. Thus, a
	$2\pi$ rotation acts as $-I$ on spin-$1/2$ states, while a $4\pi$
	rotation returns to $I$.
\end{definition}

\begin{definition}[Wigner representations and spinor lift]
	For each integer angular-momentum label $l\geq0$, let $V_l$ denote
	the $(2l+1)$-dimensional orbital representation space. A spatial
	rotation $R\in SO(3)$ acts on $V_l$ through a Wigner representation
	matrix
	\[
	D^l(R)\in\mathbb{R}^{(2l+1)\times(2l+1)}
	\]
	in a real spherical-harmonic basis, or through the equivalent complex
	Wigner matrix in a complex spherical basis.
	
	In the SOC setting, the orbital space is tensored with the
	spin-$1/2$ space. If
	$U^{1/2}(R)\in SU(2)$ denotes the corresponding spinor lift of $R$,
	then the orbital--spin representation used in this paper is
	\[
	\widetilde{D}^l(R)
	=
	D^l(R)\otimes U^{1/2}(R).
	\]
\end{definition}

\begin{definition}[Direct-product space and representation]
	\label{def:direct_product_space_representation}
	Throughout this paper, following common usage in physics, the term
	\emph{direct-product space} refers to the tensor-product space
	$V_1\otimes V_2$.
	
	Let $V_1$ and $V_2$ be finite-dimensional vector spaces with bases
	$\{e_i\}_{i=1}^{m}$ and $\{f_j\}_{j=1}^{n}$, respectively. Their
	direct-product space is
	\[
	V_1\otimes V_2
	=
	\operatorname{span}
	\left\{
	e_i\otimes f_j
	\right\}_{i=1,\ldots,m;\,j=1,\ldots,n}.
	\]
	The basis degrees of freedom are therefore indexed jointly by a pair
	$(i,j)$, and
	\[
	\dim(V_1\otimes V_2)
	=
	(\dim V_1)(\dim V_2)
	=
	mn.
	\]
	A general element $T\in V_1\otimes V_2$ can be written as
	\[
	T
	=
	\sum_{i=1}^{m}
	\sum_{j=1}^{n}
	T_{ij}\,e_i\otimes f_j,
	\]
	so its coefficients naturally form a two-index array.
	
	If
	\[
	\rho_1:\mathcal{G}\rightarrow GL(V_1),
	\qquad
	\rho_2:\mathcal{G}\rightarrow GL(V_2)
	\]
	are representations of a group $\mathcal{G}$, their direct-product
	representation acts simultaneously on the two factors:
	\[
	(\rho_1\otimes\rho_2)(g)
	(v_1\otimes v_2)
	=
	\rho_1(g)v_1
	\otimes
	\rho_2(g)v_2.
	\]
	
	For the operator spaces relevant to this work, let
	$\{f^j\}_{j=1}^{n}$ denote the dual basis of $V_2^{*}$. Here the
	superscript ${}^{*}$ denotes the algebraic dual space. The canonical
	isomorphism
	\[
	\operatorname{Hom}(V_2,V_1)
	\cong
	V_1\otimes V_2^{*}
	\]
	identifies a linear operator $Q:V_2\rightarrow V_1$ with
	\[
	Q
	=
	\sum_{i=1}^{m}
	\sum_{j=1}^{n}
	Q_{ij}\,e_i\otimes f^j,
	\]
	whose coefficients form an $m\times n$ matrix. The induced group
	action is
	\[
	Q
	\mapsto
	\rho_1(g)\,
	Q\,
	\rho_2(g)^{-1}.
	\]
	For orthogonal representations,
	$\rho_2(g)^{-1}=\rho_2(g)^\top$, whereas for unitary representations,
	$\rho_2(g)^{-1}=\rho_2(g)^\dagger$.
\end{definition}

\begin{definition}[Direct-sum space and representation]
	\label{def:direct_sum_space_representation}
	Let $V_1$ and $V_2$ be finite-dimensional vector spaces with bases
	$\{e_i\}_{i=1}^{m}$ and $\{f_j\}_{j=1}^{n}$, respectively. Their
	\emph{direct-sum} space is
	\[
	V_1\oplus V_2
	=
	\left\{
	(v_1,v_2):
	v_1\in V_1,\ v_2\in V_2
	\right\},
	\]
	with basis
	\[
	\left\{
	(e_i,0)
	\right\}_{i=1}^{m}
	\cup
	\left\{
	(0,f_j)
	\right\}_{j=1}^{n}.
	\]
	The two sets of basis degrees of freedom are concatenated rather than
	pairwise combined, and hence
	\[
	\dim(V_1\oplus V_2)
	=
	\dim V_1+\dim V_2
	=
	m+n.
	\]
	A general element is written as
	\[
	v_1\oplus v_2
	=
	(v_1,v_2).
	\]
	
	If
	\[
	\rho_1:\mathcal{G}\rightarrow GL(V_1),
	\qquad
	\rho_2:\mathcal{G}\rightarrow GL(V_2)
	\]
	are representations of $\mathcal{G}$, their direct-sum representation acts separately
	on the two components:
	\[
	(\rho_1\oplus\rho_2)(g)(v_1,v_2)
	=
	\bigl(
	\rho_1(g)v_1,\,
	\rho_2(g)v_2
	\bigr).
	\]
	In a basis adapted to the two summands, the corresponding
	representation matrix is block diagonal:
	\[
	(\rho_1\oplus\rho_2)(g)
	=
	\begin{bmatrix}
		\rho_1(g) & 0\\
		0 & \rho_2(g)
	\end{bmatrix}.
	\]
	Thus, a vector initially belonging to one summand remains in that
	summand under the group action. More generally,
	$\bigoplus_{\lambda}V_{\lambda}$ carries the block-diagonal
	representation
	\[
	\bigoplus_{\lambda}\rho_{\lambda}(g).
	\]
	In summary, the direct-product space pairwise combines basis indices
	and has multiplicative dimension, whereas the direct-sum space
	concatenates separate basis sets and has additive dimension.
\end{definition}

\begin{definition}[Direct-product unit as a local coupling operator] 
	\label{def:direct_product_unit} 
	Let $V_u$ and $V_v$ be the orbital subspaces spanned by two local 
	orbital sets $u$ and $v$, with angular degrees $l_u$ and $l_v$. 
	We write $\operatorname{Hom}(X,Y)$ for the vector space of all linear 
	maps from $X$ to $Y$. For finite-dimensional spaces, there is a 
	canonical isomorphism 
	\[ 
	\operatorname{Hom}(X,Y) 
	\cong 
	Y\otimes X^*. 
	\] 
	A \emph{direct-product unit} coupling the $v$-side orbital set to the 
	$u$-side orbital set is therefore an operator 
	\[ 
	\mathbf{Q}^{\,u\otimes v} 
	\in 
	\operatorname{Hom}(V_v,V_u) 
	\cong 
	V_u\otimes V_v^*. 
	\] 
	In an orthonormal real orbital basis, this unit is represented by a 
	matrix 
	\[ 
	\mathbf{Q}^{\,u\otimes v} 
	\in 
	\mathbb{R}^{(2l_u+1)\times(2l_v+1)} 
	\] 
	and transforms under an active spatial rotation as 
	\begin{equation} 
		\mathcal{T}_R 
		\left( 
		\mathbf{Q}^{\,u\otimes v} 
		\right) 
		= 
		D^{l_u}(R)\, 
		\mathbf{Q}^{\,u\otimes v}\, 
		D^{l_v}(R)^\top. 
	\end{equation} 
	If the complete operator block $\mathbf Q$ is inversion homogeneous 
	with operator-level intrinsic parity $\eta_Q\in\{+1,-1\}$, shared by 
	all of its non-zero direct-product units, then 
	\[ 
	\mathbf{Q}^{\,u\otimes v}(\mathcal I X) 
	= 
	\eta_Q(-1)^{l_u+l_v} 
	\mathbf{Q}^{\,u\otimes v}(X). 
	\] 
	Here, the factor $(-1)^{l_u+l_v}$ arises from the orbital parities of 
	the row and column subspaces, while $\eta_Q$ characterizes the 
	intrinsic inversion parity of the operator itself. A complete atom-pair 
	Hamiltonian block is assembled from all direct-product units associated 
	with the local orbital sets on the two atoms. The full-block rotation 
	and inversion laws for Hamiltonians, together with their corresponding 
	extension to complex orbital--spin bases in the SOC setting, are 
	formalized in Appendix~\ref{app:symmetry_background}. 
\end{definition}

\begin{definition}[Clebsch--Gordan decomposition]
	\label{def:cg_decomposition}
	As introduced in Definition~\ref{def:direct_product_unit}, the local
	orbital subspaces $V_u$ and $V_v$ carry irreducible $SO(3)$
	representations of angular degrees $l_u$ and $l_v$, respectively.
	We therefore write
	\[
	V_u \cong V_{l_u},
	\qquad
	V_v \cong V_{l_v},
	\]
	so that a direct-product unit can be regarded as an operator
	\[
	\mathbf{Q}^{\,u\otimes v}
	\in
	V_{l_u}\otimes V_{l_v}^{*}.
	\]
	
	Choose the magnetic basis
	$\{\lvert l_u,m_u\rangle\}_{m_u=-l_u}^{l_u}$ of $V_{l_u}$ and the
	dual basis
	$\{\langle l_v,m_v\rvert\}_{m_v=-l_v}^{l_v}$ of $V_{l_v}^{*}$.
	The operator can then be written as
	\begin{equation}
		\mathbf{Q}^{\,u\otimes v}
		=
		\sum_{m_u=-l_u}^{l_u}
		\sum_{m_v=-l_v}^{l_v}
		Q^{u\otimes v}_{m_u,m_v}\,
		\lvert l_u,m_u\rangle
		\otimes
		\langle l_v,m_v\rvert,
	\end{equation}
	where $Q^{u\otimes v}_{m_u,m_v}$ are the matrix elements of
	$\mathbf{Q}^{\,u\otimes v}$ in the direct-product basis.
	
	To express the same operator in irreducible angular-momentum
	components, we use the fact that every integer-angular-momentum
	irreducible representation of $SO(3)$ is self-dual:
	\[
	V_{l_v}^{*}
	\cong
	V_{l_v}.
	\]
	After choosing a fixed equivariant identification between these two
	spaces, the Clebsch--Gordan decomposition gives
	\begin{equation}
		V_{l_u}\otimes V_{l_v}^{*}
		\cong
		V_{l_u}\otimes V_{l_v}
		\cong
		\bigoplus_{L=|l_u-l_v|}^{l_u+l_v}V_L.
	\end{equation}
	This decomposition reorganizes the matrix elements of
	$\mathbf{Q}^{\,u\otimes v}$ into irreducible components with total
	angular momentum $L$.
	
Using the coefficient tensor
$C^{\,L,\,l_u,\,l_v}_{M,\,m_u,\,m_v}$, the Clebsch--Gordan
decomposition maps the direct-product matrix/tensor components into
the following irreducible components in the  direct-sum space:
\begin{equation}
	q_M^L
	=
	\sum_{m_u,m_v}
	C^{\,L,\,l_u,\,l_v}_{M,\,m_u,\,m_v}
	Q^{u\otimes v}_{m_u,m_v},
	\qquad
	L=|l_u-l_v|,\ldots,l_u+l_v,
\end{equation}
where $M=-L,\ldots,L$. 

The original direct-product matrix/tensor components can
 be recovered through the inverse transformation:
\begin{equation}
	Q^{u\otimes v}_{m_u,m_v}
	=
	\sum_{L=|l_u-l_v|}^{l_u+l_v}
	\sum_{M=-L}^{L}
	\operatorname{conj}\!\left(
	C^{\,L,\,l_u,\,l_v}_{M,\,m_u,\,m_v}
	\right)
	q_M^L.
\end{equation}

\end{definition}

\section{Electronic Structure Calculations: from Density Functional Theory to Deep Learning Methods}
\label{eed}

Density Functional Theory (DFT)~\citep{hohenberg1964inhomogeneous,kohn1965self} is one of the central computational frameworks for electronic-structure calculations in condensed matter physics, quantum chemistry, and materials science. Instead of explicitly treating the full many-electron wavefunction, DFT uses the electron density \( \rho(\mathbf r) \) as the fundamental variable. This reformulation greatly reduces the complexity of the quantum many-body problem while retaining the essential information required for many ground-state properties. Owing to this balance between physical fidelity and computational tractability, DFT has become a standard tool for computing band structures, orbital energies, relaxed geometries, and electronic, magnetic, optical, and transport-related properties.

The practical formulation most widely used in materials calculations is Kohn--Sham DFT~\citep{kohn1965self}. It maps the interacting many-electron problem to an auxiliary single-particle problem,
\begin{equation}
	\hat{H} \psi_i(\mathbf r) = \epsilon_i \psi_i(\mathbf r),
	\qquad
	\hat{H}
	=
	-\frac{\hbar^2}{2m}\nabla^2
	+ V_{\mathrm{ext}}(\mathbf r)
	+ V_{\mathrm{HXC}}[\rho](\mathbf r),
	\label{eq:KS}
\end{equation}
where \( \hat H \) is the effective Kohn--Sham Hamiltonian. The potential consists of the external potential \(V_{\mathrm{ext}}\) and the Hartree--exchange--correlation contribution
\[
V_{\mathrm{HXC}}[\rho](\mathbf r)
=
V_{\mathrm{H}}[\rho](\mathbf r)
+
V_{\mathrm{XC}}[\rho](\mathbf r),
\]
where \(V_{\mathrm{H}}\) and \(V_{\mathrm{XC}}\) are both evaluated as functionals of the density. For periodic systems, the density is obtained from the occupied Kohn--Sham orbitals as
\begin{equation}
	\rho(\mathbf r)
	=
	\sum_{\mathbf k} w_{\mathbf k}
	\sum_{n}
	f_{n\mathbf k}
	\left|\psi_{n\mathbf k}(\mathbf r)\right|^2 ,
	\label{psi2rho}
\end{equation}
where \(w_{\mathbf k}\) is the weight of the \(\mathbf k\)-point and \(f_{n\mathbf k}\) denotes the occupation of band \(n\) at \(\mathbf k\).

To solve Eq.~(\ref{eq:KS}) numerically, the orbitals are expanded in a finite basis. In localized-basis electronic-structure codes, numerical atomic orbitals (NAOs) are commonly used because of their compact spatial support and favorable efficiency. A basis function is written as the product of a radial function and a real spherical harmonic,
\begin{equation}
	\phi_{\kappa \zeta l m}(\mathbf r)
	=
	f_{\kappa \zeta l}(r)\,
	\widetilde{Y}_{lm}(\tilde{\mathbf r}),
	\label{eq:nao}
\end{equation}
where \(\kappa\) denotes the chemical species, \(\zeta\) labels the radial channel, \(l\) and \(m\) are the angular momentum and magnetic quantum numbers, and \(\tilde{\mathbf r}\) denotes the angular direction of \(\mathbf r\). The radial part is tabulated on a radial grid and is truncated beyond a cutoff radius \(r_c\). In many implementations it can be represented by a finite spherical-Bessel expansion,
\begin{equation}
	f_{\kappa \zeta l}(r)
	=
	\begin{cases}
		\displaystyle
		\sum_q c_{\kappa \zeta l q} j_l(qr),
		& r < r_c, \\[6pt]
		0,
		& r \ge r_c .
	\end{cases}
\end{equation}

Let \(\alpha=(i,u)\) be a composite index denoting the \(u\)-th atomic orbital centered on atom \(i\) in the reference unit cell, and define
\[
\phi_{\alpha\mathbf R}(\mathbf r)
=
\phi_u(\mathbf r-\tau_i-\mathbf R),
\]
where \(\tau_i\) is the atomic position within the unit cell and \(\mathbf R\) is a lattice vector. The Bloch eigenstates are then expanded as
\begin{equation}
	\psi_{n\mathbf k}(\mathbf r)
	=
	\frac{1}{\sqrt{N_k}}
	\sum_{\mathbf R}
	\sum_{\alpha}
	C_{n\alpha,\mathbf k}
	e^{i\mathbf k\cdot\mathbf R}
	\phi_{\alpha\mathbf R}(\mathbf r),
	\label{ks-eig}
\end{equation}
where \(C_{n\alpha,\mathbf k}\) is the expansion coefficient of orbital \(\alpha\) for band \(n\) at \(\mathbf k\), and \(N_k\) is the number of unit cells in the Born--von--K\'arm\'an supercell, equivalently the number of sampled \(\mathbf k\)-points.

Substituting Eq.~(\ref{ks-eig}) into the Kohn--Sham equation gives the generalized eigenvalue problem
\begin{equation}
	H(\mathbf k) C_{\mathbf k}
	=
	S(\mathbf k) C_{\mathbf k} E_{\mathbf k},
	\label{eq:HKS0}
\end{equation}
where \(H(\mathbf k)\), \(S(\mathbf k)\), and \(C_{\mathbf k}\) are the Hamiltonian matrix, overlap matrix, and eigenvector matrix at \(\mathbf k\), respectively. The diagonal matrix \(E_{\mathbf k}\) contains the Kohn--Sham eigenvalues \(\epsilon_{n\mathbf k}\).

The reciprocal-space Hamiltonian is obtained from real-space matrix elements. For orbitals \(\phi_{\alpha 0}\) and \(\phi_{\beta\mathbf R}\), with
\[
\phi_{\alpha 0}(\mathbf r)=\phi_u(\mathbf r-\tau_i),
\qquad
\phi_{\beta\mathbf R}(\mathbf r)=\phi_v(\mathbf r-\tau_j-\mathbf R),
\]
the real-space Hamiltonian element is
\begin{equation}
	H_{\alpha\beta}(\mathbf R)
	=
	\left\langle
	\phi_{\alpha 0}
	\middle|
	-\frac{\hbar^2}{2m}\nabla^2
	+ V_{\mathrm{ext}}
	+ V_{\mathrm{HXC}}[\rho]
	\middle|
	\phi_{\beta\mathbf R}
	\right\rangle .
	\label{ham_construct}
\end{equation}
The corresponding matrix at \(\mathbf k\) is obtained through the lattice Fourier transform
\begin{equation}
	H_{\alpha\beta}(\mathbf k)
	=
	\sum_{\mathbf R}
	e^{i\mathbf k\cdot\mathbf R}
	H_{\alpha\beta}(\mathbf R).
\end{equation}
The overlap matrix is transformed in the same way:
\begin{equation}
	S_{\alpha\beta}(\mathbf k)
	=
	\sum_{\mathbf R}
	e^{i\mathbf k\cdot\mathbf R}
	S_{\alpha\beta}(\mathbf R),
	\qquad
	S_{\alpha\beta}(\mathbf R)
	=
	\langle \phi_{\alpha 0} | \phi_{\beta\mathbf R} \rangle .
\end{equation}

A standard Kohn--Sham calculation proceeds through a self-consistent-field (SCF) iteration:
\begin{enumerate}
	\item \textbf{Initialization}: choose an initial density \( \rho^{(0)}(\mathbf r) \) and set the iteration index \(t=0\).
	\item \textbf{Potential update}: construct \(V_{\mathrm{HXC}}[\rho^{(t)}](\mathbf r)\) from the current density.
	\item \textbf{Hamiltonian assembly}: build the Hamiltonian matrix \(\mathbf H^{(t)}\) using Eq.~(\ref{ham_construct}).
	\item \textbf{Eigenproblem solution}: Fourier transform the real-space matrices to reciprocal space and solve Eq.~(\ref{eq:HKS0}) to obtain \(\psi_{n\mathbf k}^{(t)}(\mathbf r)\) and \(\epsilon_{n\mathbf k}^{(t)}\).
	\item \textbf{Density update}: compute the next density \(\rho^{(t+1)}(\mathbf r)\) from the occupied orbitals using Eq.~(\ref{psi2rho}).
	\item \textbf{Convergence check}: update \(t\leftarrow t+1\) and repeat until the input and output densities satisfy a prescribed convergence criterion.
\end{enumerate}
Schematically, the procedure can be written as
\begin{small}
	\[
	\rho^{(0)}(\mathbf r)
	\rightarrow
	V_{\mathrm{HXC}}[\rho^{(0)}](\mathbf r)
	\rightarrow
	\mathbf H^{(0)}
	\rightarrow
	\{\psi^{(0)}_{n\mathbf k},\epsilon^{(0)}_{n\mathbf k}\}
	\rightarrow
	\rho^{(1)}(\mathbf r)
	\rightarrow
	\cdots
	\rightarrow
	\rho^{(T)}(\mathbf r)
	\rightarrow
	\mathbf H^{(T)} .
	\]
\end{small}
After convergence, \(\mathbf H^{(T)}\) defines the self-consistent electronic structure and can be used to compute band structures, orbital energies, total energies, and related electronic, magnetic, optical, and transport properties.

Although Kohn--Sham DFT has been highly successful across materials science, chemistry, and physics~\citep{nagy1998density,jones2015density}, its computational cost remains substantial for complex systems. A major cost arises from solving the generalized eigenvalue problem in Eq.~(\ref{eq:HKS0}); dense diagonalization scales cubically with the matrix dimension, i.e.,
\(\mathcal O(M_{\mathrm{bas}}^3)\), where \(M_{\mathrm{bas}}\) is the number of basis functions. For localized atomic-orbital calculations with a fixed number of basis functions per atom, \(M_{\mathrm{bas}}\) grows linearly with the number of atoms \(N\), so the dense eigensolution scales as \(\mathcal O(N^3)\). Since the eigenvalue problem must be solved repeatedly within
an SCF loop requiring $T$ iterations, a dense eigensolver can
incur a total computational cost of $\mathcal O(TN^3)$,
imposing a substantial computational burden.

Recent machine-learning approaches~\citep{schutt2019unifying,unke2021se,li2022deep,gong2023general,DBLP:conf/icml/YuXQQJ23,DBLP:conf/iclr/LiXHWHYLWZLSG25,luo2025efficient,tracegrad} address this bottleneck by learning to predict the converged Hamiltonian \(H_{\alpha\beta}^{(T)}\) directly from atomic structures and related descriptors. In this setting, the expensive SCF trajectory is replaced by a single neural forward pass that outputs the final Hamiltonian matrix elements, thereby bypassing the repeated full-matrix eigensolutions required during conventional self-consistency.

The locality of NAOs gives the Hamiltonian a sparse block structure. Because basis functions have finite spatial extent, \(H_{\alpha\beta}(\mathbf R)\) is non-zero only when the corresponding atoms lie within an interaction cutoff. If \(N\) is the number of atoms and \(\overline{E}\) denotes the average number of neighboring atoms within this cutoff, then the number of local atomic pairs, and hence the number of Hamiltonian blocks to predict, scales as \(\mathcal O(N\overline{E})\). For small systems in which \(\overline{E}\) grows with \(N\), the number of non-zero blocks can exhibit \(N^2\) scaling. In sufficiently large systems, however, \(\overline{E}\) approaches a constant, and the number of non-zero Hamiltonian blocks scales linearly as \(\mathcal O(N)\).

Once the Hamiltonian has been predicted, many downstream electronic properties do not require the full eigenspectrum. In particular, electronic and transport properties are often governed primarily by bands within a limited energy window around the Fermi level. In such cases, one may compute only the targeted eigenstates near this energy window using sparse iterative eigensolvers, including shift-invert methods implemented in ARPACK~\citep{Lehoucq1998}. When only a small fraction of the spectrum is required, the resulting computational cost can be substantially lower than that of full diagonalization. Learned Hamiltonian models therefore eliminate the repeated diagonalizations of the SCF procedure and can further combine sparse Hamiltonian representations with targeted local eigensolutions for downstream analysis. For very large materials systems, this combination provides a scalable pathway in which Hamiltonian prediction can approach \(\mathcal O(N)\) scaling, while the subsequent spectral calculation is restricted to the physically relevant energy window near the Fermi level, allowing the relevant electronic states to be computed efficiently. This substantially improves the feasibility of large-scale materials analysis, screening, and design, compared to the traditional DFT method.

\section{Symmetry of Hamiltonians}
\label{app:symmetry_background}

This section provides a rigorous formulation of the symmetry laws obeyed by Hamiltonians under spatial rotations and inversion, expanding the compact statements given in Section~\ref{sec:preliminary}. The associated group-theoretic concepts have been introduced in Definitions~\ref{def:direct_product_space_representation}, \ref{def:direct_sum_space_representation}, and~\ref{def:direct_product_unit}.

\paragraph{Hamiltonian-block and direct-product-unit organization.}
For an atomic pair $(a,b)$, the complete atom-pair Hamiltonian block $H_{ab}$ couples the localized basis of atom $b$ to that of atom $a$. Let $u\in\mathcal B_a$ and $v\in\mathcal B_b$ index local orbital sets on the two atoms, with angular degrees $l_u$ and $l_v$, respectively. Let $V_u$ and $V_v$ denote the corresponding local orbital subspaces. The direct-product unit $H_{ab}^{\,u\otimes v}$ is a linear map from $V_v$ to $V_u$, or equivalently an element of
\[
V_u\otimes V_v^{*}.
\]
The complete atom-pair block is assembled from all such direct-product units:
\[
H_{ab}
=
\big[H_{ab}^{\,u\otimes v}\big]_
{u\in\mathcal B_a,\,v\in\mathcal B_b}.
\]
Each unit represents one row--column coupling between a pair of local orbital sets and is associated with the angular-momentum-degree pair $(l_u,l_v)$. 

For systems without spin-orbit coupling, the Hamiltonian block is typically represented in a real orbital basis. Under an active spatial rotation $R\in SO(3)$, the real spherical-harmonic basis associated with angular-momentum label $l$ transforms according to the orthogonal Wigner-$D$ representation matrix
\begin{equation}
	D^{l}(R)\in\mathbb{R}^{(2l+1)\times(2l+1)} .
\end{equation}
Each direct-product unit satisfies
\begin{equation}
	\mathcal{T}_R
	\left(
	H_{ab}^{\,u\otimes v}
	\right)
	=
	D^{l_u}(R)
	H_{ab}^{\,u\otimes v}
	D^{l_v}(R)^{\top}.
\end{equation}
Defining the atom-wise block-diagonal representation matrices
\begin{equation}
	D_a(R)=\bigoplus_{u\in\mathcal B_a}D^{l_u}(R),
	\qquad
	D_b(R)=\bigoplus_{v\in\mathcal B_b}D^{l_v}(R),
\end{equation}
gives the block-level equivariant form in Eq.~(\ref{eq:so3_full_block_equivariance}).

When spin-orbit coupling is included, the basis expands from pure orbital states to orbital--spin states $|l,m_l\rangle\otimes |s,m_s\rangle$. Each direct-product unit becomes a complex spinor tensor
\begin{equation}
	H_{ab}^{\,u\otimes v}
	\in
	\mathbb{C}^{(2l_u+1)\times 2\times(2l_v+1)\times 2},
\end{equation}
with components $(H_{ab}^{\,u\otimes v})_{m_u\alpha,m_v\beta}$, where $m_u,m_v$ index magnetic basis components and $\alpha,\beta\in\{1,2\}$ index spin.

For a spatial rotation $R$, let $U^{1/2}(R)\in\mathbb{C}^{2\times 2}$ denote its spin-$1/2$ representation, i.e., the corresponding $SU(2)$ lift. The orbital--spin representation for angular-momentum label $l$ is
\begin{equation}
	\widetilde{D}^{l}(R)
	=
	D^{l}(R)\otimes U^{1/2}(R).
\end{equation}
The spinor Hamiltonian block transforms through contractions over both orbital and spin indices:
\begin{equation}
	\mathcal{T}_R
	\left(
	(H_{ab}^{\,u\otimes v})_{m_u\alpha,m_v\beta}
	\right)
	=
	\sum_{n_u\mu,n_v\nu}
	\widetilde{D}^{l_u}_{m_u\alpha,n_u\mu}(R)
	(H_{ab}^{\,u\otimes v})_{n_u\mu,n_v\nu}
	\widetilde{D}^{l_v\dagger}_{n_v\nu,m_v\beta}(R).
\end{equation}
At the full-block level, defining
\begin{equation}
	\widetilde{D}_a(R)=\bigoplus_{u\in\mathcal B_a}\widetilde{D}^{l_u}(R),
	\qquad
	\widetilde{D}_b(R)=\bigoplus_{v\in\mathcal B_b}\widetilde{D}^{l_v}(R),
\end{equation}
gives the block-level equivariant form in Eq.~(\ref{eq:su2_full_block_equivariance}).

\paragraph{Spatial-inversion representation.}
Let $\mathcal I$ denote spatial inversion, $\mathbf r\mapsto-\mathbf r$.
Since an orbital subspace of angular degree $l$ has parity $(-1)^l$ and
spin is an axial degree of freedom, we can define the parity action on each
local orbital set $u\in\mathcal B_\xi$ by
\begin{equation}
    \mathsf P_{\xi,u}
    =
    \begin{cases}
        (-1)^{l_u}I_{2l_u+1}, & \text{non-SOC},\\
        (-1)^{l_u}I_{2l_u+1}\otimes I_2, & \text{SOC},
    \end{cases}
    \qquad
    \mathsf P_\xi
    =
    \bigoplus_{u\in\mathcal B_\xi}\mathsf P_{\xi,u}.
    \label{eq:inversion_parity_operator}
\end{equation}
The physical Hamiltonian is inversion homogeneous with operator-level
intrinsic parity $\eta_H=+1$. Its direct-product units obey
\begin{equation}
    H_{ab}^{u\otimes v}(\mathcal I X)
    =
    \mathsf P_{a,u}H_{ab}^{u\otimes v}(X)\mathsf P_{b,v}^{\star}
    =
    (-1)^{l_u+l_v}H_{ab}^{u\otimes v}(X).
    \label{eq:hamiltonian_unit_inversion_parity}
\end{equation}
Assembling all units gives the equivalent full-block law
\begin{equation}
    H_{ab}(\mathcal I X)
    =
    \mathsf P_aH_{ab}(X)\mathsf P_b^\star.
    \label{eq:hamiltonian_block_inversion_parity}
\end{equation}

More generally, an operator block $Q_{ab}$ is called inversion
homogeneous with operator-level (block-level) intrinsic parity
$\eta_Q\in\{+1,-1\}$ if the same $\eta_Q$ is shared by all of its
non-zero direct-product units and
\begin{equation}
	Q_{ab}^{u\otimes v}(\mathcal I X)
	=
	\eta_Q\,
	\mathsf P_{a,u}Q_{ab}^{u\otimes v}(X)\mathsf P_{b,v}^{\star}
	=
	\eta_Q(-1)^{l_u+l_v}Q_{ab}^{u\otimes v}(X).
	\label{eq:operator_intrinsic_inversion_parity}
\end{equation}
For the Hamiltonian block, $\eta_Q=\eta_H=+1$, whereas the
Lie-generator-based directional embedding has intrinsic parity
$\eta_V=-1$.

\section{Related Work}
\label{related_work}

\subsection{Equivariant Neural Networks for Electronic-Structure Hamiltonian Prediction: Current Status and Bottlenecks}

Symmetry is a fundamental principle in physics and is closely connected
to conservation laws. Symmetry-preserving neural networks have therefore
become a standard paradigm for learning atomistic quantities from
three-dimensional structures. Early models such as SchNet construct
scalar representations from invariant geometric descriptors, primarily
interatomic distances \citep{schutt2018schnet}. Subsequent equivariant
architectures propagate vectors and higher-order tensors transforming
under irreducible representations of spatial rotations, using
equivariant linear maps, tensor products, Clebsch--Gordan decompositions, and
non-linear gating mechanisms
\citep{thomas2018tensor,e3nn,gasteiger2021gemnet,batzner20223,batatia2022mace}.
By retaining directional information, these models provide richer
representations than purely invariant scalar networks and have become
the foundation of modern deep learning models of atomic systems.

Deep-learning Hamiltonian models directly predict Kohn--Sham
Hamiltonians from material structures, bypassing expensive
self-consistent iterations and enabling efficient calculations of band
structures, densities of states, and related electronic properties. For
example, the original DeepH framework uses local coordinates to transform
the rotation-equivariant Hamiltonian prediction problem into an invariant
learning problem \citep{li2022deep}; DeepH-E3 upgrades this local-frame
invariant formulation with explicit $SO(3)$-equivariant message passing
over irreducible tensor features and Wigner--Eckart reconstruction,
providing a more expressive representation of directional orbital
couplings and supporting spin--orbit coupling
\citep{gong2023general}; DeepH-2 retains a bond-aligned local $z$-axis, realizes global
$SO(3)$ equivariance through more efficient $SO(2)$ operations, and
combines this formulation with an equivariant Transformer
\citep{wang2024universal}. Other studies improve transferability,
many-body modeling, sparsity, non-linear expressiveness, or tensor-product
efficiency
\citep{DBLP:conf/icml/YuXQQJ23,maceH,zhong2023transferable,
	zhong2024universal,luo2025efficient,tracegrad,liang2026hot}.

Many existing Hamiltonian-prediction models use learnable element
embeddings for atoms and, in some architectures, explicit element-pair
embeddings for atom pairs. These chemical descriptors are typically
initialized randomly and optimized from the training data, with limited
explicit electronic-structure priors.
Consequently, element-dependent orbital characteristics and
element-pair-dependent electronic couplings must be inferred largely
through data-driven learning, which can limit generalization across chemically
diverse systems. Because elements and element pairs are highly imbalanced
in materials databases, embeddings associated with rare elements,
unusual bonds, and underrepresented chemical environments receive
limited supervision and can generalize poorly to unseen material
systems. Moreover, maintaining separate lookup tables requires
$\mathcal{O}(M)$ parameters for element embeddings and, when explicit
element-pair embeddings are used, up to $\mathcal{O}(M^2)$ parameters
for edge embeddings, where $M$ is the number of covered elements. The
resulting data sparsity and rapidly growing representation cost pose
substantial challenges to constructing universal electronic-structure
Hamiltonian models that span broad regions of the periodic table.

NextHAM addresses these limitations by replacing conventional node and
edge embeddings with the zeroth-step Hamiltonian $H^{(0)}$ as a
physically informed input descriptor \citep{yin2026advancing}. Unlike
randomly initialized element and element-pair embeddings, $H^{(0)}$
incorporates electronic-structure information from the atomic
configuration and initial charge density, including electron--ion
interactions and an initial approximation to electron--electron
interactions. This substantially alleviates the data-sparsity
problem associated with rare elements and infrequent element pairs,
while avoiding embedding tables whose size grows with the number of
chemical species, making the input representation naturally scalable to
broad chemical spaces. At the output stage, NextHAM only learns the
residual correction
\begin{equation}
	\Delta H = H^{(T)}-H^{(0)}
\end{equation}
rather than reconstructing the self-consistent Hamiltonian from scratch.
NextHAM has achieved remarkable cross-material generalization over a
chemically diverse materials space containing more than 60 elements
\citep{yin2026advancing}. These results demonstrate the effectiveness of
$H^{(0)}$-conditioned learning as a practical route toward universal
electronic-structure Hamiltonian prediction.

However, a fundamental representation mismatch remains in most of the existing
equivariant methods for Hamiltonian prediction: the Hamiltonian is
naturally an operator in the direct-product space, whereas the neural network usually
represents and propagates information through a direct sum of
irreducible tensor features. Specifically, an atom-pair Hamiltonian block
is a linear map from the local basis space of atom $b$ to that of atom
$a$,
\begin{equation}
	H_{ab}
	\in
	\operatorname{Hom}(V_b,V_a)
	\cong
	V_a\otimes V_b^{*},
	\qquad
	H_{ab}\in\mathbb{C}^{d_a\times d_b}.
\end{equation}
The full block is assembled from orbital units in the direct-product space,
\begin{equation}
	H_{ab}
	=
	\left[
	H_{ab}^{u\otimes v}
	\right]_{u\in\mathcal B_a,\,v\in\mathcal B_b},
	\qquad
	H_{ab}^{u\otimes v}
	\in
	V_u\otimes V_v^{*},
\end{equation}
where, for orbital angular momenta $l_u$ and $l_v$,
\begin{equation}
	H_{ab}^{u\otimes v}
	\in
	\mathbb{C}^{(2l_u+1)\times(2l_v+1)}.
\end{equation}
Thus, in a fixed localized basis, each direct-product unit is naturally
represented as a regular matrix with explicit row--column structure.

However, most of the existing equivariant Hamiltonian models
\citep{gong2023general,DBLP:conf/icml/YuXQQJ23,zhong2023transferable,wang2024universal,zhong2024universal,liang2026hot,maceH}
adopt direct sums of irreducible tensor features as their persistent internal
representation. For an orbital pair in the $SO(3)$ setting, the
corresponding direct-product representation decomposes as
\begin{equation}
	V_u\otimes V_v^{*}
	\cong
	\bigoplus_{L=|l_u-l_v|}^{l_u+l_v} V_L,
	\qquad
	\dim V_L = 2L+1,
	\label{eq:product_to_directsum}
\end{equation}
so that a complete coupling is represented by a collection of
irreducible channels
\begin{equation}
	\left\{
	q_{ab}^{uv,L}\in V_L
	\right\}_{L=|l_u-l_v|}^{l_u+l_v}.
\end{equation}
Although the total number of degrees of freedom is preserved,
\begin{equation}
	(2l_u+1)(2l_v+1)
	=
	\sum_{L=|l_u-l_v|}^{l_u+l_v}(2L+1),
\end{equation}
a regular direct-product matrix is fragmented and
reorganized into multiple irreducible tensor channels of heterogeneous
sizes. 

Even in NextHAM, where the input $H^{(0)}$ is already a Hamiltonian
operator in its native direct-product space, the corresponding
blocks are still transformed into irreducible tensor
features in the direct-sum space, propagated in that feature space, and subsequently
reconstructed to obtain the Hamiltonian residual ($\Delta H$)
\citep{yin2026advancing} in the direct-product space. Thus, although $H^{(0)}$ provides a strong
Hamiltonian-level physical prior, its native direct-product structure is
not preserved in the internal representation throughout the
network.

This representation mismatch introduces both representation-learning
bottlenecks and computational overheads. From a representation-learning
perspective, representing a Hamiltonian block that naturally lives in
the direct-product space with direct sums of irreducible tensor features
fragments its coherent row--column coupling structure across multiple
angular-momentum channels, making the underlying orbital-coupling
relationships less explicit in the internal representation. From a
computational perspective, representing complete Hamiltonian matrices as
direct sums of irreducible tensors and processing them without information
loss suffers from high computational complexity. At the hardware-execution level,
the heterogeneous dimensions $(2L+1)$ implied by
Eq.~(\ref{eq:product_to_directsum}) lead to irregular feature layouts and
angular-momentum-dependent routing paths, as analyzed in
Appendix~\ref{app:complexity_efficiency}. The
introduction of spin degrees of freedom in SOC systems
further increases this heterogeneity. Practical implementations therefore
involve extensive irregular indexing and gathering, masking, padding,
coefficient lookup, and many small tensor-product operations, which are
poorly matched to the dense matrix-multiplication kernels highly
optimized on modern graphics processing units, particularly for rich
basis sets containing $d$- and $f$-orbitals.

Our framework is designed to address this representation and computational bottleneck.
Since the Hamiltonian is intrinsically an operator in the direct-product space, we
develop encoding and message-passing mechanisms that keep
Hamiltonian-like features in the same direct-product matrix or tensor
space throughout the network. This aligns the internal neural
representation with the native representation of the Hamiltonian, while
reducing computational complexity and enabling efficient execution
through dense matrix operations and hardware acceleration. As demonstrated
in our experiments, this design yields substantial advantages in both
prediction accuracy and computational efficiency in terms of time and
memory cost.

\subsection{Benchmarks and Data Resources for Deep Learning of Hamiltonians: Current Landscape and Remaining Gaps}
\label{benchmark}

Large-scale datasets such as QH9 and $\nabla^2$DFT
\citep{QH9,khrabrov2024nabla2dft} provide valuable benchmarks for
molecular Hamiltonian learning. However, their elemental coverage is
restricted to five elements (H, C, N, O, and F) in QH9 and eight
elements (H, C, N, O, F, S, Cl, and Br) in $\nabla^2$DFT.
Both datasets focus on finite molecular systems and do not provide
Hamiltonian data for periodic crystals. Consequently, despite their large scale, these datasets lack the broad elemental coverage and periodic crystal structures needed to train and evaluate generalizable Hamiltonian models across chemically diverse periodic materials.

In contrast, publicly available Hamiltonian datasets for periodic
materials remain limited. Early studies have largely focused on individual material classes
or system-specific benchmarks, such as graphene, MoS$_2$,
Bi$_2$Se$_3$, and amorphous HfO$_2$
\citep{li2022deep,gong2023general,xia2025learning}.
More recent work, such as MACE-H~\citep{maceH}, has likewise
evaluated periodic Hamiltonian prediction on material
datasets covering C, Mo, S, Bi, Se, Te, and Au. Although these
studies established the feasibility of deep Hamiltonian prediction,
their restricted material scopes limit both the training and systematic
evaluation of cross-material generalization, preventing them from
serving as large-scale universal benchmarks for diverse periodic
materials.

More recent efforts have extended Hamiltonian learning toward broader
material spaces. DeepH-2~\citep{wang2024universal} and
HamGNN~\citep{zhong2024universal} expand the elemental and structural
coverage beyond earlier system-specific benchmarks. However, the
associated datasets and electronic-structure resources are not yet
available as fully open and standardized benchmarks for unified training
and evaluation across broad materials spaces. In addition, compromises
in dataset construction and basis selection may limit physical coverage.
For example, DeepH-2 reduces computational complexity by adopting a
truncated basis without $f$-orbitals, which may affect the description
of materials in which higher-angular-momentum orbitals play important
roles.

The emerging $H^{(0)}$-conditioned paradigm for Hamiltonian prediction, such as
NextHAM~\citep{yin2026advancing}, requires benchmark datasets to provide
a broader range of electronic-structure quantities for developing
generalizable models. Such methods require not only the final
self-consistent Hamiltonian, but also the zeroth-step Hamiltonian,
overlap matrices, and additional electronic-structure information for
reciprocal-space supervision.
Materials-HAM-SOC~\citep{yin2026advancing} addresses this need by
providing a broad SOC Hamiltonian benchmark containing $17{,}000$
periodic structures spanning 68 elements, together with
high-quality localized basis sets, zeroth-step Hamiltonians,
self-consistent Hamiltonians, and overlap matrices. Its explicit
treatment of SOC provides a stringent benchmark for spinor Hamiltonian
learning, but also substantially increases the computational and memory
costs of data generation, training, and inference, and can make
self-consistent DFT calculations more difficult to converge.

From a physical perspective, explicit SOC effects are relatively
weak for many light-element systems. Although SOC is essential
for describing spin-dependent phenomena and topological
properties in many heavy-element compounds, scalar-relativistic
non-SOC calculations remain useful for studying equilibrium
structures and atomic dynamics, including in some systems
containing heavy elements. For equilibrium properties, a systematic PAW study of 56 solids
found that relativistic corrections beyond the scalar-relativistic
approximation caused small changes in the lattice constants
and bulk moduli of some systems containing fifth- or sixth-period
elements, while changes in the other systems examined were below
the numerical resolution~\citep{dalcorso2012projector}.
For atomic dynamics, a study of iodide-defect migration in
$\mathrm{CsPbI_3}$ found that SOC had no significant influence
on the forces in their evaluations.
The authors therefore omitted SOC from the force-field
training calculations to reduce computational
cost~\citep{tyagi2025tracing}. These findings illustrate the practical value of non-SOC
reference data for structural and dynamical studies and
motivate complementary SOC and non-SOC Hamiltonian benchmarks.
A large-scale non-SOC benchmark with broad chemical coverage,
rich orbital bases, and the electronic-structure annotations
required for $H^{(0)}$-conditioned learning therefore provides
a useful resource for scalable materials modeling.

To bridge this gap and provide a substantially larger and more scalable
data resource, we construct MP-HAM-110K, a benchmark dataset for
$H^{(0)}$-conditioned Hamiltonian learning in the non-SOC orbital
$SO(3)$ setting. MP-HAM-110K contains $113{,}800$ diverse periodic
material structures sampled from the Materials Project
\citep{jain2013commentary}, covers 68 elements from the first six rows
of the periodic table, and employs high-quality localized
numerical atomic-orbital basis sets with up to $4s2p2d1f$ orbitals per
element. For each structure, it provides the atomic structure,
zeroth-step Hamiltonian $H^{(0)}$, self-consistent Hamiltonian $H^{(T)}$,
overlap matrix $S$, band-structure references, and wavefunction-related
annotations required for reciprocal-space supervision. These quantities
support physically informed residual learning, and joint real- and reciprocal-space supervision.

Compared with Materials-HAM-SOC~\citep{yin2026advancing}, MP-HAM-110K
contains more than six times as many material structures and extends the
system size to unit cells containing up to 169 atoms, thereby providing
a broader testbed for large-scale and cross-material Hamiltonian
learning. At the same time, its non-SOC formulation avoids the additional
orbital--spin degrees of freedom and offers a computationally lighter
setting for data generation, model training, and inference.
The two datasets therefore serve complementary roles:
MP-HAM-110K supports scalable orbital Hamiltonian learning over a
substantially larger materials space, whereas Materials-HAM-SOC
provides a benchmark for spinor Hamiltonian prediction with explicit SOC.
Together, they enable systematic evaluation of generalizable
Hamiltonian models across both large-scale non-SOC and explicitly
spin--orbit-coupled settings.

\section{Construction of Directional Embeddings Using Lie Algebra Generators}
\label{app:generator}

To incorporate bond directions into equivariant feature updates in the direct-product space, we encode each direction as an equivariant operator acting on the local orbital or orbital--spin basis. The operator is constructed by contracting the Cartesian components of the direction with fixed Lie-algebra generators and combining the resulting channels with invariant scalar weights. Below, we define these generators and construct the corresponding directional embeddings for non-SOC and SOC systems.

\subsection{Orbital rotation generators}

We first define the matrices that describe rotations of the local orbital basis. A local orbital set $u$ contains the $2l_u+1$ basis functions with a common radial function and orbital angular momentum $l_u$. Rotations mix these basis functions through a real orthogonal matrix $D^{(l_u)}(R)$, which forms a $(2l_u+1)$-dimensional representation of $SO(3)$. The infinitesimal rotations about the three Cartesian axes are described by the fixed matrices
\begin{equation}
	\mathbf K^{(l_u)}
	=
	\left(K_x^{(l_u)},K_y^{(l_u)},K_z^{(l_u)}\right),
	\qquad
	K_{\rho}^{(l_u)}\in\mathbb R^{(2l_u+1)\times(2l_u+1)}.
\end{equation}
These generators are real and antisymmetric. They satisfy the $\mathfrak{so}(3)$ commutation relations,
\begin{equation}
	[K_x^{(l_u)},K_y^{(l_u)}]=K_z^{(l_u)},
\end{equation}
together with cyclic permutations. A finite rotation is obtained by exponentiating the generators:
\begin{equation}
	D^{(l_u)}(R)
	=
	\exp\!\left(\sum_{\rho\in\{x,y,z\}}\theta_{\rho}K_{\rho}^{(l_u)}\right),
\end{equation}
where $\boldsymbol\theta=\theta\hat{\mathbf n}$ specifies a rotation by angle $\theta$ about the unit axis $\hat{\mathbf n}$.

The full orbital basis of atom $\xi$ consists of a collection of local orbital sets, denoted by $\mathcal B_{\xi}$. Sets with different radial functions are treated separately, even if they have the same $l_u$. Since rotations act within each set, the generators and rotation matrices for the full orbital basis have the block-diagonal form
\begin{equation}
	K_{\xi,\rho}
	=
	\bigoplus_{u\in\mathcal B_{\xi}}K_{\rho}^{(l_u)},
	\qquad
	D_{\xi}(R)
	=
	\bigoplus_{u\in\mathcal B_{\xi}}D^{(l_u)}(R).
\end{equation}

\subsection{Orbital--spin generators}

For SOC systems, a spatial rotation acts on both the orbital and spin degrees of freedom. We therefore extend the orbital generators to the tensor product of the local orbital space and the two-dimensional spin space. Using $\hbar=1$, we define the Hermitian orbital angular-momentum generators and the spin-$1/2$ generators as
\begin{equation}
	L_{\xi,\rho}=iK_{\xi,\rho},
	\qquad
	S_{\rho}=\frac{1}{2}\sigma_{\rho},
\end{equation}
where $\sigma_{\rho}$ are the Pauli matrices. The generator of a simultaneous orbital and spin rotation is
\begin{equation}
	J_{\xi,\rho}
	=
	L_{\xi,\rho}\otimes I_2
	+
	I_{\mathrm{orb},\xi}\otimes S_{\rho},
\end{equation}
where $I_{\mathrm{orb},\xi}$ and $I_2$ are the identity matrices on the local orbital and spin spaces, respectively. The first term rotates the orbital components, and the second rotates the spin components. For the same axis--angle vector $\boldsymbol\theta$, the combined transformation is
\begin{equation}
	\widetilde{D}_{\xi}(R)
	=
	\exp\!\left(-i\sum_{\rho\in\{x,y,z\}}\theta_{\rho}J_{\xi,\rho}\right).
\end{equation}
This is the spinor rotation associated with $\boldsymbol\theta$. The convention $L_{\xi,\rho}=iK_{\xi,\rho}$ ensures that its orbital part agrees with $D_{\xi}(R)$ defined above.

\subsection{Directional embeddings from generator channels}

We now use these generators to encode a unit bond direction $\hat{\mathbf r}$ as an operator acting on the local basis of atom $\xi$. Each local orbital set provides an orbital generator channel and, in SOC systems, an additional spin generator channel. Within each channel, we contract the three Cartesian generator components with $\hat{\mathbf r}$ and multiply the result by an invariant scalar weight. This allows the contribution of each channel to depend on the local pair environment while preserving rotational equivariance.

Let $P_{\xi,u}$ denote the projector from the full local orbital space onto set $u\in\mathcal B_{\xi}$, including all of its $2l_u+1$ magnetic components. Because rotations preserve each set, $P_{\xi,u}$ commutes with $D_{\xi}(R)$. In the orbital--spin space, the corresponding projector $P_{\xi,u}\otimes I_2$ commutes with $\widetilde{D}_{\xi}(R)$.

For non-SOC systems, the orbital generator channel associated with set $u$ is
\begin{equation}
	G_{\xi,\rho}^{u,\mathrm{orb}}
	=
	P_{\xi,u}K_{\xi,\rho}P_{\xi,u},
	\qquad u\in\mathcal B_{\xi}.
	\label{eq:channel_generator_so3}
\end{equation}
This projection retains the generator block for set $u$ and sets all other blocks to zero. The directional orbital operator is then
\begin{equation}
	V_{\xi}^{\mathrm{SO(3)}}(\hat{\mathbf r})
	=
	\sum_{u\in\mathcal B_{\xi}}w_{\xi,u}^{\mathrm{orb}}
	\sum_{\rho\in\{x,y,z\}}
	\hat r_{\rho}G_{\xi,\rho}^{u,\mathrm{orb}},
	\label{eq:directional_operator_so3_theory}
\end{equation}
where $w_{\xi,u}^{\mathrm{orb}}$ is an invariant scalar function of the local pair environment. The same weight multiplies all three Cartesian components of a channel.

For SOC systems, we keep the orbital and spin contributions separate so that they can be assigned independent weights. Their generators are
\begin{equation}
	G_{\xi,\rho}^{u,\mathrm{orb}}
	=
	\left(P_{\xi,u}L_{\xi,\rho}P_{\xi,u}\right)\otimes I_2,
	\qquad
	G_{\xi,\rho}^{u,\mathrm{spin}}
	=
	P_{\xi,u}\otimes S_{\rho}.
	\label{eq:channel_generator_su2}
\end{equation}
The orbital channel acts on the orbital components within set $u$, while the spin channel acts on the spin components associated with that set. Combining the two gives the directional spinor operator
\begin{equation}
	V_{\xi}^{\mathrm{SU(2)}}(\hat{\mathbf r})
	=
	\sum_{u\in\mathcal B_{\xi}}
	\sum_{\rho\in\{x,y,z\}}
	\hat r_{\rho}
	\left[
	w_{\xi,u}^{\mathrm{orb}}G_{\xi,\rho}^{u,\mathrm{orb}}
	+
	w_{\xi,u}^{\mathrm{spin}}G_{\xi,\rho}^{u,\mathrm{spin}}
	\right],
	\label{eq:directional_operator_su2_theory}
\end{equation}
where $w_{\xi,u}^{\mathrm{orb}}$ and $w_{\xi,u}^{\mathrm{spin}}$ are independent invariant scalar functions of the local pair environment.

To express both constructions in a common notation, we collect their channel indices into
\begin{equation}
	\mathcal C_{\xi}^{\mathrm{SO(3)}}
	=
	\{(u,\mathrm{orb}):u\in\mathcal B_{\xi}\},
	\qquad
	\mathcal C_{\xi}^{\mathrm{SU(2)}}
	=
	\{(u,\mathrm{orb}),(u,\mathrm{spin}):u\in\mathcal B_{\xi}\}.
\end{equation}
Here $\lambda=(u,\mathrm{orb})$ or $(u,\mathrm{spin})$ identifies a channel, and $\chi\in\{\mathrm{SO(3)},\mathrm{SU(2)}\}$ specifies the non-SOC or SOC construction. Writing the corresponding generator and weight as $G_{\xi,\rho}^{\chi,\lambda}$ and $w_{\xi}^{\chi,\lambda}$, respectively, Eqs.~(\ref{eq:directional_operator_so3_theory}) and~(\ref{eq:directional_operator_su2_theory}) become
\begin{equation}
	V_{\xi}^{\chi}(\hat{\mathbf r})
	=
	\sum_{\lambda\in\mathcal C_{\xi}^{\chi}}
	\sum_{\rho\in\{x,y,z\}}
	w_{\xi}^{\chi,\lambda}\hat r_{\rho}G_{\xi,\rho}^{\chi,\lambda},
	\qquad
	\chi\in\{\mathrm{SO(3)},\mathrm{SU(2)}\}.
	\label{eq:directional_operator_unified}
\end{equation}

The directional embeddings defined above satisfy the equivariance laws for spatial rotations and spatial inversion stated in Theorem~\ref{theorem:covariance}, with operator-level intrinsic parity $\eta_V=-1$. The proofs are provided in Appendix~\ref{app:theory_proofs}.

\section{Proofs for Section~\ref{sec:theory}}
\label{app:theory_proofs}

\begin{proof}[Proof of Theorem~\ref{theorem:covariance}]
    We use the active rotation convention in Eqs.~(\ref{eq:so3_full_block_equivariance}) and~(\ref{eq:su2_full_block_equivariance}). Under a proper rotation, the three generators in each channel transform as the components of a vector:
\begin{equation}
	\mathcal R_{\xi}(R)
	G_{\xi,\rho}^{\chi,\lambda}
	\mathcal R_{\xi}(R)^{\star}
	=
	\sum_{\kappa\in\{x,y,z\}}
	R_{\kappa\rho}G_{\xi,\kappa}^{\chi,\lambda},
	\label{eq:generator_channel_covariance}
\end{equation}
where $\mathcal R_{\xi}(R)=D_{\xi}(R)$ and $\star=\top$ for non-SOC systems, while $\mathcal R_{\xi}(R)=\widetilde{D}_{\xi}(R)$ and $\star=\dagger$ for SOC systems.
    Applying Eq.~(\ref{eq:generator_channel_covariance}) to Eq.~(\ref{eq:directional_operator_so3_theory}) gives
    \begin{align}
        D_{\xi}(R)V_{\xi}^{\mathrm{SO(3)}}(\hat{\mathbf r})D_{\xi}(R)^{\top}
        &=
        \sum_{\lambda}w_{\xi}^{\mathrm{SO(3)},\lambda}
        \sum_{\rho}\hat r_{\rho}
        \left[D_{\xi}(R)G_{\xi,\rho}^{\mathrm{SO(3)},\lambda}D_{\xi}(R)^{\top}\right] \nonumber\\
        &=
        \sum_{\lambda}w_{\xi}^{\mathrm{SO(3)},\lambda}
        \sum_{\kappa}\left(\sum_{\rho}R_{\kappa\rho}\hat r_{\rho}\right)
        G_{\xi,\kappa}^{\mathrm{SO(3)},\lambda} \nonumber\\
        &=
        V_{\xi}^{\mathrm{SO(3)}}(R\hat{\mathbf r}).
    \end{align}
    Here $\sum_{\rho}R_{\kappa\rho}\hat r_{\rho}=(R\hat{\mathbf r})_{\kappa}$, and the invariant channel weights commute with the group action.

    For the SOC representation, the same calculation uses Hermitian conjugation and the SOC generator set:
    \begin{align}
        &\sum_{r\gamma,s\delta}
        (\widetilde{D}_{\xi})_{i\alpha,r\gamma}(R)
        \left(V_{\xi}^{\mathrm{SU(2)}}(\hat{\mathbf r})\right)_{r\gamma,s\delta}
        (\widetilde{D}_{\xi}^{\dagger})_{s\delta,p\mu}(R) \nonumber\\
        &=
        \sum_{\lambda}w_{\xi}^{\mathrm{SU(2)},\lambda}
        \sum_{\rho}\hat r_{\rho}
        \left[
        \sum_{r\gamma,s\delta}
        (\widetilde{D}_{\xi})_{i\alpha,r\gamma}(R)
        (G_{\xi,\rho}^{\mathrm{SU(2)},\lambda})_{r\gamma,s\delta}
        (\widetilde{D}_{\xi}^{\dagger})_{s\delta,p\mu}(R)
        \right] \nonumber\\
        &=
        \sum_{\lambda}w_{\xi}^{\mathrm{SU(2)},\lambda}
        \sum_{\kappa}(R\hat{\mathbf r})_{\kappa}
        (G_{\xi,\kappa}^{\mathrm{SU(2)},\lambda})_{i\alpha,p\mu} \nonumber\\
        &=
        \left(V_{\xi}^{\mathrm{SU(2)}}(R\hat{\mathbf r})\right)_{i\alpha,p\mu}.
    \end{align}
    This proves the proper-rotation statements in Eqs.~(\ref{eq:directional_operator_so3}) and~(\ref{eq:directional_operator_su2}).

    For spatial inversion, each generator channel acts within a fixed local orbital set. Since the parity operator is proportional to the identity on each such set, with an additional spin identity in the SOC setting, every channel satisfies
    \begin{equation}
        \mathsf P_{\xi}
        G_{\xi,\rho}^{\chi,\lambda}
        \mathsf P_{\xi}^{\star}
        =
        G_{\xi,\rho}^{\chi,\lambda},
        \qquad
        \chi\in\{\mathrm{SO(3)},\mathrm{SU(2)}\}.
        \label{eq:generator_inversion_even}
    \end{equation}
    Because the channel weights are invariant and $\mathcal I$ sends $\hat{\mathbf r}\mapsto-\hat{\mathbf r}$, Eq.~(\ref{eq:directional_operator_unified}) gives
    \begin{align}
        V_{\xi}^{\chi}(-\hat{\mathbf r})
        &=-\sum_{\lambda\in\mathcal C_{\xi}^{\chi}}
        \sum_{\rho}w_{\xi}^{\chi,\lambda}\hat r_{\rho}
        G_{\xi,\rho}^{\chi,\lambda} \nonumber\\
        &=-\sum_{\lambda\in\mathcal C_{\xi}^{\chi}}
        \sum_{\rho}w_{\xi}^{\chi,\lambda}\hat r_{\rho}
        \mathsf P_{\xi}G_{\xi,\rho}^{\chi,\lambda}\mathsf P_{\xi}^{\star} \nonumber\\
        &=-\mathsf P_{\xi}V_{\xi}^{\chi}(\hat{\mathbf r})\mathsf P_{\xi}^{\star}.
    \end{align}
    Taking $\star=\top$ in the non-SOC setting and $\star=\dagger$ in the SOC setting proves Eq.~(\ref{eq:directional_operator_inversion_odd}). Equivalently, every non-zero directional unit obeys
    \begin{equation}
        \left[V_{\xi}^{\chi}(-\hat{\mathbf r})\right]^{u\otimes u'}
        =
        \eta_V(-1)^{l_u+l_{u'}}
        \left[V_{\xi}^{\chi}(\hat{\mathbf r})\right]^{u\otimes u'},
        \qquad
        \eta_V=-1,
        \label{eq:directional_unit_inversion_odd}
    \end{equation}
    establishing that both directional embeddings are inversion homogeneous with operator-level intrinsic parity $-1$.
\end{proof}

\begin{proof}[Proof of Theorem~\ref{thm:so3_equivariance}]
	Under the convention in Eq.~(\ref{eq:so3_full_block_equivariance}), the relevant transformations are
\begin{align}
	\mathcal{T}_R(A_a^{(m)}) &= D_a(R) A_a^{(m)} D_a(R)^{\top}, \\
	\mathcal{T}_R(B_b^{(m)}) &= D_b(R) B_b^{(m)} D_b(R)^{\top}, \\
	\mathcal{T}_R(F_{ab}) &= D_a(R) F_{ab} D_b(R)^{\top}.
\end{align}
	Because $D_a(R)$ and $D_b(R)$ are orthogonal, $D_a(R)^{\top}D_a(R)=I_a$ and $D_b(R)^{\top}D_b(R)=I_b$. Applying the contraction after rotation gives
	\begin{align}
		\mathcal M_{SO(3)}(\mathcal T_R F_{ab})
		&=
		\sum_m c_m \bigl(D_a A_a^{(m)}D_a^{\top}\bigr)
		\bigl(D_a F_{ab}D_b^{\top}\bigr)
		\bigl(D_b B_b^{(m)}D_b^{\top}\bigr) \nonumber\\
		&=
		D_a\left(\sum_m c_m A_a^{(m)}F_{ab}B_b^{(m)}\right)D_b^{\top}
		=
		D_a(R)\mathcal M_{SO(3)}(F_{ab})D_b(R)^{\top},
	\end{align}
	where the argument $R$ has been suppressed in the intermediate expressions for readability. Thus the output obeys the same transformation rule as the input Hamiltonian block.
\end{proof}

\begin{proof}[Proof of Theorem~\ref{thm:su2_equivariance}]
	Under the convention in Eq.~(\ref{eq:su2_full_block_equivariance}),
	\begin{align}
		\mathcal{T}_R(A_a^{(m)})_{i\alpha,p\mu}
		&=\sum_{k\gamma,r\rho}(\widetilde{D}_a)_{i\alpha,k\gamma}(A_a^{(m)})_{k\gamma,r\rho}(\widetilde{D}_a^{\dagger})_{r\rho,p\mu},\\
		\mathcal{T}_R(F_{ab})_{p\mu,q\nu}
		&=\sum_{s\sigma,t\tau}(\widetilde{D}_a)_{p\mu,s\sigma}F_{s\sigma,t\tau}(\widetilde{D}_b^{\dagger})_{t\tau,q\nu},\\
		\mathcal{T}_R(B_b^{(m)})_{q\nu,j\beta}
		&=\sum_{u\upsilon,l\delta}(\widetilde{D}_b)_{q\nu,u\upsilon}(B_b^{(m)})_{u\upsilon,l\delta}(\widetilde{D}_b^{\dagger})_{l\delta,j\beta}.
	\end{align}
	Substituting these expressions into Eq.~(\ref{eq:su2_primitive_contraction}) produces internal contractions of the form
	\begin{equation}
		\sum_{p\mu}(\widetilde{D}_a^{\dagger})_{r\rho,p\mu}(\widetilde{D}_a)_{p\mu,s\sigma}=\delta_{rs}\delta_{\rho\sigma},
		\qquad
		\sum_{q\nu}(\widetilde{D}_b^{\dagger})_{t\tau,q\nu}(\widetilde{D}_b)_{q\nu,u\upsilon}=\delta_{tu}\delta_{\tau\upsilon},
	\end{equation}
	using the unitarity of the orbital--spin representations. All internal representation matrices cancel, leaving
	\begin{align}
		\mathcal M_{SU(2)}(\mathcal T_R F_{ab})_{i\alpha,j\beta}
		&=
		\sum_{k\gamma,l\delta}(\widetilde{D}_a)_{i\alpha,k\gamma}
		\left[\mathcal M_{SU(2)}(F_{ab})\right]_{k\gamma,l\delta}
		(\widetilde{D}_b^{\dagger})_{l\delta,j\beta}.
	\end{align}
	This exactly matches the spinor-tensor transformation rule in Eq.~(\ref{eq:su2_full_block_equivariance}).
\end{proof}

\begin{proof}[Proof of Theorem~\ref{thm:inversion_equivariance}]
For either $\chi\in\{\mathrm{SO(3)},\mathrm{SU(2)}\}$, the
$u\otimes v$ output unit can be written as
\begin{equation}
    \left[\mathcal M_\chi(F_{ab})\right]^{u\otimes v}
    =
    \sum_{m=1}^{M}
    \sum_{u'\in\mathcal B_a}
    \sum_{v'\in\mathcal B_b}
    c_m
    \left[A_a^{(m)}\right]^{u\otimes u'}
    F_{ab}^{u'\otimes v'}
    \left[B_b^{(m)}\right]^{v'\otimes v}.
    \label{eq:proof_unitwise_contraction}
\end{equation}
Because $A_a^{(m)}$, $F_{ab}$, and $B_b^{(m)}$ are inversion
homogeneous, their non-zero units transform according to
Eq.~(\ref{eq:operator_intrinsic_inversion_parity}). One contracted path
therefore acquires the factor
\begin{align}
    &\eta_A^{(m)}(-1)^{l_u+l_{u'}}
    \cdot
    \eta_F(-1)^{l_{u'}+l_{v'}}
    \cdot
    \eta_B^{(m)}(-1)^{l_{v'}+l_v}
    \nonumber\\
    &\qquad
    =
    \eta_A^{(m)}\eta_F\eta_B^{(m)}
    (-1)^{l_u+l_v},
    \label{eq:proof_unitwise_parity_factor}
\end{align}
because the internal orbital parities associated with $u'$ and $v'$ each
appear twice. The spin identity in the SOC setting contributes no
additional inversion sign. Since $c_m$ is inversion invariant, the
$m$-th contraction term is inversion homogeneous with
$\eta_{\mathrm{out}}^{(m)}=\eta_A^{(m)}\eta_F\eta_B^{(m)}$. If these
products are equal for all retained terms, the sum has the same
operator-level intrinsic parity. For an input $F_{ab}$ that obeys the Hamiltonian transformation law
under spatial inversion, $\eta_F=+1$. Requiring
$\eta_A^{(m)}\eta_B^{(m)}=+1$ for every retained term then gives
$\eta_{\mathrm{out}}^{(m)}=+1$. Consequently, each output unit
transforms with the same factor $(-1)^{l_u+l_v}$ as the corresponding
Hamiltonian unit, establishing spatial-inversion equivariance.
\end{proof}

\paragraph{Summary of operator-level intrinsic parities.}
The Hamiltonian, every persistent Hamiltonian-shaped feature, the
attention-derived context, and the identity operator have
\begin{equation}
    \eta_H=\eta_F=\eta_N=\eta_I=+1,
\end{equation}
whereas Eq.~(\ref{eq:directional_unit_inversion_odd}) gives
$\eta_V=-1$. Taking a transpose or Hermitian adjoint does not change the
operator-level intrinsic parity. Consequently, the bilateral
directional action has
$\eta_V\eta_F\eta_V=(-1)(+1)(-1)=+1$, while each one-sided or bilateral
context action contains only factors with intrinsic parity $+1$.
Per-unit gating, residual addition, and LayerScale modulation
use invariant scalars and preserve the same operator-level parity and its
unitwise factor $\eta_Q(-1)^{l_u+l_v}$.

\begin{proof}[Proof of
	Proposition~\ref{prop:canonical_context_equivariance}]
	Let $\mathcal R_v(R)$ denote the local representation associated with
	atom $v$, with $\star=\top$ for the non-SOC $SO(3)$ setting and
	$\star=\dagger$ for the SOC spinor setting. Since all incoming
	operators are expressed in the same carrier space and common local
	basis, they obey
	\[
	F_e(RX)
	=
	\mathcal R_v(R)
	F_e(X)
	\mathcal R_v(R)^\star .
	\]
	Using the rotational invariance of $\alpha_e$, the aggregated context
	therefore transforms as
	\begin{align}
		N_v(RX)
		&=
		\sum_{e:\,\mathrm{dst}(e)=v}
		\alpha_e(RX)F_e(RX)
		\nonumber\\
		&=
		\sum_{e:\,\mathrm{dst}(e)=v}
		\alpha_e(X)
		\mathcal R_v(R)F_e(X)\mathcal R_v(R)^\star
		\nonumber\\
		&=
		\mathcal R_v(R)
		\left[
		\sum_{e:\,\mathrm{dst}(e)=v}
		\alpha_e(X)F_e(X)
		\right]
		\mathcal R_v(R)^\star
		\nonumber\\
		&=
		\mathcal R_v(R)N_v(X)\mathcal R_v(R)^\star .
	\end{align}
	Hence $N_v$ obeys the same proper-rotation law as every incoming
	operator $F_e$.
	
	For spatial inversion, let the incoming operators share the
	operator-level intrinsic parity $\eta_F$. Restricting the aggregation
	to one direct-product unit $u\otimes u'$, with $u,u'\in\mathcal B_v$, gives
	\begin{align}
		N_v^{u\otimes u'}(\mathcal I X)
		&=
		\sum_{e:\,\mathrm{dst}(e)=v}
		\alpha_e(\mathcal I X)
		F_e^{u\otimes u'}(\mathcal I X)
		\nonumber\\
		&=
		\eta_F(-1)^{l_u+l_{u'}}
		\sum_{e:\,\mathrm{dst}(e)=v}
		\alpha_e(X)F_e^{u\otimes u'}(X)
		\nonumber\\
		&=
		\eta_F(-1)^{l_u+l_{u'}}
		N_v^{u\otimes u'}(X).
	\end{align}
	The same $\eta_F$ is therefore shared by every non-zero context unit,
	so $\eta_N=\eta_F$. For the Hamiltonian-shaped feature stream involved
	in the architecture, $\eta_F=+1$ and hence $\eta_N=+1$.
\end{proof}

\section{Architectural Details of the Proposed Neural Network}
\label{app:architecture_details}

\subsection{Delta Representation Learning from $H^{(0)}$ in the Direct-Product Space}
\label{subsec:app_residual_initialization}

The equivariant atom-pair feature $F_{ab}^{(k)}$ passed from one layer to the next is represented in the same direct-product space as the Hamiltonian block for atoms $(a,b)$. Its indices refer to the orbital basis for non-SOC systems,
\begin{equation}
    F_{ab}^{(k)}\in\mathbb{R}^{d_a\times d_b},
    \qquad (F_{ab}^{(k)})_{ij},
\end{equation}
and to the orbital--spin basis for SOC systems,
\begin{equation}
    F_{ab}^{(k)}\in\mathbb{C}^{L_a\times 2\times L_b\times 2},
    \qquad (F_{ab}^{(k)})_{i\alpha,j\beta}.
\end{equation}
We initialize this feature with the zeroth-step Hamiltonian block $H_{ab}^{(0)}$ and accumulate a correction in that basis:
\begin{equation}
    \Delta F_{ab}^{(0)}=0,
    \qquad F_{ab}^{(k)}=H_{ab}^{(0)}+\Delta F_{ab}^{(k)}.
    \label{eq:app_delta_state}
\end{equation}
Alongside these equivariant atom-pair features, the network maintains invariant node and edge features that describe the local environment and determine the coefficients for the equivariant updates. The following subsections describe how the two types of features are initialized and updated within each layer.

Both implementations pad the atom-wise bases to a common $4s2p2d1f$ orbital-set layout, with a consistent ordering of orbital and spin components. Corresponding basis slots thus carry the same rotation and parity representations on every atom. In this layout, each equivariant atom-pair feature is a real $27\times27$ matrix for non-SOC systems or a complex $27\times2\times27\times2$ tensor for SOC systems. The latter is reshaped into a $54\times54$ matrix for contractions. The common matrix dimension is therefore $D=27$ for non-SOC and $D=54$ for SOC. This shared representation allows incoming edge matrices to be aggregated into equivariant atom-wise context features.

\subsection{Dual Conversion between Equivariant and Invariant Features}
\label{subsec:app_invariant_equivariant_routing}

The invariant node and edge features provide the inputs to scalar neural networks that control the equivariant updates. They receive information about pair distances and the current equivariant atom-pair features through quantities that are unchanged by rotations and spatial inversion. We encode the pair distance $r_{ab}=\lVert\mathbf r_{ab}\rVert$ with Gaussian radial basis functions,
\begin{equation}
    \phi_t(r_{ab})=\exp\!\left[-\frac{(r_{ab}-\mu_t)^2}{2\sigma^2}\right],
    \qquad t=1,\ldots,T_{\mathrm{rbf}},
    \label{eq:app_rbf}
\end{equation}
and summarize each direct-product unit of $F_{ab}^{(k)}$ by a normalized Frobenius norm. For non-SOC units, this summary is
\begin{equation}
    n_{ab,uv}^{(k),\mathrm{SO(3)}}=
    \left[
    \frac{1}{d_u^{\mathrm{orb}}d_v^{\mathrm{orb}}}
    \sum_{m_u,m_v}\left|F_{ab,m_um_v}^{u\otimes v,(k)}\right|^2+\varepsilon_n
    \right]^{1/2},
    \label{eq:app_norm_so3}
\end{equation}
and for SOC units, it is
\begin{equation}
    n_{ab,uv}^{(k),\mathrm{SU(2)}}=
    \left[
    \frac{1}{d_ud_v}
    \sum_{m_u,\alpha,m_v,\beta}
    \left|F_{ab,m_u\alpha,m_v\beta}^{u\otimes v,(k)}\right|^2+\varepsilon_n
    \right]^{1/2}.
    \label{eq:app_norm_su2}
\end{equation}
Here $d_u^{\mathrm{orb}}$ is the number of orbital basis functions in set $u$, and $d_u$ also counts the spin components in the SOC setting. Orthogonal or unitary left--right transformations and spatial inversion preserve these norms, so their concatenation is an invariant vector.

For a directed edge $e=(a,b)$, we take $a$ as the source atom, $\mathrm{src}(e)$, and $b$ as the destination atom, $\mathrm{dst}(e)$. We denote the concatenated norms by $\mathbf n_e^{(k)}$ and the radial features by $\boldsymbol\phi_e$. At initialization, an MLP combines the norms of $F_e^{(0)}=H_e^{(0)}$ with the radial features to construct the invariant edge feature. Summing these edge features over the incoming edges of each atom then provides the input for its invariant node feature:
\begin{equation}
    \mathbf z_e^{(0)}=
    \operatorname{MLP}_{\mathrm{init}}\!\left(
    \mathbf n_e^{(0)}\parallel\boldsymbol\phi_e\right),
\end{equation}
\begin{equation}
    \mathbf x_v^{(0)}=
    \operatorname{MLP}_{\mathrm{node\text{-}init}}\!\left(
    \sum_{e:\,\mathrm{dst}(e)=v}\mathbf z_e^{(0)}\right).
    \label{eq:app_node_initialization}
\end{equation}

At layer $k$, the scalar networks receive the current Frobenius summaries, the invariant edge feature, the invariant node features of the source and destination atoms, and the radial features:
\begin{equation}
    \boldsymbol\eta_e^{(k)}=
    \mathbf n_e^{(k)}\parallel\mathbf z_e^{(k)}\parallel
    \mathbf x_{\mathrm{src}(e)}^{(k)}\parallel
    \mathbf x_{\mathrm{dst}(e)}^{(k)}\parallel\boldsymbol\phi_e.
    \label{eq:app_routing_input}
\end{equation}
Here $\parallel$ denotes concatenation. MLPs map $\boldsymbol\eta_e^{(k)}$ to per-unit gates, directional weights, a bilateral directional coefficient, attention logits, and an edge gate. Later in the layer, the updated node features and new Frobenius summaries define a second invariant input for the context coefficients and invariant edge update, as described in Appendix~\ref{subsec:app_attention_context}. The two stages connect the feature types in both directions: norms convey information from equivariant atom-pair features to the scalar networks, and invariant coefficients control the equivariant operations.

The first operation within a layer applies a gate to each direct-product unit of $F_{ab}^{(k)}$, producing $\widetilde F_{ab}^{(k)}$. Each gate is shared across all magnetic components of its row--column orbital-set pair and, for SOC systems, across the spin components as well:
\begin{equation}
    \widetilde F_{ab,m_u,m_v}^{u\otimes v,(k)}=
    \left(1+g_{ab,uv}^{\mathrm{SO(3)}}\right)
    F_{ab,m_u,m_v}^{u\otimes v,(k)},
\end{equation}
\begin{equation}
    \widetilde F_{ab,m_u\alpha,m_v\beta}^{u\otimes v,(k)}=
    \left(1+g_{ab,uv}^{\mathrm{SU(2)}}\right)
    F_{ab,m_u\alpha,m_v\beta}^{u\otimes v,(k)}.
\end{equation}
Using projectors onto the local orbital sets, the same operation can be written as
\begin{equation}
    \widetilde F_{ab}^{(k)}=
    \sum_{u\in\mathcal B_a}\sum_{v\in\mathcal B_b}
    (1+g_{ab,uv})
    P_{a,u}F_{ab}^{(k)}\left(P_{b,v}\right)^\star,
    \label{eq:app_basis_pair_gate_projector}
\end{equation}
For SOC systems, each projector includes the spin identity, $P_{\xi,u}\otimes I_2$. Throughout this appendix, $\star=\top$ for non-SOC matrices and $\star=\dagger$ for SOC tensors. Because the projectors commute with rotations and spatial inversion and the gates are invariant scalars, the scalar networks can adjust the direct-product units independently without changing the transformation laws of the equivariant atom-pair features.

\subsection{Equivariant Updates Using Directions}
\label{subsec:app_directional_lifting}

To incorporate bond orientation, we construct directional operators
acting on the local bases of atoms $a$ and $b$ and apply them to
the gated equivariant atom-pair feature $\widetilde F_{ab}^{(k)}$.
For edge $e=(a,b)$, the directional MLP uses
$\boldsymbol\eta_e^{(k)}$ to predict a residual weight
$\Delta w_e^{\chi,\lambda}$ for each generator channel $\lambda$.
The same residual is used for the corresponding channel on both
atoms in the common basis layout. Substituting the parameterization
$w=1+\Delta w$ into Eq.~(\ref{eq:directional_operator_unified}) gives
\begin{equation}
    \begin{aligned}
        w_{\xi}^{\chi,\lambda}
        &=1+\Delta w_e^{\chi,\lambda},\\
        V_{\xi}^{\chi}(\hat{\mathbf r}_{ab})
        &=\sum_{\lambda\in\mathcal C_{\xi}^{\chi}}
        \sum_{\rho\in\{x,y,z\}}
        \left(1+\Delta w_e^{\chi,\lambda}\right)
        \hat r_{ab,\rho}G_{\xi,\rho}^{\chi,\lambda},
        \qquad \xi\in\{a,b\}.
    \end{aligned}
    \label{eq:app_directional_weight_parameterization}
\end{equation}
Here, $\chi\in\{\mathrm{SO(3)},\mathrm{SU(2)}\}$ and the fixed
generators $G_{\xi,\rho}^{\chi,\lambda}$ are defined in
Appendix~\ref{app:generator}. The operator acts on the orbital
basis in non-SOC systems and on the orbital--spin basis in SOC
systems. For on-site pairs, setting the direction vector to zero
removes the explicit directional contribution; the implementation
uses this convention whenever $r_{ab}\leq10^{-6}$~\AA.

The weights $w_{\xi}^{\chi,\lambda}$ determine the contribution
of each generator channel to the directional operator.
The directional MLP also predicts a separate invariant scalar
coefficient $c_{ab,VHV}^{\chi}$ that scales the bilateral
contraction of these operators with $\widetilde F_{ab}^{(k)}$.
For non-SOC systems,
\begin{align}
    \widehat F_{ab,ij}^{(k)}={}&\widetilde F_{ab,ij}^{(k)}
    +\frac{c_{ab,VHV}^{\mathrm{SO(3)}}}{\sqrt{d_ad_b}}
    \sum_{p,q}(V_a^{\mathrm{SO(3)}})_{ip}
    \widetilde F_{ab,pq}^{(k)}
    [(V_b^{\mathrm{SO(3)}})^\top]_{qj},
    \label{eq:app_direction_so3}
\end{align}
and for SOC systems,
\begin{align}
    \widehat F_{ab,i\alpha,j\beta}^{(k)}={}&
    \widetilde F_{ab,i\alpha,j\beta}^{(k)}
    +\frac{c_{ab,VHV}^{\mathrm{SU(2)}}}{\sqrt{(2L_a)(2L_b)}}
    \sum_{p,\mu,q,\nu}(V_a^{\mathrm{SU(2)}})_{i\alpha,p\mu}
    \widetilde F_{ab,p\mu,q\nu}^{(k)}
    [(V_b^{\mathrm{SU(2)}})^\dagger]_{q\nu,j\beta}.
    \label{eq:app_direction_su2}
\end{align}
With the matrix dimension $D$ defined in Appendix~\ref{subsec:app_residual_initialization}, both expressions take the form
\begin{equation}
    \widehat F_e^{(k)}
    =
    \widetilde F_e^{(k)}
    +
    \frac{c_{e,VHV}}{D}
    V_{\mathrm{src}(e)}^\chi
    \widetilde F_e^{(k)}
    \left(V_{\mathrm{dst}(e)}^\chi\right)^\star,
    \label{eq:app_direction_dense}
\end{equation}
where $\chi\in\{\mathrm{SO(3)},\mathrm{SU(2)}\}$. The use of a bilateral product follows from spatial-inversion symmetry. Each directional operator has operator-level intrinsic parity $\eta_V=-1$, whereas the equivariant atom-pair feature has $\eta_F=+1$. Their product therefore has $\eta_{\mathrm{out}}=\eta_V\eta_F\eta_V=+1$ and follows the Hamiltonian transformation law under inversion. A one-sided directional product would have $\eta_{\mathrm{out}}=-1$ and is excluded from this update.

\subsection{Attention-Based Updates of Node, Edge, and Context Features}
\label{subsec:app_attention_context}
\label{subsec:app_operator_attention}

Attention aggregates the incoming edge information at each atom through two branches. The equivariant branch combines the atom-pair features $\widehat F_e^{(k)}$ obtained after incorporating the bond direction to construct a context feature. The invariant branch combines the current invariant edge features $\mathbf z_e^{(k)}$ to update the node features. Both branches use logits predicted from $\boldsymbol\eta_e^{(k)}$, whose Frobenius summaries are computed from the input feature $F_e^{(k)}$ of the current layer. The attention MLP also predicts an edge gate:
\begin{equation}
    (\ell_e^{\mathrm{op}},b_e,
    \ell_{e,1}^{\mathrm{inv}},\ldots,\ell_{e,H}^{\mathrm{inv}})
    =\operatorname{MLP}_{\mathrm{att}}(\boldsymbol\eta_e^{(k)}).
\end{equation}
The superscripts $\mathrm{op}$ and $\mathrm{inv}$ distinguish the equivariant and invariant branches, respectively, and $H$ is the number of invariant attention heads.

For destination atom $v$, let $\mathcal E_{\to v}$ be its set of incoming edges. The equivariant branch normalizes its logits over this set and uses the resulting invariant weights to construct the atom-wise context:
\begin{equation}
    \alpha_e^{\mathrm{op}}=
    \frac{\exp(\ell_e^{\mathrm{op}}-m_v^{\mathrm{op}})}
    {\sum_{e'\in\mathcal E_{\to v}}
    \exp(\ell_{e'}^{\mathrm{op}}-m_v^{\mathrm{op}})+\varepsilon_\alpha},
    \qquad
    m_v^{\mathrm{op}}=\max_{e'\in\mathcal E_{\to v}}\ell_{e'}^{\mathrm{op}},
\end{equation}
\begin{equation}
    N_v^{(k)}=
    \sum_{e\in\mathcal E_{\to v}}
    \alpha_e^{\mathrm{op}}\widehat F_e^{(k)}.
    \label{eq:app_operator_context}
\end{equation}
The incoming equivariant atom-pair features use the same representation space and common local basis, as required by Proposition~\ref{prop:canonical_context_equivariance}. Their sum with invariant weights is therefore an equivariant context feature $N_v^{(k)}$. It also inherits their intrinsic parity, giving $\eta_N=\eta_F=+1$. Each non-zero direct-product unit consequently satisfies
\begin{equation}
[N_v^{(k)}]^{u\otimes u'}(\mathcal I X)
=\eta_N(-1)^{l_u+l_{u'}}[N_v^{(k)}]^{u\otimes u'}(X)
=(-1)^{l_u+l_{u'}}[N_v^{(k)}]^{u\otimes u'}(X).
\end{equation}

In the invariant branch, each edge feature is projected into $H$ attention heads, $\mathbf v_{e,h}=W_{V,h}\mathbf z_e^{(k)}+\mathbf b_{V,h}$. The weights for each head are normalized over the same set of incoming edges:
\begin{equation}
    \alpha_{e,h}^{\mathrm{inv}}=
    \frac{\exp(\ell_{e,h}^{\mathrm{inv}}-m_{v,h}^{\mathrm{inv}})}
    {\sum_{e'\in\mathcal E_{\to v}}
    \exp(\ell_{e',h}^{\mathrm{inv}}-m_{v,h}^{\mathrm{inv}})+\varepsilon_\alpha},
\end{equation}
Here $m_{v,h}^{\mathrm{inv}}=\max_{e'\in\mathcal E_{\to v}}\ell_{e',h}^{\mathrm{inv}}$. The displayed denominators follow the SOC implementation, which adds $\varepsilon_\alpha=10^{-8}$ to each exponential sum. The non-SOC implementation instead bounds each sum below by $\varepsilon_\alpha$, for both equivariant and invariant attention. Each head aggregates its projected edge features, and their concatenation determines the residual update of the destination-node feature:
\begin{equation}
    \mathbf m_{v,h}^{(k)}=
    \sum_{e\in\mathcal E_{\to v}}
    \alpha_{e,h}^{\mathrm{inv}}\mathbf v_{e,h},
    \qquad
    \mathbf x_v^{(k+1)}=
    \mathbf x_v^{(k)}+
    \gamma_{\mathrm{node}}\operatorname{MLP}_{\mathrm{node}}
    (\operatorname{Concat}_h\mathbf m_{v,h}^{(k)}).
    \label{eq:app_node_attention_update}
\end{equation}

After these aggregations, the edge gate $b_e$ scales $\widehat F_e^{(k)}$ to produce the intermediate equivariant atom-pair feature used in the remaining operations:
\begin{equation}
    \overline F_e^{(k)}=(1+b_e)\widehat F_e^{(k)}.
\end{equation}
We compute new normalized Frobenius summaries from $\overline F_e^{(k)}$ and concatenate them with the updated invariant node features $\mathbf x_{\mathrm{src}(e)}^{(k+1)}$ and $\mathbf x_{\mathrm{dst}(e)}^{(k+1)}$ of the source and destination atoms, the current invariant edge feature $\mathbf z_e^{(k)}$, and the radial features $\boldsymbol\phi_e$. The resulting invariant vector $\boldsymbol\xi_e^{(k)}$ supplies the input for the edge update,
\begin{equation}
    \mathbf z_e^{(k+1)}=\mathbf z_e^{(k)}+
    \gamma_{\mathrm{edge}}\operatorname{MLP}_{\mathrm{feat}}(\boldsymbol\xi_e^{(k)}),
    \label{eq:app_coupled_state_update}
\end{equation}
and for the context-contraction coefficients described next. Thus, the context matrices aggregate $\widehat F_e^{(k)}$, while the subsequent context contractions and their coefficients use $\overline F_e^{(k)}$ and its summaries.

\subsection{Equivariant Updates Using Context Features}
\label{subsec:app_operator_update}

The context features $N_a^{(k)}$ and $N_b^{(k)}$ summarize
the equivariant atom-pair features on edges directed into
atoms $a$ and $b$, respectively. To incorporate this
neighborhood information into the correction, we contract
$\overline F_{ab}^{(k)}$ with $N_a^{(k)}$ on the left,
$(N_b^{(k)})^\star$ on the right, or both. An MLP maps
$\boldsymbol\xi_e^{(k)}$ to three invariant coefficients,
one for each contribution.

Before applying LayerScale, the contribution from context
features for non-SOC systems is
\begin{align}
	\mathcal C_{ab,ij}^{\mathrm{SO(3)},(k)}={}&
	\frac{u_{ab,1}^{\mathrm{SO(3)}}}{\sqrt{d_a}}
	\sum_p
	(N_a^{\mathrm{SO(3)}})_{ip}
	\overline F_{ab,pj}^{(k)}
	\nonumber\\
	&+
	\frac{u_{ab,2}^{\mathrm{SO(3)}}}{\sqrt{d_b}}
	\sum_q
	\overline F_{ab,iq}^{(k)}
	[(N_b^{\mathrm{SO(3)}})^\top]_{qj}
	\nonumber\\
	&+
	\frac{u_{ab,3}^{\mathrm{SO(3)}}}{\sqrt{d_ad_b}}
	\sum_{p,q}
	(N_a^{\mathrm{SO(3)}})_{ip}
	\overline F_{ab,pq}^{(k)}
	[(N_b^{\mathrm{SO(3)}})^\top]_{qj}.
\end{align}
For SOC systems, the corresponding contribution is
\begin{align}
	\mathcal C_{ab,i\alpha,j\beta}^{\mathrm{SU(2)},(k)}={}&
	\frac{u_{ab,1}^{\mathrm{SU(2)}}}{\sqrt{2L_a}}
	\sum_{p,\mu}
	(N_a^{\mathrm{SU(2)}})_{i\alpha,p\mu}
	\overline F_{ab,p\mu,j\beta}^{(k)}
	\nonumber\\
	&+
	\frac{u_{ab,2}^{\mathrm{SU(2)}}}{\sqrt{2L_b}}
	\sum_{q,\nu}
	\overline F_{ab,i\alpha,q\nu}^{(k)}
	[(N_b^{\mathrm{SU(2)}})^\dagger]_{q\nu,j\beta}
	\nonumber\\
	&+
	\frac{u_{ab,3}^{\mathrm{SU(2)}}}{\sqrt{(2L_a)(2L_b)}}
	\sum_{p,\mu,q,\nu}
	(N_a^{\mathrm{SU(2)}})_{i\alpha,p\mu}
	\overline F_{ab,p\mu,q\nu}^{(k)}
	[(N_b^{\mathrm{SU(2)}})^\dagger]_{q\nu,j\beta}.
\end{align}

Multiplying $\mathcal C_{ab}^{\chi,(k)}$ by the LayerScale
parameter $\gamma_H$ gives the equivariant feature correction,
where $\chi\in\{\mathrm{SO(3)},\mathrm{SU(2)}\}$.
For $e=(a,b)$ and the common matrix dimension $D$, the unified
matrix form is
\begin{equation}
	\begin{aligned}
		\delta F_e^{(k)}
		&=
		\gamma_H\mathcal C_{ab}^{\chi,(k)}\\
		&=
		\gamma_H\Biggl[
		\frac{u_{e,1}}{\sqrt D}
		N_{\mathrm{src}(e)}^{(k)}
		\overline F_e^{(k)}
		+
		\frac{u_{e,2}}{\sqrt D}
		\overline F_e^{(k)}
		\left(N_{\mathrm{dst}(e)}^{(k)}\right)^\star\\
		&\qquad\quad+
		\frac{u_{e,3}}{D}
		N_{\mathrm{src}(e)}^{(k)}
		\overline F_e^{(k)}
		\left(N_{\mathrm{dst}(e)}^{(k)}\right)^\star
		\Biggr].
	\end{aligned}
	\label{eq:app_context_dense}
\end{equation}
Here, the orbital and spin indices are combined into matrix
indices in the SOC setting.

The context features and the equivariant atom-pair feature
all have intrinsic parity $+1$. For a one-sided contraction,
the identity operator on the unused side also has intrinsic
parity $+1$. All three contributions therefore satisfy the
parity-product rule in
Theorem~\ref{thm:inversion_equivariance}, so each output unit
has the same spatial-inversion transformation law as the
corresponding Hamiltonian unit.

We accumulate the resulting correction:
\begin{equation}
	\Delta F_e^{(k+1)}
	=
	\Delta F_e^{(k)}+\delta F_e^{(k)}.
	\label{eq:app_equivariant_residual_update}
\end{equation}
The next layer receives
$F_e^{(k+1)}=H_e^{(0)}+\Delta F_e^{(k+1)}$,
together with the updated invariant node and edge features.

\subsection{Stabilization Techniques and $H^{(0)}$-Centered Residual Correction}
\label{subsec:app_norm_stability}
\label{subsec:app_stabilization}

Repeated contractions combine feature magnitudes across basis indices. We control their scale with normalization factors determined by the basis dimension, and regulate the learned updates with LayerNorm and LayerScale. The normalized Frobenius summaries in Eqs.~(\ref{eq:app_norm_so3})--(\ref{eq:app_norm_su2}) use root-mean-square magnitudes so that units with different numbers of components have comparable summaries. Dividing one-sided contractions by $\sqrt D$ and bilateral contractions by $D$ controls growth with the basis dimension. These factors are invariant scalars and preserve equivariance.

LayerNorm is used in the hidden layers of the invariant MLPs:
\begin{equation}
    \mathbf h_1=\operatorname{SiLU}(\operatorname{LN}(W_1\mathbf x+\mathbf b_1)),
    \qquad
    \mathbf h_{\ell+1}=\operatorname{SiLU}(\operatorname{LN}(W_{\ell+1}\mathbf h_\ell+\mathbf b_{\ell+1})),
\end{equation}
Each MLP ends with a linear output without normalization. LayerNorm is not applied to the raw physical inputs or to the equivariant atom-pair features, whose magnitudes remain available to the network. In the large-scale non-SOC implementation, an optional operation limits exceptionally large equivariant feature magnitudes: if the Frobenius norm exceeds a specified threshold, the entire matrix is rescaled by a single invariant scalar detached from gradient computation. This rescaling also preserves equivariance.

Each layer has separate LayerScale parameters
$(\gamma_H,\gamma_{\mathrm{edge}},\gamma_{\mathrm{node}})$ for its
equivariant atom-pair, invariant edge, and invariant node updates.
We initialize these parameters to small values and zero initialize
the final affine layers of the MLPs for gating, directional
operations, attention, context coefficients, and invariant edge
and node updates. At initialization, the resulting weights and
coefficients satisfy
\begin{equation}
    \begin{aligned}
        \Delta w_e^{\chi,\lambda}&=0,
        &w_{\xi}^{\chi,\lambda}&=1,
        &c_{e,VHV}&=0,\\
        g_{e,uv}&=0,
        &b_e&=0,
        &u_{e,j}&=0,\qquad j=1,2,3.
    \end{aligned}
    \label{eq:app_correction_initialization}
\end{equation}
These conditions apply to every edge and layer. The unit weights
retain the fixed generator contributions to $V_a$ and $V_b$ in
Eq.~(\ref{eq:app_directional_weight_parameterization}). Since
$c_{e,VHV}=0$, however, their bilateral contraction makes no change
to the gated feature: Eq.~(\ref{eq:app_direction_dense}) gives
$\widehat F_e^{(k)}=\widetilde F_e^{(k)}$. The zero gate residuals
$g_{e,uv}$ and $b_e$ also leave the features unchanged by gating.
Finally, the three context coefficients $u_{e,j}$ are zero, so
Eq.~(\ref{eq:app_context_dense}) gives $\delta F_e^{(k)}=0$.
Starting from $\Delta F_e^{(0)}=0$, the accumulation in
Eq.~(\ref{eq:app_equivariant_residual_update}) therefore leaves
$\Delta F_e^{(K)}=0$, and the initial prediction equals $H_e^{(0)}$.

After $K$ layers, the output module maps the accumulated feature correction $\Delta F_{ab}^{(K)}$ to the Hamiltonian correction $\Delta H_{ab}^{\mathrm{pred}}$, which is added to the input Hamiltonian block:
\begin{equation}
    H_{ab}^{\mathrm{pred}}
    =
    H_{ab}^{(0)}
    +
    \Delta H_{ab}^{\mathrm{pred}}.
    \label{eq:app_final_output}
\end{equation}
Here $\Delta F_{ab}^{(K)}$ is the correction accumulated within the network, whereas $\Delta H_{ab}^{\mathrm{pred}}$ is the correction used in the final prediction. The prediction remains in the target orbital or orbital--spin basis and follows the Hamiltonian transformation laws under rotations and spatial inversion.
\section{Analysis of Computational Complexity and Hardware Efficiency}
\label{app:complexity_efficiency}

We compare the computational efficiency of our proposed equivariant
learning method in the direct-product space with that of conventional
methods in the direct-sum space, including $SO(3)$ tensor
products~\citep{e3nn} and $SO(2)$ convolutions used in
eSCN~\citep{scmr}, in the context of Hamiltonian prediction. We distinguish two settings: fixed channel
multiplicity at each angular degree, and channel multiplicities
sufficient to represent a complete Hamiltonian block. The latter setting reflects the computational complexity of
updating features that represent complete Hamiltonian blocks
and is therefore directly relevant to Hamiltonian prediction.

We use $T^{\oplus}$ and $T^{\otimes}$ to denote computational costs
for operations in the direct-sum and direct-product spaces,
respectively. The subscript identifies the method or operation.

Unless stated otherwise, operation counts are per directed edge.
For a Hamiltonian block $H_{ab}\in\mathbb F^{D_a\times D_b}$,
$P=D_aD_b$ denotes the number of coefficients. We assume that the local basis dimensions at atoms $a$ and $b$ are comparable, $D_a\asymp D_b\asymp D$, so that
$P=\Theta(D^2)$. Here, $D$ is the dimension of a local orbital or
orbital--spin basis, whereas $L$ is the maximum angular degree of a
direct-sum feature. All counts use conventional dense matrix
multiplication. The spin dimension is fixed and therefore affects
constant factors only. The resulting asymptotic orders apply to the
update rules and basis families specified below.

\subsection{Theoretical Complexity}
\label{subsec:app_complexity_theory}

\paragraph{Encoding features in the direct-sum space with fixed channel multiplicity.}
We begin with a direct-sum feature containing $C$ channels at each
angular degree from $0$ to $L$:
\begin{equation}
    \mathcal X_L
    =\bigoplus_{l=0}^{L}\left(\mathbb F^C\otimes V_l\right),
    \qquad \dim V_l=2l+1.
    \label{eq:app_complexity_fixed_state}
\end{equation}
Its coefficient count is
\begin{equation}
    \dim\mathcal X_L=C\sum_{l=0}^{L}(2l+1)
    =C(L+1)^2=\Theta(CL^2).
    \label{eq:app_complexity_fixed_state_size}
\end{equation}
This setting fixes the multiplicity independently of the angular
cutoff; it does not yet specify a complete Hamiltonian block.

For the $SO(3)$ tensor product in this direct-sum space, the feature
is coupled to spherical harmonics of the edge direction. We retain
all input and output degrees $l_i,l_o\leq L$ and all filter degrees
$|l_i-l_o|\leq l_f\leq l_i+l_o$. These choices give
\begin{equation}
    \sum_{l_i=0}^{L}\sum_{l_o=0}^{L}
    \left[2\min(l_i,l_o)+1\right]=\Theta(L^3)
    \label{eq:app_complexity_number_of_paths}
\end{equation}
angular coupling paths. On each path, we evaluate a dense
Clebsch--Gordan contraction for every input channel, at cost
\begin{equation*}
    \Theta\!\left(C(2l_i+1)(2l_f+1)(2l_o+1)\right).
\end{equation*}
We then apply a $C\times C$ weight matrix to mix the channels for
each of the $2l_o+1$ output magnetic components. This second step
costs $\Theta\!\left(C^2(2l_o+1)\right)$ per path. Summing the
angular contractions and channel mixing over all paths yields
\begin{equation}
    T_{\mathrm{SO(3)}}^{\oplus}(L,C)=\Theta(CL^6+C^2L^4).
    \label{eq:app_complexity_so3_standard}
\end{equation}
Thus, for fixed $C$, the dense angular contractions give a leading
cost of $\Theta(L^6)$~\citep{scmr}.

The $SO(2)$ convolution in the direct-sum space first transforms
the features using Wigner matrices associated with a rotation
that maps the edge direction to a fixed reference axis.
In a complex basis, the convolution then mixes coefficients
with the same signed magnetic order $m$ across angular degrees
and channels. Since each degree $l=|m|,\ldots,L$ contributes
$C$ coefficients, the corresponding block has dimension
$d_m=C(L-|m|+1)$. In a real basis, the components associated
with $m$ and $-m$ are processed jointly, which preserves the
asymptotic complexity. Dense mixing over all orders
$-L\leq m\leq L$ therefore costs
\begin{equation}
	T_{\mathrm{SO(2)},\mathrm{mix}}^{\oplus}(L,C)
	=\Theta\!\left(\sum_{m=-L}^{L}d_m^2\right)
	=\Theta(C^2L^3).
	\label{eq:app_complexity_so2_standard_mixing}
\end{equation}

After mixing, the output is transformed back to the original
coordinate frame using the inverse rotation. Applying the
forward and inverse rotations through dense Wigner matrices costs
\begin{equation}
	T_{\mathrm{rot}}^{\oplus}(L,C)
	=\Theta\!\left(C\sum_{l=0}^{L}(2l+1)^2\right)
	=\Theta(CL^3).
	\label{eq:app_complexity_so2_standard_rotation}
\end{equation}
Combining the mixing and rotation operations gives
\begin{equation}
	T_{\mathrm{SO(2)}}^{\oplus}(L,C)
	=\Theta(C^2L^3+CL^3).
	\label{eq:app_complexity_so2_standard}
\end{equation}
For fixed $C$, the total cost is
$\Theta(L^3)$~\citep{scmr}.

The $\Theta(L^6)$ cost of the $SO(3)$ tensor product and
the $\Theta(L^3)$ cost of the $SO(2)$ convolution therefore
both apply here to direct-sum features with fixed channel
multiplicity and $\Theta(CL^2)$ coefficients. To compare
these operations for features representing a complete
Hamiltonian block, we must first determine how the required
channel multiplicities grow with the orbital basis.

\paragraph{Encoding features in the direct-sum space for representing a complete Hamiltonian block.}
Let $\mathcal B_a$ and $\mathcal B_b$ denote the retained orbital
sets on atoms $a$ and $b$, respectively. Their largest angular degree is
\begin{equation}
    \ell_{\mathrm{orb}}
    =\max\left(\{l_u:u\in\mathcal B_a\}
    \cup\{l_v:v\in\mathcal B_b\}\right)
\end{equation}
The orbital space at atom $a$ is
\begin{equation}
    \mathcal W_a
    =\bigoplus_{l=0}^{\ell_{\mathrm{orb}}}
    \left(\mathbb F^{r_{a,l}}\otimes V_l\right),
    \qquad
    D_a^{\mathrm{orb}}
    =\sum_{l=0}^{\ell_{\mathrm{orb}}}r_{a,l}(2l+1),
    \label{eq:app_complexity_local_basis}
\end{equation}
where $r_{a,l}$ counts the radial copies at degree $l$; the space
at atom $b$ is defined analogously. We study a family of bases in
which $\ell_{\mathrm{orb}}$ increases, radial multiplicities remain
bounded, and each atom retains $\Theta(\ell_{\mathrm{orb}})$
angular degrees. With spin multiplicity $\nu_\chi=1$ for non-SOC
and $\nu_\chi=2$ for SOC, the local basis dimensions are
$D_a=\nu_\chi D_a^{\mathrm{orb}}$ and
$D_b=\nu_\chi D_b^{\mathrm{orb}}$. Under these assumptions,
\begin{equation}
    D_a,D_b=\Theta(\ell_{\mathrm{orb}}^2),
    \qquad P=\Theta(D^2)=\Theta(\ell_{\mathrm{orb}}^4).
    \label{eq:app_complexity_hamiltonian_size}
\end{equation}
This relation determines the conversion from angular-degree scaling
to scaling in the coefficient count $P$.

The space of complete Hamiltonian blocks is
\begin{equation}
    \mathcal H_{ab}
    =\operatorname{Hom}(\mathcal W_b^\chi,\mathcal W_a^\chi),
    \quad
    \mathcal W_a^\chi=\mathcal W_a\otimes\mathbb F^{\nu_\chi},
    \quad \dim\mathcal H_{ab}=P.
    \label{eq:app_complexity_hamiltonian_space}
\end{equation}
Its orbital part decomposes into the direct-sum space as
\begin{equation}
    \operatorname{Hom}(\mathcal W_b,\mathcal W_a)
    \cong\bigoplus_{J=0}^{2\ell_{\mathrm{orb}}}
    \left(\mathbb F^{\mu_{ab,J}}\otimes V_J\right).
    \label{eq:app_complexity_hamiltonian_decomposition}
\end{equation}
The multiplicity at coupled degree $J$ is
\begin{equation}
    \mu_{ab,J}
    =\sum_{l_a,l_b=0}^{\ell_{\mathrm{orb}}}
    r_{a,l_a}r_{b,l_b}
    \mathbf 1\!\left[|l_a-l_b|\leq J\leq l_a+l_b\right],
    \label{eq:app_complexity_hamiltonian_multiplicity}
\end{equation}
because each orbital-set pair contributes one copy of every degree
allowed by the angular-momentum triangle condition. In particular,
the decomposition preserves the full orbital coefficient count:
\begin{equation}
    \sum_{J=0}^{2\ell_{\mathrm{orb}}}\mu_{ab,J}(2J+1)
    =D_a^{\mathrm{orb}}D_b^{\mathrm{orb}}=\Theta(P).
    \label{eq:app_complexity_dimension_identity}
\end{equation}
Each multiplicity is at most
$\mathcal O(\ell_{\mathrm{orb}}^2)$, and
\begin{equation}
    \mu_{ab,J}=\Theta(\ell_{\mathrm{orb}}^2)=\Theta(D)
    \label{eq:app_complexity_multiplicity_scaling}
\end{equation}
holds for $\Theta(\ell_{\mathrm{orb}})$ degrees with
$J=\Theta(\ell_{\mathrm{orb}})$. The fixed spin dimension adds
components in SOC without changing these orders. Hence a direct-sum
representation of the complete Hamiltonian requires multiplicities
that grow with the basis size. The fixed-$C$ estimates above cannot
be applied while keeping $C$ constant in this setting.

For the same dense $SO(3)$ tensor product in the direct-sum space,
replace $C$ on each angular path by the relevant input and output
multiplicities. The angular contraction and channel mixing then cost,
respectively,
\begin{equation*}
    \Theta\!\left(\mu_{ab,J_i}(2J_i+1)(2J_f+1)(2J_o+1)\right)
    \quad\text{and}\quad
    \Theta\!\left(\mu_{ab,J_i}\mu_{ab,J_o}(2J_o+1)\right).
\end{equation*}
Summing over all admissible paths gives
$\ell_{\mathrm{orb}}^8$ operations in each contribution, so
\begin{equation}
    T_{\mathrm{SO(3)}}^{\oplus}(P)
    =\Theta(\ell_{\mathrm{orb}}^8)
    =\Theta(P^2)=\Theta(D^4).
    \label{eq:app_complexity_so3_hamiltonian}
\end{equation}
This is the cost of applying the specified dense direct-sum update
while retaining all coefficients of the Hamiltonian block.

For the $SO(2)$ convolution in the direct-sum space, a magnetic-order
block now includes the multiplicities of every degree $J\geq |m|$:
\begin{equation}
    d_m=\sum_{J=|m|}^{2\ell_{\mathrm{orb}}}\mu_{ab,J}
    \label{eq:app_complexity_so2_hamiltonian_block}
\end{equation}
There are $\Theta(\ell_{\mathrm{orb}})$ such orders with
$d_m=\Theta(\ell_{\mathrm{orb}}^3)$. Dense mixing over all magnetic
orders therefore requires
\begin{equation}
    T_{\mathrm{SO(2)},\mathrm{mix}}^{\oplus}(P)
    =\Theta\!\left(\sum_{m=-2\ell_{\mathrm{orb}}}^{2\ell_{\mathrm{orb}}}
    d_m^2\right)
    =\Theta(P^{7/4})=\Theta(D^{7/2}).
    \label{eq:app_complexity_so2_hamiltonian}
\end{equation}
The accompanying edge-frame rotations cost
\begin{equation}
    T_{\mathrm{rot}}^{\oplus}(P)
    =\Theta\!\left(\sum_{J=0}^{2\ell_{\mathrm{orb}}}
    \mu_{ab,J}(2J+1)^2\right)
    =\Theta(P^{5/4})=\Theta(D^{5/2}).
    \label{eq:app_complexity_so2_frame_rotation}
\end{equation}
Mixing determines the leading total cost,
$T_{\mathrm{SO(2)}}^{\oplus}(P)=\Theta(P^{7/4})$. This result assumes that
every magnetic order is included in the direct-sum convolution.
If mixing is restricted to a fixed range $|m|\leq m_{\max}$, its
cost becomes $\Theta(P^{3/2})$ for the same basis family, with a
corresponding restriction on the couplings performed by each update.

\paragraph{Encoding features in the direct-product space for representing a complete Hamiltonian block.}
DP-Learner retains the same $P$ coefficients as a matrix feature
$F_{ab}^{(k)}\in\mathbb F^{D_a\times D_b}$. Each layer updates this equivariant atom-pair feature using bond
directions and context features through a fixed number of left, right,
and bilateral matrix products, independent of $D$. For operators
$A_a\in\mathbb F^{D_a\times D_a}$ and
$B_b\in\mathbb F^{D_b\times D_b}$ acting on the row and column
bases of $F_{ab}$, respectively, the individual costs are
\begin{align}
    T^{\otimes}(A_aF_{ab})&=\Theta(D_a^2D_b),\nonumber\\
    T^{\otimes}(F_{ab}B_b)&=\Theta(D_aD_b^2),\nonumber\\
    T^{\otimes}(A_aF_{ab}B_b)&=\Theta(D_a^2D_b+D_aD_b^2).
    \label{eq:app_complexity_matrix_products}
\end{align}
For comparable local basis dimensions, $D_a\asymp D_b\asymp D$, these products and their
intermediate matrices give the layer cost and feature storage
\begin{equation}
    T_{\text{DP-Learner}}^{\otimes}(P)=\Theta(P^{3/2})=\Theta(D^3),
    \qquad M_{\mathrm{feature}}=\Theta(P)=\Theta(D^2),
    \label{eq:app_complexity_dp_learner}
\end{equation}
Here, $M_{\mathrm{feature}}$ includes the matrix feature and its
intermediate matrices within one layer. The leading cubic cost
follows from the fixed number of contractions on complete
Hamiltonian matrices.

To account for the remaining operations, let $h$ be the invariant
hidden width and $q=|\mathcal B_a||\mathcal B_b|$ the number of
direct-product units. Per-unit norms, directional-operator
construction, scalar weighting, attention aggregation, and residual
additions require $\mathcal O(P)$ work. With fixed MLP depth,
radial-feature width, and attention-head count, the invariant MLPs
require $\mathcal O(qh+h^2)$ operations per edge and
$\mathcal O(h^2)$ per node. For the basis family above,
$q=\Theta(D)$. Thus, when $h$ is also fixed, these operations
have lower order in $D$ and leave the leading
$\Theta(D^3)$ layer cost unchanged.

Across $K$ layers and $|\mathcal E|$ directed edges, the matrix
contractions require $\Theta(K|\mathcal E|D^3)$ operations, and
the invariant node updates add $\mathcal O(K|\mathcal V|h^2)$.
For a fixed basis and network architecture, a bounded number of
neighbors per atom therefore gives total cost linear in the number
of atoms.

\paragraph{Input and output conversion for methods in the direct-sum space.}
We now count the transformations needed at the input and output stages of a network operating in the direct-sum space. At the input stage, it
takes the complete matrix $H_{ab}^{(0)}$ as input, converting it into  irreducible components. At the output stage, it  returns a complete
Hamiltonian matrix by converting features in the direct-sum space to the direct-product space. These input and output conversions are distinct from the repeated Clebsch--Gordan decompositions of feature tensor products within $SO(3)$ message-passing layers, whose costs are included in the layer-update analysis above.

For an orbital-set pair $(u,v)$, let
$p_{uv}=(2l_u+1)(2l_v+1)$ be the number of coefficients in its
direct-product unit. The magnetic selection rule permits
$\mathcal O(\min(l_u,l_v)+1)$ coupled coefficients per matrix
entry. Evaluating only these permitted positions gives a conversion
cost of
\begin{align}
    T_{\mathrm{conversion,sparse}}(P)
    &=\Theta\!\left(\sum_{u\in\mathcal B_a}\sum_{v\in\mathcal B_b}
    p_{uv}\bigl(\min(l_u,l_v)+1\bigr)\right)\nonumber\\
    &=\Theta(\ell_{\mathrm{orb}}^5)
    =\Theta(P^{5/4})=\Theta(D^{5/2}).
    \label{eq:app_complexity_boundary_conversion}
\end{align}
Alternatively, multiplying by a dense $p_{uv}\times p_{uv}$
coefficient matrix for each orbital-set pair costs
\begin{equation}
    T_{\mathrm{conversion,dense}}(P)
    =\Theta\!\left(\sum_{u\in\mathcal B_a}\sum_{v\in\mathcal B_b}
    p_{uv}^2\right)
    =\Theta(\ell_{\mathrm{orb}}^6)
    =\Theta(P^{3/2})=\Theta(D^3).
    \label{eq:app_complexity_dense_conversion}
\end{equation}
The inverse transformation has the same order. Thus, a complete
prediction requires one input and one output transformation, each
costing up to $\mathcal O(P^{3/2})$ for these implementations.
DP-Learner updates and predicts the Hamiltonian in the target basis,
so neither conversion is needed.

\paragraph{Comparison of computational costs.}
We now compare the three methods when their equivariant features
retain all coefficients of an atom-pair Hamiltonian block. In the
direct-sum space, the channel multiplicities required to retain
this information depend on the orbital basis and grow with the
basis family considered here. The comparison therefore differs
from the setting in which the number of channels at each angular
degree is held fixed. Table~\ref{tab:app_complexity_comparison}
summarizes the results for both settings.

For the basis family in
Eq.~(\ref{eq:app_complexity_hamiltonian_size}), with
$D_a\asymp D_b\asymp D$, the costs of updating these equivariant
features for one directed edge in a single layer are
\begin{equation}
	\begin{aligned}
		T_{\mathrm{SO(3)}}^{\oplus}&=\Theta(D^4),\\
		T_{\mathrm{SO(2)}}^{\oplus}&=\Theta(D^{7/2}),\\
		T_{\text{DP-Learner}}^{\otimes}&=\Theta(D^3).
	\end{aligned}
	\label{eq:app_complexity_ordering}
\end{equation}
Here, the first two costs correspond to $SO(3)$ tensor products
and $SO(2)$ convolutions in the direct-sum space, respectively.

To obtain the cost of a complete prediction from $H_{ab}^{(0)}$
to $H_{ab}^{\mathrm{pred}}$, we account for all $K$ layers.
The direct-sum networks additionally require a conversion of
the input Hamiltonian into direct-sum features and a reconstruction
of the output Hamiltonian from those features. Using
$P=D_aD_b=\Theta(D^2)$, the total costs per directed edge are
\begin{align}
	T_{\mathrm{net},\mathrm{SO(3)}}^{\oplus}(P,K)
	&=\Theta(KP^2)+2T_{\mathrm{conversion}},\nonumber\\
	T_{\mathrm{net},\mathrm{SO(2)}}^{\oplus}(P,K)
	&=\Theta(KP^{7/4})+2T_{\mathrm{conversion}},\nonumber\\
	T_{\mathrm{net},\text{DP-Learner}}^{\otimes}(P,K)
	&=\Theta(KP^{3/2}),
	\label{eq:app_complexity_network_total}
\end{align}
where $T_{\mathrm{conversion}}$ denotes the cost of one forward
or inverse Clebsch--Gordan transformation between a Hamiltonian
block and its direct-sum representation.

The computational advantages of DP-Learner arise from two
aspects. First, under the assumptions above, its matrix
contractions require $\Theta(P^{3/2})$ operations per layer,
compared with $\Theta(P^2)$ for $SO(3)$ tensor products and
$\Theta(P^{7/4})$ for $SO(2)$ convolutions in the direct-sum
space. Second, its equivariant features retain the orbital or
orbital--spin matrix structure throughout the network,
eliminating the input decomposition and output reconstruction
described above. The next subsection explains how these design
choices also enable efficient implementation on hardware
accelerators.

\begin{table}[t]
    \centering
\caption{Comparison of feature sizes and computational costs per
	directed edge. The upper part assumes $C$ channels at each angular
	degree. The lower part compares methods that represent a complete
	Hamiltonian block with $P$ coefficients, using the basis family
	in Eq.~(\ref{eq:app_complexity_hamiltonian_size}).
	The layer-update column reports the cost of one message-passing
	layer. Direct-sum methods additionally convert the input Hamiltonian
	into irreducible features and reconstruct the output Hamiltonian
	after message passing. The last column gives an upper bound on
	the cost of each conversion. DP-Learner operates directly on Hamiltonian
	blocks and requires neither conversion.}
    \label{tab:app_complexity_comparison}
    \resizebox{\textwidth}{!}{%
    \begin{tabular}{@{}llccc@{}}
        \toprule
        Representation & Method & Feature size & Layer-update cost
        & Input/output conversion\\
        \midrule
        \multicolumn{5}{l}{\textit{Fixed channel multiplicity $C$}}\\
        \multirow{2}{*}{Direct-sum}
        & $SO(3)$ tensor product & $\Theta(CL^2)$
        & $\mathcal O(CL^6+C^2L^4)$ & --\\
        & $SO(2)$ convolution (eSCN) & $\Theta(CL^2)$
        & $\mathcal O(C^2L^3+CL^3)$ & --\\
        \midrule
        \multicolumn{5}{l}{\textit{Retaining all coefficients of a complete
        Hamiltonian block: $P=D_aD_b=\Theta(D^2)$}}\\
        \multirow{2}{*}{Direct-sum}
        & $SO(3)$ tensor product & $P$
        & $\Theta(P^2)$ & Up to $\mathcal O(P^{3/2})$\\
        & $SO(2)$ convolution (eSCN) & $P$
        & $\Theta(P^{7/4})$ & Up to $\mathcal O(P^{3/2})$\\
        Direct-product & DP-Learner & $P$
        & $\Theta(P^{3/2})$ & Not required\\
        \bottomrule
    \end{tabular}}
\end{table}

\subsection{Hardware Efficiency}
\label{subsec:app_complexity_hardware}

Both Materials-HAM-SOC~\citep{yin2026advancing} and MP-HAM-110K
use the common $4s2p2d1f$ basis layout described in
Appendix~\ref{subsec:app_residual_initialization}. The local basis dimensions on atoms $a$ and $b$
are therefore equal, $D_a=D_b=D$, with
$D=27$ for non-SOC and $D=54$ for SOC. DP-Learner stores
equivariant atom-pair features in arrays of shape
$[|\mathcal E|,D,D]$ and equivariant atom-wise context features in
arrays of shape $[|\mathcal V|,D,D]$. The storage required for these
features and their intermediate matrices within one layer is
$\mathcal O((|\mathcal E|+|\mathcal V|)P)$, where $P=D^2$.

The common matrix dimensions allow the network to update many
atom-pair features together using optimized dense matrix
multiplication kernels. In particular, the contractions that
incorporate bond directions and context features can be evaluated
in batches without separating the features by angular degree or
magnetic order. The computation of per-unit Frobenius norms,
application of invariant weights, and attention aggregation also
operate directly on the same matrix arrays. The representation thus
supports the full sequence of equivariant feature operations in a
consistent layout.

In comparison, $SO(3)$ tensor products in the direct-sum space
perform contractions along angular coupling paths with different
tensor dimensions and access the corresponding Clebsch--Gordan
coefficients. $SO(2)$ convolutions in the direct-sum space rotate
features into and out of the edge-aligned frame and mix coefficients
in blocks whose sizes depend on the magnetic order. Implementations
that process these paths or blocks separately require repeated
indexing and coefficient or feature reads. Batching tensors with
different shapes may also require padding and masking. These
operations can increase memory traffic and kernel-launch overhead,
while many small contractions can limit data reuse. DP-Learner
organizes its equivariant feature updates as a fixed number of
batched matrix products with common dimensions, allowing these
operations to use the dense computation supported by modern
hardware accelerators.

DP-Learner avoids the computation and memory traffic required
to convert between Hamiltonian blocks and irreducible features.
When a complete Hamiltonian block is processed in the direct-sum
space, its orbital or orbital--spin entries must first be
decomposed into coupled angular components. The predicted
components must then be converted back into Hamiltonian matrix
entries at the output. In addition to the arithmetic costs
derived in Appendix~\ref{subsec:app_complexity_theory}, these
conversions require access to Clebsch--Gordan coefficients,
reorganization of feature indices, and temporary storage for
intermediate results. The number of components grows with the
angular degrees of the orbital basis, increasing this overhead
for the rich basis sets considered here.

DP-Learner retains the orbital or orbital--spin matrix structure
throughout message passing and updates the couplings directly
through matrix contractions. It therefore avoids costly conversions
between direct-product and direct-sum representations. The common
matrix layout also allows features from many atom pairs to be
processed together using optimized dense matrix multiplication
kernels. Avoiding these conversions and performing feature updates
through batched matrix operations contribute to the inference-time
and GPU-memory improvements reported in Table~\ref{tab:main_res}.

\section{Training Loss Function}
\label{app:loss}

DP-Learner learns a Hamiltonian correction so that the prediction
\[
\mathbf{H}^{\mathrm{pred}}(\mathbf{R})
=
\mathbf{H}^{(0)}(\mathbf{R})
+
\Delta\mathbf{H}^{\mathrm{pred}}(\mathbf{R})
\]
approximates the self-consistent ground-truth Hamiltonian. The loss
therefore supervises the correction through the resulting full
Hamiltonian. The output assembly used for evaluation is specified in
Eq.~(\ref{eq:app_final_output}). Here, $\mathbf R$ denotes the lattice
vector connecting the reference unit cell and a neighboring unit cell.

Hamiltonian comparison must account for the arbitrary choice of energy
reference. In a non-orthogonal basis, adding $\mu\mathbf S$ to a
Hamiltonian, where $\mu$ is a real scalar and $\mathbf S$ is the overlap
matrix, shifts all generalized eigenvalues by $\mu$ while preserving
eigenvectors and energy differences. Following the gauge-error
formulation of \citet{wang2024universal}, we use a common shift $\mu$ for
the real-space and reciprocal-space supervision targets and determine
it by minimizing their joint loss.

\paragraph{Real-space supervision.}
The real-space loss compares the predicted Hamiltonian with the
ground-truth Hamiltonian after alignment of the energy reference:
\begin{equation}
	\begin{split}
		\label{loss_real_detail}
		& \mathrm{loss}(\mathbf{R})
		=
		{\mathbb{E}}_{\mathbf{R}}
		\left[
		\lambda_R \cdot \mathrm{loss}_{H}(\mathbf{R})
		\right], \\
		& \mathrm{loss}_{H}(\mathbf{R})
		=
		MSE\big(
		\mathbf{H}^{\mathrm{pred}}(\mathbf{R}),
		\mathbf{H}^{gt}(\mathbf{R},\mu)
		\big),
	\end{split}
\end{equation}
where $MSE$ denotes the mean squared error over Hamiltonian matrix
elements and $\lambda_R$ controls the weight of the real-space term.
The unshifted ground-truth Hamiltonian is the self-consistent result,
\[
\mathbf{H}^{gt}(\mathbf{R})
=
\mathbf{H}^{(T)}(\mathbf{R}).
\]
The aligned supervision target is
\begin{equation}
	\label{H_T_mu}
	\mathbf{H}^{gt}(\mathbf{R},\mu)
	=
	\mathbf{H}^{gt}(\mathbf{R})
	+
	\mu \cdot \mathbf{S}(\mathbf{R}),
\end{equation}
where $\mathbf S(\mathbf R)$ is the real-space overlap matrix.

\paragraph{Reciprocal-space supervision.}
Following NextHAM~\citep{yin2026advancing}, we also supervise the
Hamiltonian in reciprocal space, where band structures and
observables that depend on eigenstates are evaluated. In a
non-orthogonal localized basis, small real-space Hamiltonian errors
can be amplified by the generalized eigenvalue problem involving the
overlap matrix, affecting both eigenvalues and eigenvectors. The
reciprocal-space term addresses these errors by comparing the
Hamiltonians in the ground-truth eigenbasis, with separate weights
for the low-energy subspace, the high-energy subspace, and the
coupling between them.

The predicted reciprocal-space Hamiltonian is obtained by Fourier
transforming the full real-space prediction:
\[
\mathbf{H}^{\mathrm{pred}}(\mathbf{k})
=
\sum_{\mathbf{R}}
e^{i\mathbf{k}\cdot\mathbf{R}}
\mathbf{H}^{\mathrm{pred}}(\mathbf{R}).
\]
The aligned ground-truth Hamiltonian $\mathbf H^{gt}(\mathbf k,\mu)$
is obtained by the same Fourier transform of
$\mathbf H^{gt}(\mathbf R,\mu)$. We represent both Hamiltonians in
the ground-truth eigenbasis
\[
\mathcal{U}(\mathbf{k})
=
[
\,\mathcal{P}(\mathbf{k}),\,
\mathcal{Q}(\mathbf{k})\,
].
\]
The columns of $\mathcal P(\mathbf k)$ and $\mathcal Q(\mathbf k)$
span the low-energy and high-energy subspaces, respectively, with
the energy threshold specified in Appendix~\ref{app:implementation}.
The generalized eigenvectors are normalized with respect to the
overlap matrix:
\[
\mathcal U(\mathbf k)^\dagger
\mathbf S(\mathbf k)
\mathcal U(\mathbf k)=I.
\]
Consequently, adding $\mu\mathbf S(\mathbf k)$ to the ground-truth
Hamiltonian adds $\mu I$ in this basis. The shift changes only the
diagonal entries and leaves the blocks coupling the two subspaces
unchanged. The transformed Hamiltonians are partitioned as
\[
\widetilde{\mathbf{H}}(\mathbf{k})
=
\mathcal{U}(\mathbf{k})^\dagger
\mathbf{H}^{\mathrm{pred}}(\mathbf{k})
\mathcal{U}(\mathbf{k})
=
\begin{bmatrix}
	\widetilde{\mathbf{H}}_{PP}(\mathbf{k}) &
	\widetilde{\mathbf{H}}_{PQ}(\mathbf{k}) \\
	\widetilde{\mathbf{H}}_{QP}(\mathbf{k}) &
	\widetilde{\mathbf{H}}_{QQ}(\mathbf{k})
\end{bmatrix},
\]
\[
\widetilde{\mathbf{H}}^{gt}(\mathbf{k},\mu)
=
\mathcal{U}(\mathbf{k})^\dagger
\mathbf{H}^{gt}(\mathbf{k},\mu)
\mathcal{U}(\mathbf{k})
=
\begin{bmatrix}
	\widetilde{\mathbf{H}}_{PP}^{gt}(\mathbf{k},\mu) &
	\widetilde{\mathbf{H}}_{PQ}^{gt}(\mathbf{k},\mu) \\
	\widetilde{\mathbf{H}}_{QP}^{gt}(\mathbf{k},\mu) &
	\widetilde{\mathbf{H}}_{QQ}^{gt}(\mathbf{k},\mu)
\end{bmatrix}.
\]
Because this basis consists of ground-truth eigenvectors, the
ground-truth block coupling the two subspaces vanishes:
\[
\widetilde{\mathbf{H}}_{PQ}^{gt}(\mathbf{k},\mu)
=
\mathbf{0}.
\]
Prediction errors can produce nonzero entries in
$\widetilde{\mathbf H}_{PQ}(\mathbf k)$, mixing the low-energy and
high-energy subspaces and potentially distorting the resulting
bands. We therefore penalize errors in the two diagonal blocks
and in the coupling block:
\begin{equation}
	\begin{split}
		\label{loss_k_detail}
		\mathrm{loss}(\mathbf{k})
		&=
		\mathbb{E}_{\mathbf{k}}
		\Big[
		\lambda_{\mathcal{P}}
		\cdot
		MSE\big(
		\widetilde{\mathbf{H}}_{PP}(\mathbf{k}),
		\widetilde{\mathbf{H}}_{PP}^{gt}(\mathbf{k},\mu)
		\big)
		\\
		&\quad
		+
		\lambda_{\mathcal{Q}}
		\cdot
		MSE\big(
		\widetilde{\mathbf{H}}_{QQ}(\mathbf{k}),
		\widetilde{\mathbf{H}}_{QQ}^{gt}(\mathbf{k},\mu)
		\big)
		\\
		&\quad
		+
		\lambda_{PQ}
		\cdot
		MSE\big(
		\widetilde{\mathbf{H}}_{PQ}(\mathbf{k}),
		\widetilde{\mathbf{H}}_{PQ}^{gt}(\mathbf{k},\mu)
		\big)
		\Big],
	\end{split}
\end{equation}
The weights $\lambda_{\mathcal P}$, $\lambda_{\mathcal Q}$, and
$\lambda_{PQ}$ control the relative contributions of the three
terms.

\paragraph{Joint loss and optimal energy shift.}
Combining the real-space and reciprocal-space terms gives
\begin{equation}
	\begin{split}
		\label{loss_all}
		\mathrm{loss}_{all}
		=
		&
		\mathrm{loss}(\mathbf{R})
		+
		\mathrm{loss}(\mathbf{k})
		\\
		=
		&
		{\mathbb{E}}_{\mathbf{R}}
		[
		\lambda_R
		\cdot
		\mathrm{loss}_{H}(\mathbf{R})
		]
		\\
		&
		+
		{\mathbb{E}}_{\mathbf{k}}
		[
		\lambda_{\mathcal{P}}
		\cdot
		\mathrm{loss}_{\mathcal{P}}(\mathbf{k})
		+
		\lambda_{\mathcal{Q}}
		\cdot
		\mathrm{loss}_{\mathcal{Q}}(\mathbf{k})
		+
		\lambda_{PQ}
		\cdot
		\mathrm{loss}_{PQ}(\mathbf{k})
		].
	\end{split}
\end{equation}
Here, $\mathrm{loss}_{\mathcal P}(\mathbf k)$,
$\mathrm{loss}_{\mathcal Q}(\mathbf k)$, and
$\mathrm{loss}_{PQ}(\mathbf k)$ refer to the three mean-squared-error
terms in Eq.~(\ref{loss_k_detail}).

Since the supervision targets depend linearly on $\mu$, the joint
loss is quadratic in this scalar. We determine the common energy
shift by the stationarity condition
\[
\frac{\partial \mathrm{loss}_{all}}{\partial \mu}=0.
\]
The $PQ$ term does not contribute to this condition because its
target is unchanged by the energy shift. The real-space term and
the diagonal entries of the $PP$ and $QQ$ blocks give the analytical
solution
\begin{equation}
	\label{getmu}
	\begin{aligned}
		& \mu
		=
		\frac{\Delta_1}{\Delta_2},
		\\
		& \Delta_1
		=
		\frac{\lambda_R}{\text{N}_R}
		\sum_{\mathbf{R},\alpha\beta}
		\text{Re}
		\Bigl(
		\operatorname{conj}\!\left(
		[
		\mathbf{H}^{\mathrm{pred}}(\mathbf{R})
		-
		\mathbf{H}^{gt}(\mathbf{R})
		]_{\alpha\beta}
		\right)
		\mathbf{S}(\mathbf{R})_{\alpha\beta}
		\Bigr)
		\\
		& \qquad\quad
		+
		\frac{\lambda_{\mathcal{P}}}{\text{N}_{\mathcal{P}}}
		\sum_{\mathbf{k},\alpha}
		\text{Re}
		\Bigl(
		[
		\widetilde{\mathbf{H}}_{PP}(\mathbf{k})
		-
		\widetilde{\mathbf{H}}_{PP}^{gt}(\mathbf{k})
		]_{\alpha\alpha}
		\Bigr)
		\\
		& \qquad\quad
		+
		\frac{\lambda_{\mathcal{Q}}}{\text{N}_{\mathcal{Q}}}
		\sum_{\mathbf{k},\alpha}
		\text{Re}
		\Bigl(
		[
		\widetilde{\mathbf{H}}_{QQ}(\mathbf{k})
		-
		\widetilde{\mathbf{H}}_{QQ}^{gt}(\mathbf{k})
		]_{\alpha\alpha}
		\Bigr),
		\\
		& \Delta_2
		=
		\frac{\lambda_R}{\text{N}_R}
		\sum_{\mathbf{R},\alpha\beta}
		\operatorname{conj}\!\left(\mathbf{S}(\mathbf{R})_{\alpha\beta}\right)
		\mathbf{S}(\mathbf{R})_{\alpha\beta}
		+
		\sum_{\mathbf{k},\alpha}
		\frac{\lambda_{\mathcal{P}}}{\text{N}_{\mathcal{P}}}
		+
		\sum_{\mathbf{k},\alpha}
		\frac{\lambda_{\mathcal{Q}}}{\text{N}_{\mathcal{Q}}},
	\end{aligned}
\end{equation}
where $\operatorname{conj}(\cdot)$ denotes scalar or entrywise complex
conjugation. The quantities $\text{N}_R$, $\text{N}_{\mathcal P}$,
and $\text{N}_{\mathcal Q}$ are the total numbers of Hamiltonian matrix
elements in the real-space, $PP$, and $QQ$ losses, respectively.
The ground-truth reciprocal-space blocks written without $\mu$ in
Eq.~(\ref{getmu}) are the unshifted blocks. Substituting this value of
$\mu$ into both supervision targets aligns their energy reference
according to the joint objective.

\section{Dataset Details}
\label{data_detail}

To construct our MP-HAM-110K dataset, first-principles calculations are performed with the localized numerical
atomic-orbital implementation of ABACUS~\citep{lipengfei_2016,Linpz2023}. We
use the Perdew--Burke--Ernzerhof (PBE) exchange--correlation
functional~\citep{pbe_1996} and scalar-relativistic optimized norm-conserving
Vanderbilt (ONCV) pseudopotentials~\citep{ONCV_2013} from the PseudoDojo
library~\citep{dojo_2018}. The self-consistent calculations employ
a wave-function cutoff of 100~Ry, a $\Gamma$-centered $k$-point mesh with a
reciprocal-space spacing of
$0.08~\mathrm{Bohr}^{-1}$, Gaussian smearing of $10^{-3}$~Ry, and a
charge-density convergence threshold of $10^{-7}$. All calculations are spin-unpolarized and exclude spin--orbit coupling. For each structure, the
zeroth-step Hamiltonian $H^{(0)}$, the converged Hamiltonian $H^{(T)}$, and the
overlap matrix $S$ are generated, and PYATB~\citep{jin2023pyatb} is used for
band-structure post-processing. The element-resolved pseudopotential valence
configurations and numerical atomic-orbital (NAO) basis
sets~\citep{lin2021strategy} are summarized in
Table~\ref{tab:nsoc_pseudo_orbital}. The dataset additionally provides the wavefunction-related information required for constructing the reciprocal-space loss. For each training
sample, the ground-truth reciprocal-space Hamiltonian and overlap matrix at the sampled $\mathbf{k}$ point are used to solve the generalized Kohn--Sham eigenvalue problem, yielding the
ground-truth eigenvalues and eigenvectors. The resulting Kohn--Sham eigenvectors define the ground-truth eigenstate basis, which is partitioned into low- and high-energy subspaces and subsequently used to project the predicted and ground-truth Hamiltonians for constructing the
$\mathbf{k}$-space loss.

\input{nsoc_pseudopotential_orbital_table_longtable_fullwidth.tex}
\FloatBarrier

\begin{figure}
	\centering
	\begin{subfigure}{\linewidth}
		\centering
		\includegraphics[width=0.98\linewidth]{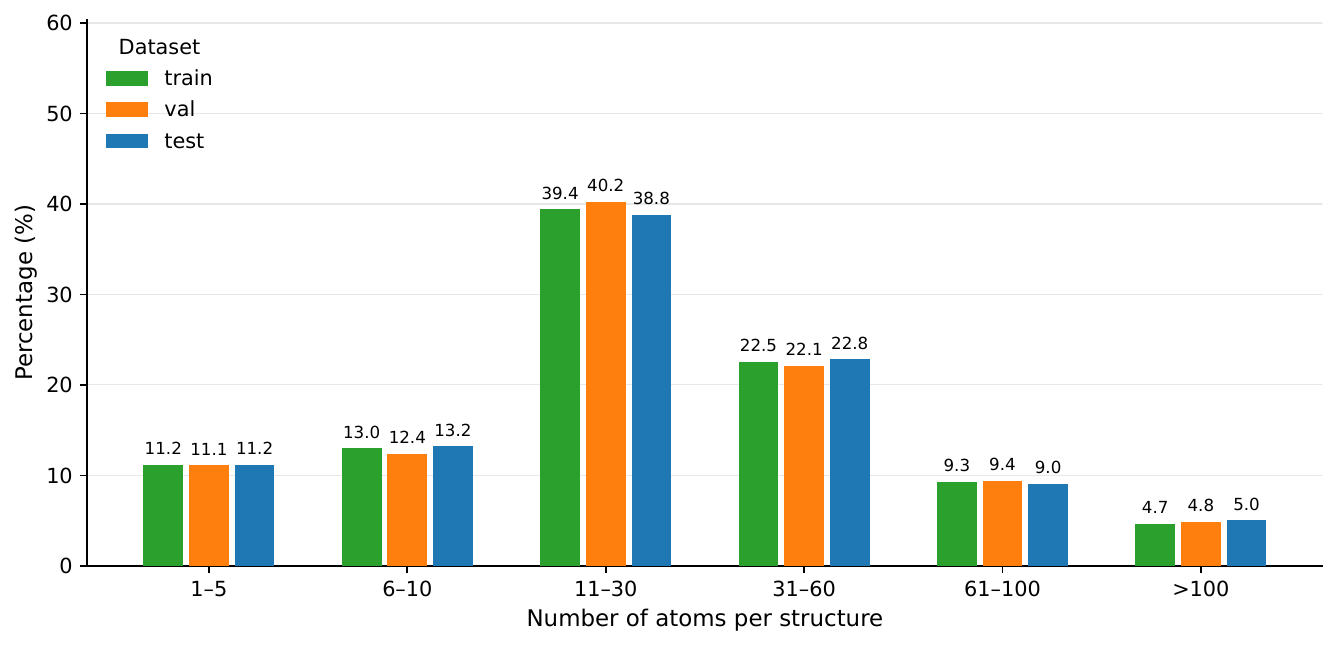}
		\caption{Distribution of the number of atoms in the supplied unit cell.}
		\label{fig:mp_ham_atom_count_distribution}
	\end{subfigure}
	\medskip
	\begin{subfigure}{\linewidth}
		\centering
		\includegraphics[width=0.98\linewidth]{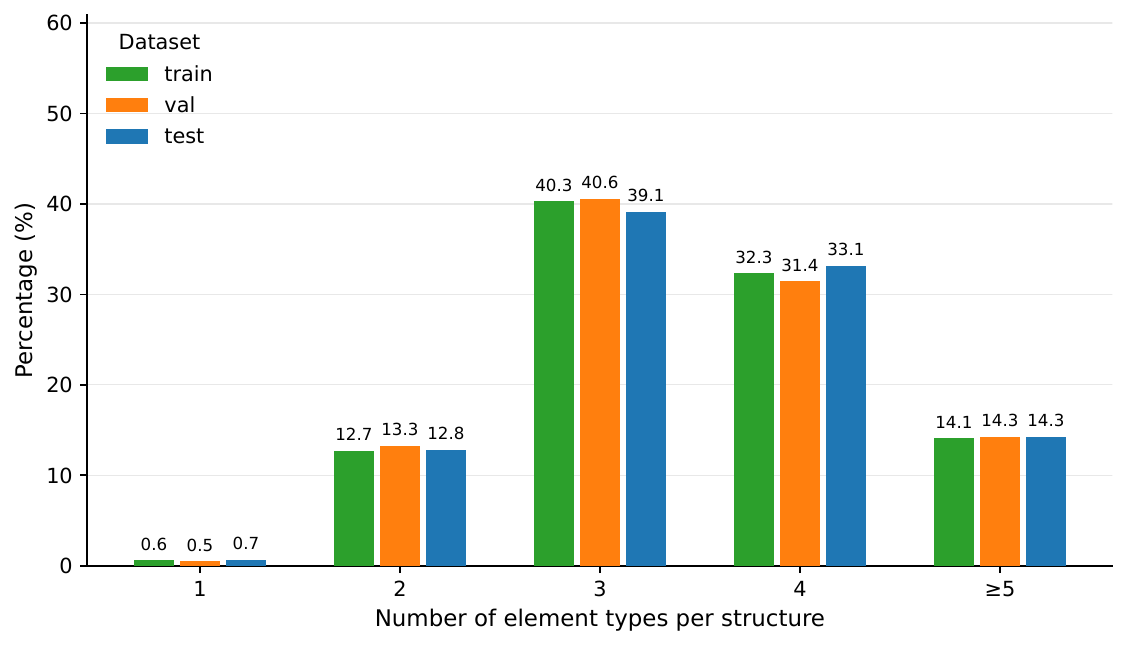}
		\caption{Distribution of the number of distinct element types per structure.}
		\label{fig:mp_ham_element_diversity_distribution}
	\end{subfigure}
	\caption{Distributions of atomic counts and numbers of element types in the
		training, validation, and testing sets.}
	\label{fig:mp_ham_dataset_distributions}
\end{figure}

\begin{figure}
	\centering
	\begin{subfigure}{\linewidth}
		\centering
		\includegraphics[width=\linewidth]{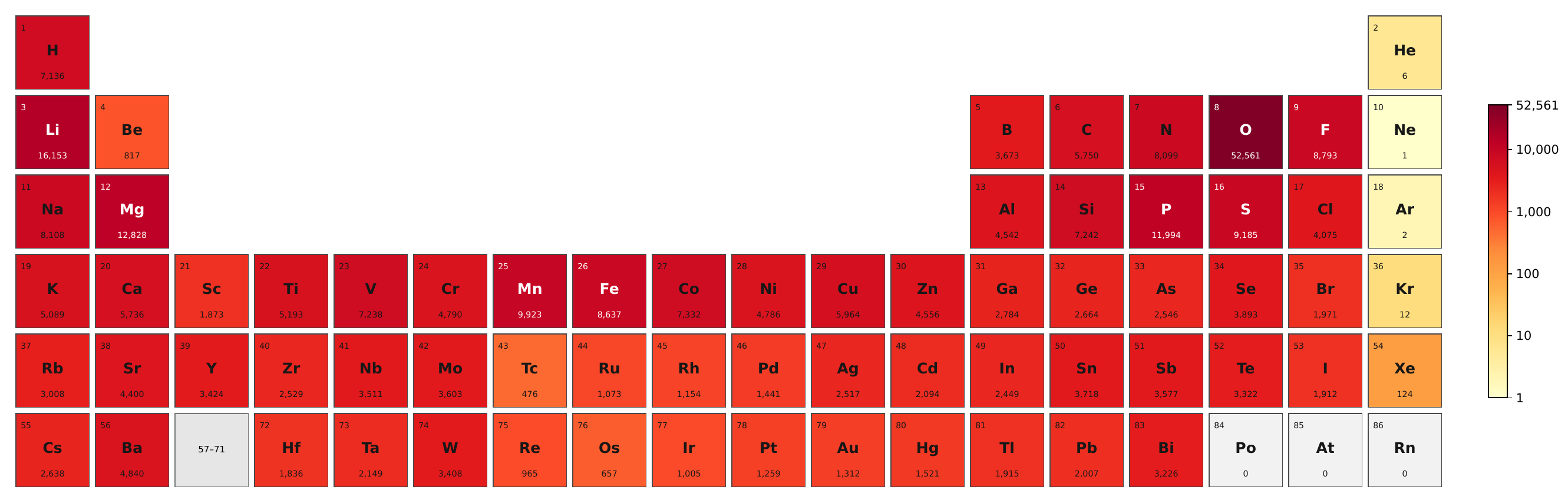}
		\caption{Training set ($91{,}040$ structures).}
		\label{fig:mp_ham_element_occurrences_train}
	\end{subfigure}
	\begin{subfigure}{\linewidth}
		\centering
		\includegraphics[width=\linewidth]{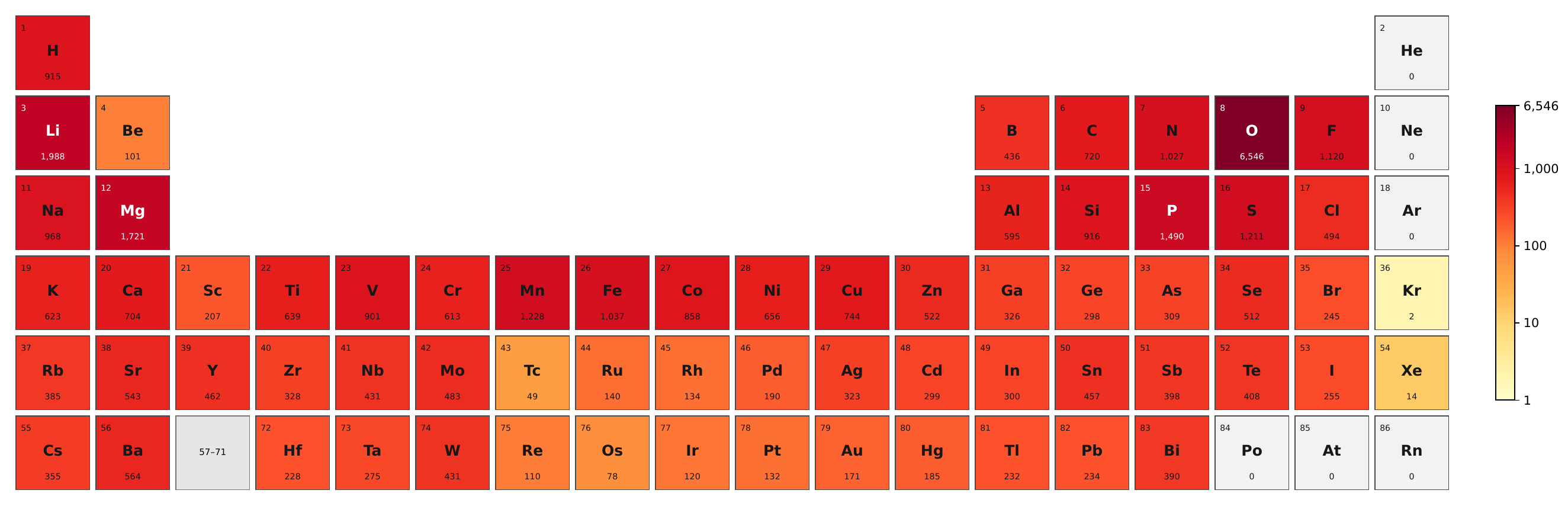}
		\caption{Validation set ($11{,}380$ structures).}
		\label{fig:mp_ham_element_occurrences_val}
	\end{subfigure}
	\begin{subfigure}{\linewidth}
		\centering
		\includegraphics[width=\linewidth]{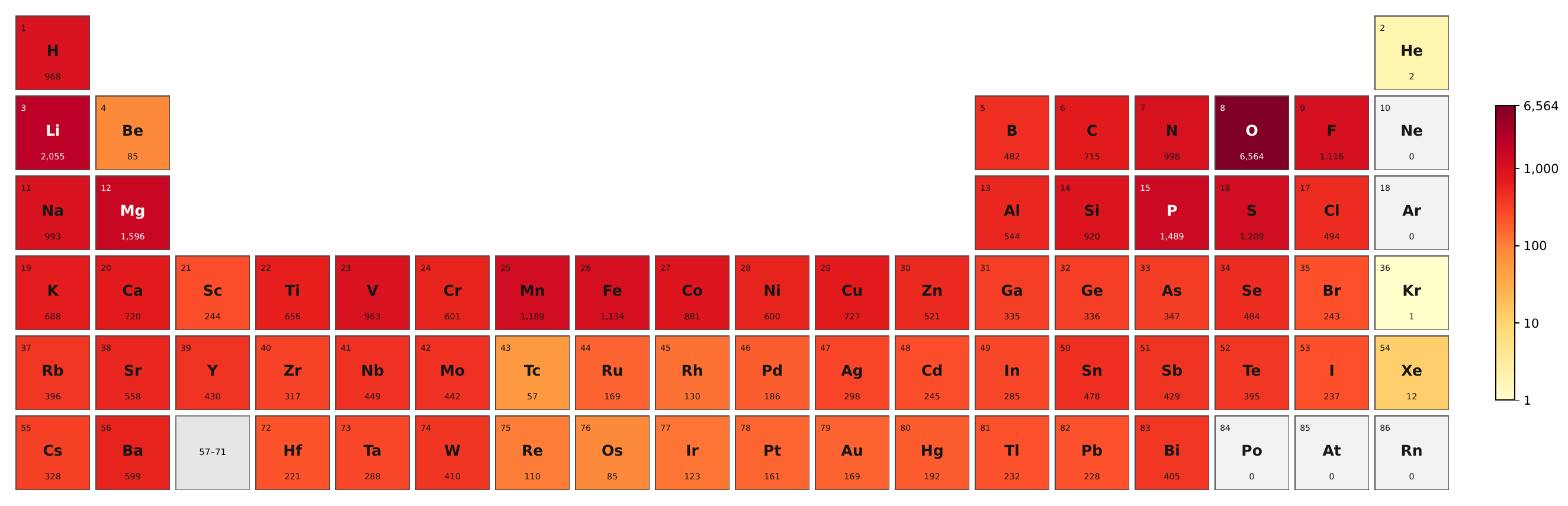}
		\caption{Testing set ($11{,}380$ structures).}
		\label{fig:mp_ham_element_occurrences_test}
	\end{subfigure}
	\caption{Element-occurrence frequencies in the training, validation, and
		testing sets.}
	\label{fig:mp_ham_element_occurrences}
\end{figure}

\begin{figure}
	\centering    
	\includegraphics[scale=0.32]{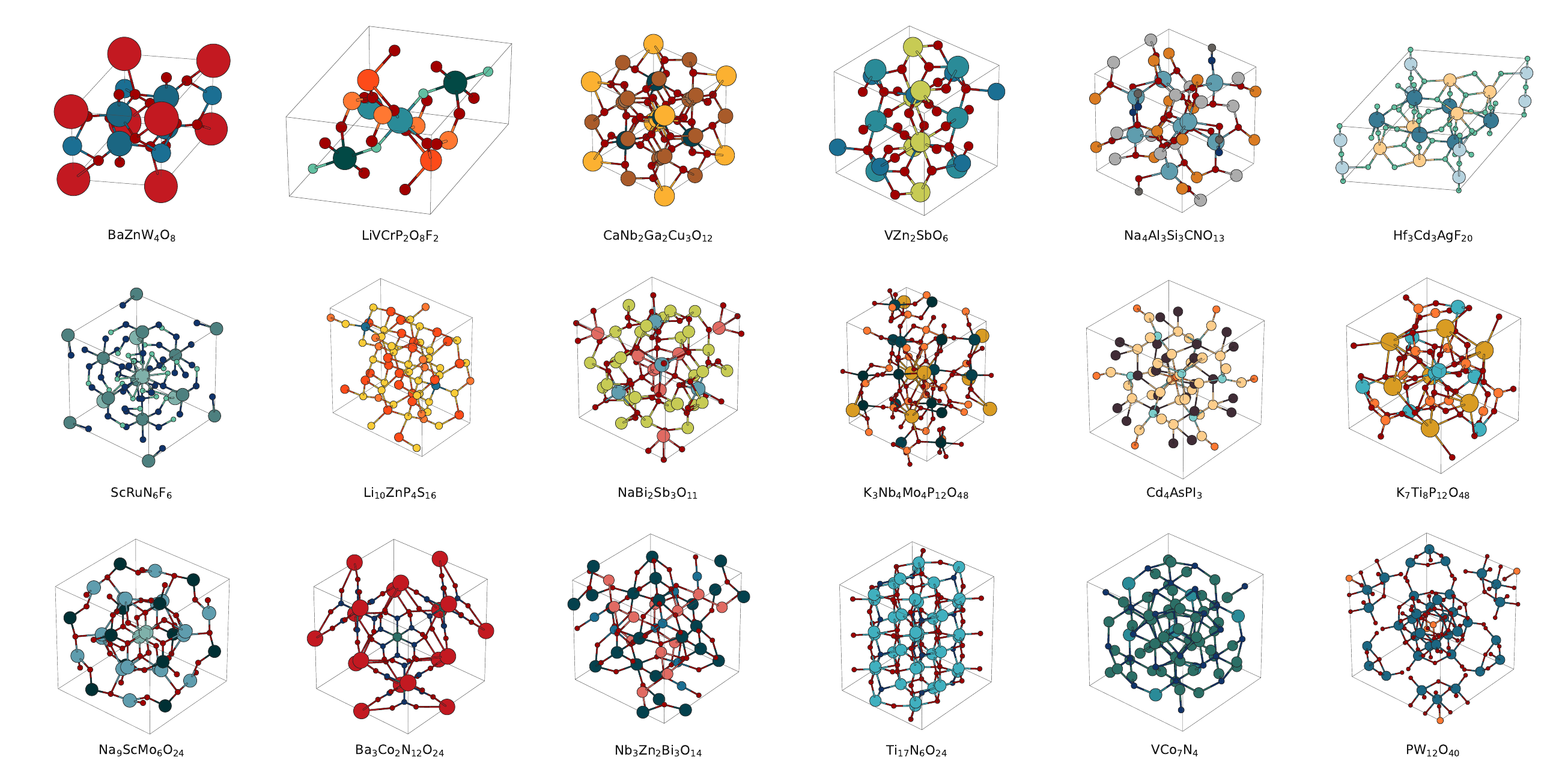}
	\caption{Sampled material structures from our MP-HAM-110K dataset.}
	\label{fig:mp_ham_crystal_gallery}
\end{figure}

The crystal structure files were obtained from the Materials Project
database~\citep{jain2013commentary}, yielding $113{,}800$ periodic structures.
The dataset is randomly divided into $91{,}040$ training, $11{,}380$ validation, and
$11{,}380$ testing structures, corresponding to an 8:1:1 split. It covers 68
elements, comprising H--Ba and Hf--Bi. The unit cells contain 1--169 atoms,
with a mean of 31.62 atoms per structure, and 1--9 distinct element types.
Figure~\ref{fig:mp_ham_crystal_gallery} shows 18 representative structures
ordered from 14 to 106 atoms, illustrating the diversity of their chemical
compositions and structural complexity. The atomic-count and element-diversity
distributions in Figure~\ref{fig:mp_ham_dataset_distributions} show that most
structures contain 11--60 atoms and three or four element types, with closely
matched profiles across the three splits. Figure~\ref{fig:mp_ham_element_occurrences}
presents the corresponding element-occurrence frequencies; all 68 elements
occur in the training set, while 65 and 66 occur in the validation and test
sets, respectively.

\section{Implementation Details}
\label{app:implementation}

\paragraph{Computational environment and baseline reproduction.}
All experiments use PyTorch 2.2.0 with CUDA 12.1 on cloud servers
equipped with NVIDIA A800 GPUs, each with 80~GiB of memory, same as that of NextHAM.
The NextHAM results in Table~\ref{tab:main_res} are obtained
using its official code \footnote{
	\url{https://github.com/DavidYin94/NextHAM}}.
For Materials-HAM-SOC, we evaluate the released pretrained
models and obtain results consistent with the original study.
For MP-HAM-110K, we train and evaluate NextHAM from scratch.
Because the released code targets SOC systems, we make only
the adaptations required for the non-SOC dataset, primarily
adjusting the input feature dimensions.

The original NextHAM framework uses four sub-models to predict
Hamiltonian corrections for different atom-pair distance
intervals~\citep{yin2026advancing}. DP-Learner instead uses a
single model to predict corrections across the full distance
range. In our experiments, this model provides sufficient
expressive capacity to capture the corrections across
different distances, without partitioning the prediction
task among separate models. We also observe that using
separate DP-Learner models for individual distance intervals
increases the tendency to overfit. Each NextHAM sub-model receives the entire structure as
input but predicts corrections only for atom pairs within
its assigned interval. The four outputs are assembled
according to these intervals to form the complete Hamiltonian
prediction. Our comparison therefore includes all four
NextHAM sub-models and the single DP-Learner model, so that
both methods perform the complete prediction task. With the official implementation, running all four NextHAM
sub-models on a single GPU causes out-of-memory errors for
some structures during both training and inference. We
therefore retain its original configuration, distributing
the four sub-models across four GPUs. DP-Learner uses one
GPU for both training and inference. The reported parameter
counts, GPU memory consumption, and inference times account
for the complete predictor of each method.

\paragraph{Graph construction and feature representations.}
We construct the atomic graph using a  cutoff distance of
$8.0$~\AA\ to define the Hamiltonian blocks of atom pairs.
Both the non-SOC and SOC models use $K=12$ layers, with
independent parameters in each layer. The invariant node and
edge features each have dimension $256$. Attention for
aggregating invariant features uses $4$ heads with $64$
channels per head, while a separate scalar attention weight
controls the aggregation of equivariant features. Pair
distances are expanded using $T_{\mathrm{rbf}}=128$ Gaussian
basis functions with fixed centers evenly spaced over
$[0,8.0]$~\AA. The standard deviation in
Eq.~(\ref{eq:app_rbf}) is
$\sigma=0.5\times8.0/128=0.03125$~\AA.

Both models use a common layout of nine local orbital sets,
with angular degrees $(0,0,0,0,1,1,2,2,3)$, corresponding to
four $s$ sets, two $p$ sets, two $d$ sets, and one $f$ set.
For the non-SOC model, the resulting $27$ orbital basis
functions give real $27\times27$ equivariant atom-pair features
in the direct-product space. The directional MLP predicts one
orbital weight for each local orbital set, and all gates and
contraction coefficients are real.

For the SOC model, each orbital has two spin components.
The equivariant atom-pair features therefore have shape
$27\times2\times27\times2$ and are reshaped into complex
$54\times54$ matrices for contractions. The implementation
permutes the external spin-major layout
$2\times27\times2\times27$ into this orbital-major layout
before the contractions and reverses the permutation at
the output. The directional MLP
predicts nine orbital weights and nine spin weights separately.
The per-unit gates, directional weights, and contraction
coefficients are complex scalars formed from pairs of real
MLP outputs. The invariant node and edge features and the
attention logits remain real. In both models, the equivariant
feature update using bond directions includes only the
bilateral directional contraction defined in
Appendix~\ref{subsec:app_directional_lifting}.

\paragraph{MLP parameterization.}
The $9\times9=81$ direct-product units provide $81$ normalized
Frobenius summaries in both models. Concatenating these
summaries with the radial features gives the MLP for initial
edge features an input dimension of $81+128=209$.
Within each layer, the MLPs for per-unit gating, directional
embedding, attention, context-contraction coefficients, and
invariant edge updates receive inputs comprising the
Frobenius summaries, the radial features, and the three
$256$-dimensional feature vectors associated with the edge,
its source atom, and its destination atom. Their input dimension is therefore
$81+128+3\times256=977$. The MLPs for initial node features
and invariant node updates each receive a $256$-dimensional
input.

The MLPs for per-unit gating, directional embedding, and
attention each have three hidden layers. The MLPs for initial
node and edge features, context-contraction coefficients,
and invariant node and edge updates each have two hidden
layers. Every hidden layer has width $256$ and applies an
affine transformation, LayerNorm, and SiLU in that order.
Each MLP ends with an affine output layer without normalization
or activation. The MLP for context-contraction coefficients
produces three real scalars in the non-SOC model and three
complex scalars in the SOC model.

The non-SOC and SOC configurations contain $26{,}093{,}805$
and $26{,}445{,}381$ parameters, respectively, counting all
registered parameter tensors.

\paragraph{Time-reversal symmetry.}

For SOC systems, we follow NextHAM~\citep{yin2026advancing}
in applying corrections to the real parts of the two
spin-conserving blocks. Its analysis on Materials-HAM-SOC
shows that the discrepancies between $H^{(0)}$ and the
self-consistent Hamiltonian are concentrated in these
components, while the errors in the spin-flip blocks and
the imaginary parts of the spin-conserving blocks are
negligible. We therefore retain these remaining components
from $H^{(0)}$. During output post-processing, we average
the real parts of the predicted corrections to the
$\uparrow\uparrow$ and $\downarrow\downarrow$ blocks and
add the resulting real matrix to both spin-conserving
blocks of $H^{(0)}$. This procedure preserves the
time-reversal symmetry of 
Hamiltonians.

\paragraph{Initialization and normalization.}
Each layer has three trainable LayerScale parameters,
initialized to $0.01$, for updating the equivariant features,
invariant edge features, and invariant node features.
The weights and biases of the final affine layers in the
MLPs for per-unit gating, directional embedding, attention,
context-contraction coefficients, and invariant node and
edge updates are initialized to zero.

With the parameterization in
Eq.~(\ref{eq:app_directional_weight_parameterization}), the
directional weights start at one. The gate residuals,
the coefficient $c_{e,VHV}$ for the bilateral directional
contraction, and the three coefficients $u_{e,j}$ for
contractions using context features start at zero, as summarized
in Eq.~(\ref{eq:app_correction_initialization}). The resulting
initial prediction is $H^{(0)}$, following the update equations
in Appendix~\ref{subsec:app_norm_stability}.

The normalized Frobenius summaries use
$\varepsilon_n=10^{-5}$ inside the square root.
Each matrix multiplication used to update equivariant
features with directional or contextual information is
divided by $\sqrt{D}$, where $D=27$ for non-SOC and $D=54$
for SOC. A bilateral contraction involving two matrix
multiplications therefore has an overall normalization
factor of $D^{-1}$.

\paragraph{Training protocol.}
For the reciprocal-space loss, ground-truth eigenstates with
energies at or below $E_F+10$~eV define the low-energy subspace
$\mathcal P$, and the remaining states define the high-energy
subspace $\mathcal Q$, following
NextHAM~\citep{yin2026advancing}. For the joint loss in
Eq.~(\ref{loss_all}), we use the same weights as NextHAM:
$\lambda_R=0.99955$, $\lambda_{\mathcal P}=0.0002$,
$\lambda_{\mathcal Q}=0.0001$, and $\lambda_{PQ}=0.00015$.
These weights control the real-space Hamiltonian loss and the
reciprocal-space losses for the $\mathcal P$, $\mathcal Q$,
and $PQ$ blocks, respectively, as defined in
Appendix~\ref{app:loss}.

We train DP-Learner for $100$ epochs with Adam and select
the checkpoint with the best  performance on the validation set for
testing. The learning rate increases linearly from $10^{-6}$
to $5\times10^{-4}$ during the first five epochs, then follows
a cosine decay to $10^{-5}$ over the remaining $95$ epochs.
On each dataset, we train the full model and each of the five
architectural ablation variants from scratch with three random
seeds ($2025$, $2026$, and $2027$) and report the mean results
over the three runs. For the full models and the ablation studies, the standard deviation of Gauge MAE
across runs is below $0.005$~meV.

\paragraph{Accuracy evaluation.}
Although mean absolute error (MAE) provides a straightforward
measure of prediction accuracy, Hamiltonians must be compared
with a common energy reference. In a non-orthogonal
atomic-orbital basis, adding $\mu\mathbf{S}$ to the Hamiltonian,
where $\mu$ is a real scalar and $\mathbf{S}$ is the overlap
matrix, shifts all eigenvalues equally while preserving
eigenvectors and energy differences. An unaligned MAE can
therefore penalize differences arising solely from the
arbitrary choice of energy reference.

To account for this gauge freedom, we evaluate Hamiltonian
prediction accuracy using Gauge MAE, following the evaluation
protocol of NextHAM~\citep{yin2026advancing} and the gauge-error
formulation of \citet{wang2024universal}. For each test
structure, a common energy-reference offset is removed before
computing the mean absolute error over entries selected by
the evaluation mask. For SOC systems, the metric uses the
complex modulus of each entrywise error and includes all four
spin blocks. Results are averaged over test structures and
reported in meV. The full model, the architectural ablations,
and the Only $H^{(0)}$ reference use the same evaluation
procedure.

\paragraph{Efficiency evaluation.}
We report the mean peak GPU memory per test structure using
PyTorch's reserved-memory metric. The measurement includes
resident model parameters and input tensors, together with the
additional memory required to compute the prediction. To account for the complete predictor, we sum the peak GPU memory
values across all four GPUs used by the complete NextHAM predictor. For
DP-Learner, which uses a single model on one GPU, we record the
peak memory on that device.

We also measure the elapsed wall-clock time for transferring model
inputs, executing the forward pass, converting the outputs into a
DFT-compatible Hamiltonian format, and saving the predicted
Hamiltonians to disk. Since both methods share the same pre-processing pipeline, including the
construction of the zeroth-step Hamiltonian $H^{(0)}$, the
pre-processing cost is not included, allowing a direct comparison of
model-dependent inference costs. The reported time
is averaged over test structures. Concurrent model computations are
counted once within this elapsed interval; runtimes across GPUs are
not summed.

\section{Additional Experimental Results}
\label{app:ablation}

We compare DP-Learner with an Only $H^{(0)}$ reference and
five architectural ablation variants on both
Materials-HAM-SOC~\citep{yin2026advancing} and MP-HAM-110K,
using the SOC and non-SOC formulations, respectively. The
Only $H^{(0)}$ reference measures the improvement achieved
by learning a correction to the zeroth-step Hamiltonian.
The five architectural ablations assess the individual
contributions of the following components to
Hamiltonian prediction accuracy: updating equivariant features
across layers, incorporating bond directions through Lie
generators, aggregating equivariant features from neighboring
atom pairs, predicting invariant coefficients from the local
environment, and scaling residual feature updates with LayerScale.
Each architectural ablation starts from the corresponding full
architecture in Appendix~\ref{app:architecture_details} and
makes the changes described below. Within each dataset, all
comparisons use the same testing structures and evaluation
procedure. The five trained variants also use the same split,
training objective, optimizer, learning-rate schedule, and
training duration as the full model.
The radial basis, cutoff radius, invariant hidden width, MLP
depth, and attention configuration also follow the corresponding
full model wherever these components are retained.

The settings below apply to both formulations, with
$\chi\in\{\mathrm{SO(3)},\mathrm{SU(2)}\}$.
We use $\chi=\mathrm{SO(3)}$, $\star=\top$, and $D=27$ for
the non-SOC model on MP-HAM-110K, and
$\chi=\mathrm{SU(2)}$, $\star=\dagger$, and $D=54$ for the
SOC model on Materials-HAM-SOC. The feature symbols follow
the corresponding formulation; we omit $\chi$ where the
same expression applies to both.

The reference and ablation settings are described below.

\paragraph{Only $H^{(0)}$.}
This reference uses the zeroth-step Hamiltonian directly as
the prediction, $H_{ab}^{\mathrm{pred}}=H_{ab}^{(0)}$, without
applying a neural network correction. It has no trainable
parameters and requires no training. We evaluate it on both
testing sets using the same Gauge MAE procedure as the learned
models. This comparison quantifies the accuracy gained by
learning the correction from the initial Hamiltonian to the
self-consistent reference.

\paragraph{Without equivariant feature updates across layers.}
In the full model, invariant node and edge features determine the
coefficients used to update equivariant atom-pair features in the
direct-product space. This variant removes those equivariant feature
updates and retains message passing through invariant node and edge
features. In particular, it removes the intermediate features
$F_e^{(k)}$, their per-unit and edge gating, the construction of
directional operators, and the attention that aggregates equivariant
features into atom-wise context features. The matrix contractions
that update equivariant features using directional information,
\begin{equation}
    V_{\mathrm{src}(e)}^{\chi}
    \widetilde F_e^{(k)}
    \left(V_{\mathrm{dst}(e)}^{\chi}\right)^\star
    \label{eq:ablation_removed_directional_actions}
\end{equation}
and contextual information,
\begin{equation}
    N_{\mathrm{src}(e)}^{(k)}\overline F_e^{(k)},
    \qquad
    \overline F_e^{(k)}
    \left(N_{\mathrm{dst}(e)}^{(k)}\right)^\star,
    \qquad
    N_{\mathrm{src}(e)}^{(k)}\overline F_e^{(k)}
    \left(N_{\mathrm{dst}(e)}^{(k)}\right)^\star
    \label{eq:ablation_removed_context_actions}
\end{equation}
are therefore absent from every layer. The retained invariant
networks use fixed summaries of $H^{(0)}$ throughout message passing
and predict scalar coefficients that modify $H^{(0)}$ at the output.

The initial invariant features are constructed in the same way as in
the full model. Let $\mathbf n_e^{(0)}$ contain the normalized
Frobenius norms of the direct-product units of $H_e^{(0)}$, computed
using Eq.~(\ref{eq:app_norm_so3}) for non-SOC and
Eq.~(\ref{eq:app_norm_su2}) for SOC. The initial edge and node
features are
\begin{equation}
    \mathbf z_e^{(0)}
    =
    \operatorname{MLP}_{\mathrm{init}}
    \left(
    \mathbf n_e^{(0)}
    \parallel
    \boldsymbol\phi_e
    \right),
    \label{eq:ablation_invariant_edge_init}
\end{equation}
\begin{equation}
    \mathbf x_v^{(0)}
    =
    \operatorname{MLP}_{\mathrm{node\text{-}init}}
    \left(
    \sum_{e:\,\mathrm{dst}(e)=v}
    \mathbf z_e^{(0)}
    \right).
    \label{eq:ablation_invariant_node_init}
\end{equation}
Since no intermediate equivariant features are updated, each layer
combines the fixed summaries $\mathbf n_e^{(0)}$ with the current
invariant features and radial information:
\begin{equation}
    \boldsymbol\eta_{e,\mathrm{inv}}^{(k)}
    =
    \mathbf n_e^{(0)}
    \parallel
    \mathbf z_e^{(k)}
    \parallel
    \mathbf x_{\mathrm{src}(e)}^{(k)}
    \parallel
    \mathbf x_{\mathrm{dst}(e)}^{(k)}
    \parallel
    \boldsymbol\phi_e.
    \label{eq:ablation_invariant_only_input}
\end{equation}
The scalar attention logits are predicted from
$\boldsymbol\eta_{e,\mathrm{inv}}^{(k)}$, and the attention value
vectors are obtained from $\mathbf z_e^{(k)}$. Multi-head attention
then updates the invariant node features according to
Eq.~(\ref{eq:app_node_attention_update}). The updated node features
enter the input for the invariant edge update,
\begin{equation}
    \boldsymbol\xi_{e,\mathrm{inv}}^{(k)}
    =
    \mathbf n_e^{(0)}
    \parallel
    \mathbf z_e^{(k)}
    \parallel
    \mathbf x_{\mathrm{src}(e)}^{(k+1)}
    \parallel
    \mathbf x_{\mathrm{dst}(e)}^{(k+1)}
    \parallel
    \boldsymbol\phi_e,
    \label{eq:ablation_invariant_post_attention}
\end{equation}
which gives
\begin{equation}
    \mathbf z_e^{(k+1)}
    =
    \mathbf z_e^{(k)}
    +
    \gamma_{\mathrm{edge}}
    \operatorname{MLP}_{\mathrm{feat}}
    \left(
    \boldsymbol\xi_{e,\mathrm{inv}}^{(k)}
    \right).
    \label{eq:ablation_invariant_edge_update}
\end{equation}
Thus, the invariant node and edge networks continue to learn
non-linear functions of the neighboring environment, while the
Hamiltonian summaries remain fixed across layers.

An output MLP maps the final invariant features to one scalar
coefficient for each direct-product unit of $H^{(0)}$. This readout
allows the variant to learn a Hamiltonian correction after the
equivariant feature updates have been removed. Its input is
\begin{equation}
    \boldsymbol\zeta_{ab}
    =
    \mathbf n_{ab}^{(0)}
    \parallel
    \mathbf z_{ab}^{(K)}
    \parallel
    \mathbf x_a^{(K)}
    \parallel
    \mathbf x_b^{(K)}
    \parallel
    \boldsymbol\phi_{ab},
    \label{eq:ablation_invariant_readout_input}
\end{equation}
and its outputs are
\begin{equation}
    \left\{
    \beta_{ab,uv}^{\chi}
    \right\}_{u\in\mathcal B_a,\;v\in\mathcal B_b}
    =
    \operatorname{MLP}_{\mathrm{out}}
    \left(
    \boldsymbol\zeta_{ab}
    \right).
    \label{eq:ablation_invariant_beta}
\end{equation}
The coefficients $\beta_{ab,uv}^{\chi}$ are real for non-SOC
and complex for SOC; in the latter case, each complex
coefficient is formed from two real MLP outputs.
To apply these coefficients to the corresponding orbital sets, define
\begin{equation}
    \Pi_{\xi,u}^{\chi}
    =
    \begin{cases}
        P_{\xi,u}, & \chi=\mathrm{SO(3)},\\
        P_{\xi,u}\otimes I_2, & \chi=\mathrm{SU(2)}.
    \end{cases}
    \label{eq:ablation_projector_chi}
\end{equation}
where $P_{\xi,u}$ projects onto the complete magnetic subspace of
orbital set $u$ on atom $\xi$. For SOC, the tensor product
with $I_2$ includes both spin components. The correction before
the final output processing is
\begin{equation}
    \Delta H_{ab}^{\mathrm{inv},\chi}
    =
    \sum_{u\in\mathcal B_a}
    \sum_{v\in\mathcal B_b}
    \beta_{ab,uv}^{\chi}
    \Pi_{a,u}^{\chi}
    H_{ab}^{(0)}
    \left(\Pi_{b,v}^{\chi}\right)^\star.
    \label{eq:ablation_invariant_readout}
\end{equation}
The same output module as in the full model maps
$\Delta H_{ab}^{\mathrm{inv},\chi}$ to the final Hamiltonian
correction $\Delta H_{ab}^{\mathrm{pred}}$. The prediction is
\begin{equation}
    H_{ab}^{\mathrm{pred}}
    =
    H_{ab}^{(0)}
    +
    \Delta H_{ab}^{\mathrm{pred}}.
    \label{eq:ablation_invariant_output}
\end{equation}
The output assembly therefore follows
Eq.~(\ref{eq:app_final_output}), while this variant does not
propagate equivariant features across layers.
The final affine layer of $\operatorname{MLP}_{\mathrm{out}}$ is
zero initialized, so that $\beta_{ab,uv}^{\chi}=0$ and the initial
prediction equals $H_{ab}^{(0)}$.

The coefficients $\beta_{ab,uv}^{\chi}$ are invariant, and the
projectors commute with the rotation and spatial-inversion
representations. The correction therefore follows the same transformation
law as the Hamiltonian. Before the common output processing,
all entries within each direct-product unit are multiplied
by the same scalar. This multiplication
can adjust the magnitude and, for complex coefficients, the
phase of each unit according to the environment, but it does
not mix its row or column components. This variant compares learning
equivariant matrix features across layers with using invariant
message passing to modulate the fixed units of $H^{(0)}$.

\paragraph{Without directional embedding.}
In this variant, we remove the directional operators in
Eq.~(\ref{eq:directional_operator_unified}) and their bilateral
contraction with the equivariant atom-pair features in
Eq.~(\ref{eq:app_direction_dense}). The gated feature therefore passes directly to the attention
calculation and aggregation of equivariant context features:
\begin{equation}
    \widehat F_e^{(k)}
    =
    \widetilde F_e^{(k)},
    \qquad
    \forall\,e,k.
    \label{eq:ablation_no_directional_lifting}
\end{equation}
This modification removes the orbital contribution to the
directional embedding in the non-SOC model and both the orbital
and spin contributions in the SOC model.

Per-unit gating, aggregation of equivariant atom-wise context features,
and updating equivariant atom-pair features using those context
features remain active. The invariant node and edge networks and the
stabilization mechanisms are also retained. Given that $H^{(0)}$ already
depends on the orbital orientation, this comparison measures the
additional contribution of using bond directions explicitly through
the Lie-generator-based embedding.

\paragraph{Without aggregation of equivariant context features.}
This variant removes the mechanism that aggregates incoming
equivariant atom-pair features into the atom-wise context
$N_v^{(k)}$ in Eq.~(\ref{eq:app_operator_context}). The associated
logits $\ell_e^{\mathrm{op}}$ and weights
$\alpha_e^{\mathrm{op}}$ are removed, and the context feature is
replaced by the identity matrix at every layer:
\begin{equation}
    N_v^{(k)}
    =
    I_D,
    \qquad
    \forall\,v,k.
    \label{eq:ablation_identity_context}
\end{equation}
The local bases of the source and destination atoms are padded
to the same dimension $D$. Substituting
$N_v^{(k)}=I_D$ into Eq.~(\ref{eq:app_context_dense}) gives
\begin{equation}
    \delta F_e^{(k)}
    =
    \gamma_H
    \left(
    \frac{u_{e,1}+u_{e,2}}{\sqrt D}
    +
    \frac{u_{e,3}}{D}
    \right)
    \overline F_e^{(k)}.
    \label{eq:ablation_context_local_update}
\end{equation}
The left, right, and bilateral matrix contractions thus reduce to
scalar multiplication of the current equivariant atom-pair feature.
The coefficients $u_{e,1}$, $u_{e,2}$, and $u_{e,3}$ are still
predicted from the invariant features of the local environment.

Updating equivariant features using directional information, scalar
attention, invariant node and edge updates, and all stabilization
mechanisms are retained. Neighboring atoms can therefore still
influence the predicted scalar coefficients through invariant message
passing. What is removed is the aggregation of neighboring
equivariant matrix features and their subsequent action on the
orbital or orbital--spin components of the current atom-pair feature.

\paragraph{Without environment-dependent invariant coefficients.}
In the full model, MLPs use the local environment to predict invariant
coefficients for updating equivariant features and logits for
attention. This ablation term replaces these MLPs with directly learnable
parameters of the same output shapes. Each layer has its own set of
parameters, shared across all atom pairs. These parameters are
optimized during training but do not vary with the input environment.
The per-unit gates, directional-weight residuals, coefficient for the
bilateral contraction using directional operators, edge gate, and
coefficients for contractions using context features become
\begin{align}
    g_{e,uv}^{(k)}&=\overline g_{uv}^{(k)},
    &\Delta w_e^{(k)}&=\overline{\Delta w}^{(k)},
    &c_{e,VHV}^{(k)}&=\overline c_{VHV}^{(k)},\nonumber\\
    b_e^{(k)}&=\overline b^{(k)},
    &u_{e,j}^{(k)}&=\overline u_j^{(k)},\quad j=1,2,3,
    &\forall\,e.
    \label{eq:ablation_static_routing}
\end{align}
Here, the overbars denote the directly learned parameters. Gates
remain distinct across direct-product units, and directional weights
remain distinct across generator channels. These coefficients
retain the real or complex type used in the corresponding full
model, while attention logits remain real. The parameters are
initialized to match the outputs of the corresponding heads in
the full model:
\begin{equation}
    \overline g_{uv}^{(k)}=0,
    \qquad
    \overline{\Delta w}^{(k)}=0,
    \qquad
    \overline c_{VHV}^{(k)}=0,
    \qquad
    \overline b^{(k)}=0,
    \qquad
    \overline u_j^{(k)}=0\quad(j=1,2,3).
    \label{eq:ablation_static_routing_initialization}
\end{equation}
The directional weights therefore still start from
$w=1+\Delta w=1$, and the initial Hamiltonian correction is zero.

The attention logits for aggregating equivariant features and for
each scalar attention head are also replaced by parameters shared
across edges within each layer:
\begin{equation}
    \ell_e^{\mathrm{op},(k)}=\overline\ell^{\mathrm{op},(k)},
    \qquad
    \ell_{e,h}^{\mathrm{inv},(k)}
    =\overline\ell_h^{\mathrm{inv},(k)},
    \qquad \forall\,e.
\end{equation}
The softmax then assigns equal weights to the incoming edges of
each destination atom:
\begin{equation}
    \alpha_e^{\mathrm{op}}
    =
    \alpha_{e,h}^{\mathrm{inv}}
    =
    \begin{cases}
        |\mathcal E_{\to v}|^{-1},
        & \chi=\mathrm{SO(3)},\\
        (|\mathcal E_{\to v}|+\varepsilon_\alpha)^{-1},
        & \chi=\mathrm{SU(2)},
    \end{cases}
    \qquad e\in\mathcal E_{\to v}.
    \label{eq:ablation_uniform_attention}
\end{equation}
These expressions follow the denominator stabilization used
in the corresponding full models, as described in
Appendix~\ref{subsec:app_attention_context}. Both assign equal
weights within each nonempty incoming-edge set; the SOC weights
include the stabilization constant $\varepsilon_\alpha$ and
are approximately $|\mathcal E_{\to v}|^{-1}$.

All matrix contractions that update equivariant features using
directional and contextual information are retained. The invariant
node and edge networks also remain in the architecture, but their
input-dependent features no longer determine the coefficients or
attention weights used in the equivariant computation. The
Hamiltonian prediction still depends on the input through $H^{(0)}$,
the bond directions, and the aggregated equivariant features. This
comparison assesses the benefit of adapting both the coefficients for
updating equivariant features and the attention weights to the local
environment.

\clearpage
\begin{table}[p]
	\centering
	\caption{Results of ablation study on the Materials-HAM-SOC and MP-HAM-110K
		testing sets, using the SOC and non-SOC formulations, respectively.
		Gauge MAE is the gauge-corrected Hamiltonian mean absolute error
		in meV. The ablation term labeled as \textit{Only} $H^{(0)}$ reports the error obtained
		by directly using the zeroth-step Hamiltonian as the prediction,
		without any learned correction. Results for the full model and
		the five other  ablation terms are averaged over three
		independent training runs. Lower values indicate better performance.}
	\label{tab:ablation_dp_learner}
	\small
	\setlength{\tabcolsep}{5.0pt}
	\renewcommand{\arraystretch}{1.08}
	\begin{tabular}{@{}p{0.50\textwidth}cc@{}}
		\toprule
		\textbf{Method} & \multicolumn{2}{c}{\textbf{Gauge MAE (meV) $\downarrow$}}\\
		\cmidrule(l){2-3}
		& \textbf{Materials-HAM-SOC} & \textbf{MP-HAM-110K}\\
		\midrule
		Only $H^{(0)}$ & 2.606 & 5.203\\
		Without equivariant feature updates across layers & 1.875 & 3.421\\
		Without directional embedding & 1.025 & 2.302\\
		Without aggregation of equivariant context features & 1.354 & 2.433\\
		Without environment-dependent invariant coefficients & 1.280 & 2.230\\
		Without LayerScale & 0.963 & 2.065\\
		\textbf{DP-Learner (full)} & \textbf{0.841} & \textbf{1.595}\\
		\bottomrule
	\end{tabular}
\end{table}

\begin{figure}[p]
	\centering
	\includegraphics[scale=0.7]{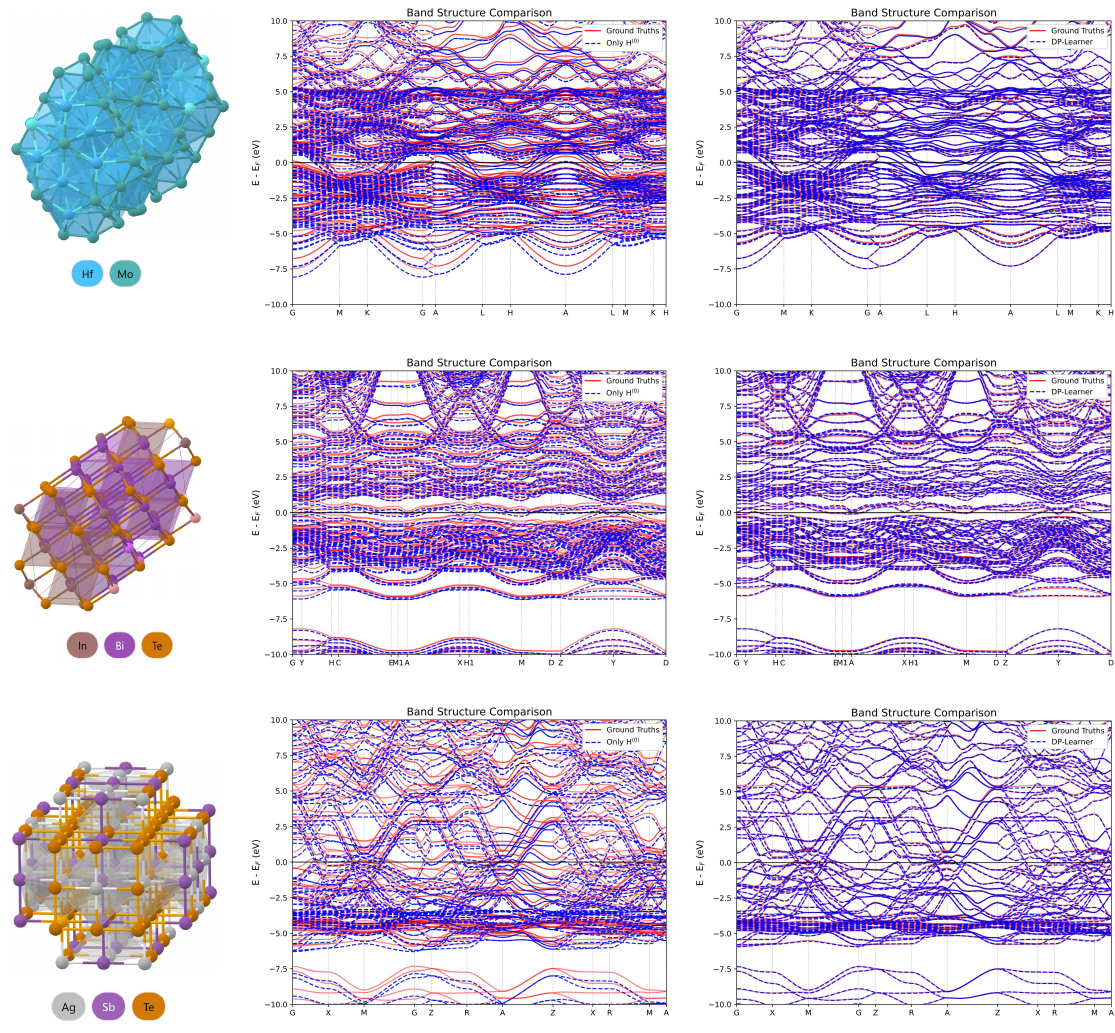}
	\caption{Band structures obtained using only $H^{(0)}$ and
		using DP-Learner for three representative structures from the MP-HAM-110K dataset. Each row
		shows the crystal structure (left), the bands obtained
		from $H^{(0)}$ (middle), and the DP-Learner bands (right).
		Red solid lines show the DFT reference bands used as ground truths in our dataset, while blue dashed lines show the results of the corresponding method. Energies are referenced to the Fermi energy $E_F$.}
	\label{fig:band_h0_compare}
\end{figure}

\paragraph{Without LayerScale.}
In the full model, LayerScale regulates the magnitudes of the residual
feature updates through trainable parameters initialized to $0.01$.
This ablation term fixes the residual scaling factors for the equivariant
atom-pair features and the invariant edge and node features to one.
In every layer, the effective multipliers in
Eqs.~(\ref{eq:app_context_dense}),
(\ref{eq:app_coupled_state_update}), and
(\ref{eq:app_node_attention_update}) satisfy
\begin{equation}
    \gamma_H
    =
    \gamma_{\mathrm{edge}}
    =
    \gamma_{\mathrm{node}}
    =
    1.
    \label{eq:ablation_no_layerscale}
\end{equation}
The residual feature updates are therefore added without learnable
LayerScale multipliers. Any bounding function applied to these
multipliers is bypassed so that their effective values are exactly
one.

All feature representations, attention mechanisms, and matrix
contractions are retained. The contraction factors $D^{-1/2}$ and
$D^{-1}$, zero initialization of the correction heads,
$w=1+\Delta w$ parameterization, and other stabilization mechanisms
are unchanged. This variant evaluates the contribution of LayerScale
to controlling the magnitudes of the three types of residual feature
updates.

Table~\ref{tab:ablation_dp_learner} reports the mean Gauge MAE
on both testing sets. The full model achieves $0.841$~meV on
Materials-HAM-SOC and $1.595$~meV on MP-HAM-110K, improving
on both the Only $H^{(0)}$ reference and every architectural
ablation. Removing any of the five components increases
the error in both the SOC and non-SOC formulations.

Using only $H^{(0)}$ gives Gauge MAEs of $2.606$~meV on
Materials-HAM-SOC and $5.203$~meV on MP-HAM-110K. DP-Learner
reduces these errors to $0.841$ and $1.595$~meV, respectively,
corresponding to absolute decreases of $1.765$ and $3.608$~meV. These reductions
show that the learned corrections substantially improve on
the initial electronic-structure estimates in both settings.
The zeroth-step Hamiltonian provides a physically informed
starting point, while the network uses the local geometric
and electronic environment to refine its orbital or
orbital--spin couplings toward the self-consistent reference.
Figure~\ref{fig:band_h0_compare} shows that the bands obtained
from $H^{(0)}$ exhibit substantial deviations from the DFT
reference, whereas DP-Learner closely reproduces the reference
band structures, demonstrating the effectiveness of the learned
Hamiltonian corrections.

Removing equivariant feature updates across layers produces the
largest error among the trained variants on both datasets:
$1.875$~meV on Materials-HAM-SOC
and $3.421$~meV on MP-HAM-110K. These correspond to increases of
$122.9\%$ and $114.5\%$ over the respective full models.
These results demonstrate the importance of learning equivariant
features across layers for both orbital and orbital--spin
Hamiltonian prediction. The variant still learns non-linear
invariant node and edge features, but its correction before
the final output processing is restricted to scaling the
direct-product units of $H^{(0)}$. The full model uses
directional operators and equivariant context features to mix
the row and column components of atom-pair features during
message passing. These matrix contractions allow the network
to represent more general changes in individual orbital or
orbital--spin couplings induced by the local geometric and
electronic environment, providing a more expressive
representation for predicting the self-consistent Hamiltonian.

Removing the directional embedding increases Gauge MAE to
$1.025$~meV on Materials-HAM-SOC and $2.302$~meV on MP-HAM-110K,
corresponding to increases of $21.9\%$ and $44.3\%$, respectively.
These increases indicate that explicitly incorporating bond
directions helps the network model the orientation dependence
of orbital and orbital--spin couplings more accurately, even
though $H^{(0)}$ already contains some orientation information.
The directional embedding contracts bond directions with
orbital Lie generators and, for SOC, spin Lie generators to
construct operators that act on the row and column components
of atom-pair features. With weights learned from the local
environment, these operators allow bond orientation to guide
feature updates at every layer while preserving equivariance.

Removing the aggregation of equivariant context features
increases Gauge MAE to $1.354$~meV on Materials-HAM-SOC and
$2.433$~meV on MP-HAM-110K, corresponding to increases of
$61.0\%$ and $52.5\%$, respectively. These increases demonstrate
the importance of incorporating information about neighboring
couplings through equivariant matrix features when modeling
how the surrounding environment affects each Hamiltonian block.
Invariant message passing still conveys information about
neighboring atoms and determines the coefficients in
Eq.~(\ref{eq:ablation_context_local_update}), but the
equivariant context matrices are no longer available for
feature updates. The full model constructs these atom-wise
context features from neighboring atom-pair features and uses
them in matrix contractions to update the row and column
components of each atom-pair feature. The neighboring
environment therefore contributes to these updates through
both scalar coefficients and equivariant matrices, enabling
a more expressive description of its influence on orbital
or orbital--spin couplings.

Removing the MLPs that predict invariant coefficients and
attention logits from the local environment increases Gauge
MAE to $1.280$~meV on Materials-HAM-SOC and $2.230$~meV on
MP-HAM-110K, corresponding to increases of $52.2\%$ and
$39.8\%$, respectively. These increases demonstrate the
importance of adapting attention and equivariant matrix
contractions to the local environment in both formulations.
In this variant, the MLP outputs are replaced with learnable
parameters shared across all atom pairs within each layer.
The matrix contractions remain available, but their
coefficients no longer depend on the input, and attention
assigns equal weights to the incoming edges at each atom.
At each layer, the full model predicts these coefficients
and logits from invariant features that encode local
geometric and electronic information. These predictions
allow the network to adaptively weight neighboring
contributions and adjust how directional and contextual
information is used to update each equivariant atom-pair feature.

Removing LayerScale increases Gauge MAE to $0.963$~meV on
Materials-HAM-SOC and $2.065$~meV on MP-HAM-110K, corresponding
to increases of $14.5\%$ and $29.5\%$, respectively.
These increases highlight the benefit of using LayerScale
to control the magnitude of residual feature updates during
training. LayerScale provides separate trainable scaling
factors for equivariant atom-pair features and invariant
node and edge features, allowing their update magnitudes
to be adjusted independently at each layer. The contraction
normalizations, zero initialization of the correction
branches, and other stabilization mechanisms remain
unchanged in this ablation.

Together, the comparison with Only $H^{(0)}$ and the ablations
on both datasets demonstrate the accuracy gained from learned
corrections and the complementary roles of invariant and
equivariant features in Hamiltonian prediction. Invariant features allow the attention
weights and contraction coefficients to adapt to the local
geometric environment and electronic couplings, while
equivariant features retain the Hamiltonian structure in the
direct-product space and are refined across layers. Explicit
bond directions and equivariant context features provide
information about bond orientation and neighboring couplings,
respectively, while LayerScale controls the magnitude of
residual feature updates. Combining these components yields
the lowest prediction error on both datasets, demonstrating
the effectiveness of this design for both SOC and non-SOC
Hamiltonian prediction.

\clearpage

\begin{figure}[p]
	\centering
	\includegraphics[scale=0.4]{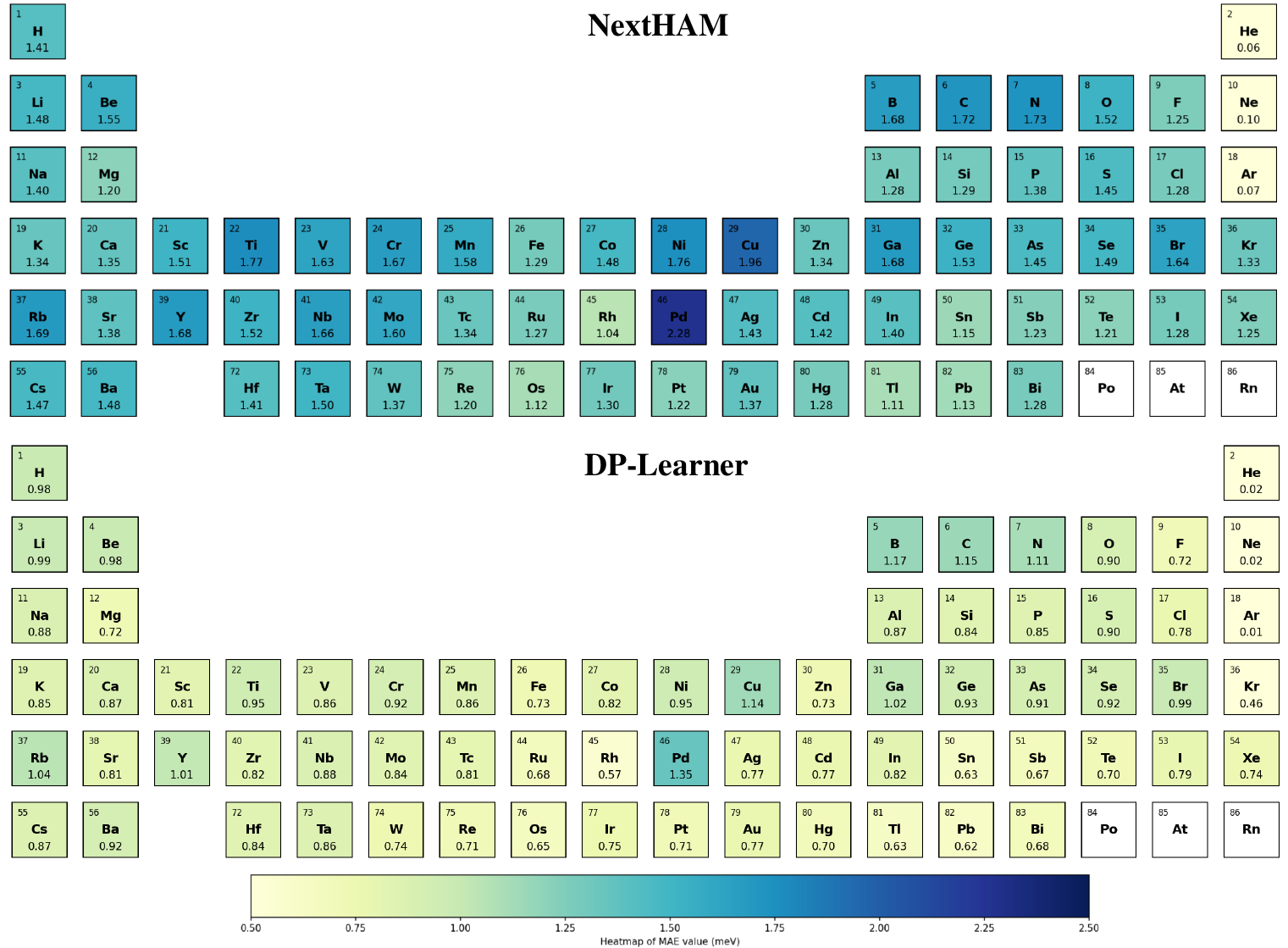}
	\caption{Gauge MAE (meV) for element-defined subsets of the Materials-HAM-SOC testing set. Each subset contains all test structures with the indicated element. Colors are clipped to the displayed range; numerical labels give the actual errors.}
	\label{fig:compare_table1}
\end{figure}

\begin{figure}[p]
	\centering
	\includegraphics[scale=0.7]{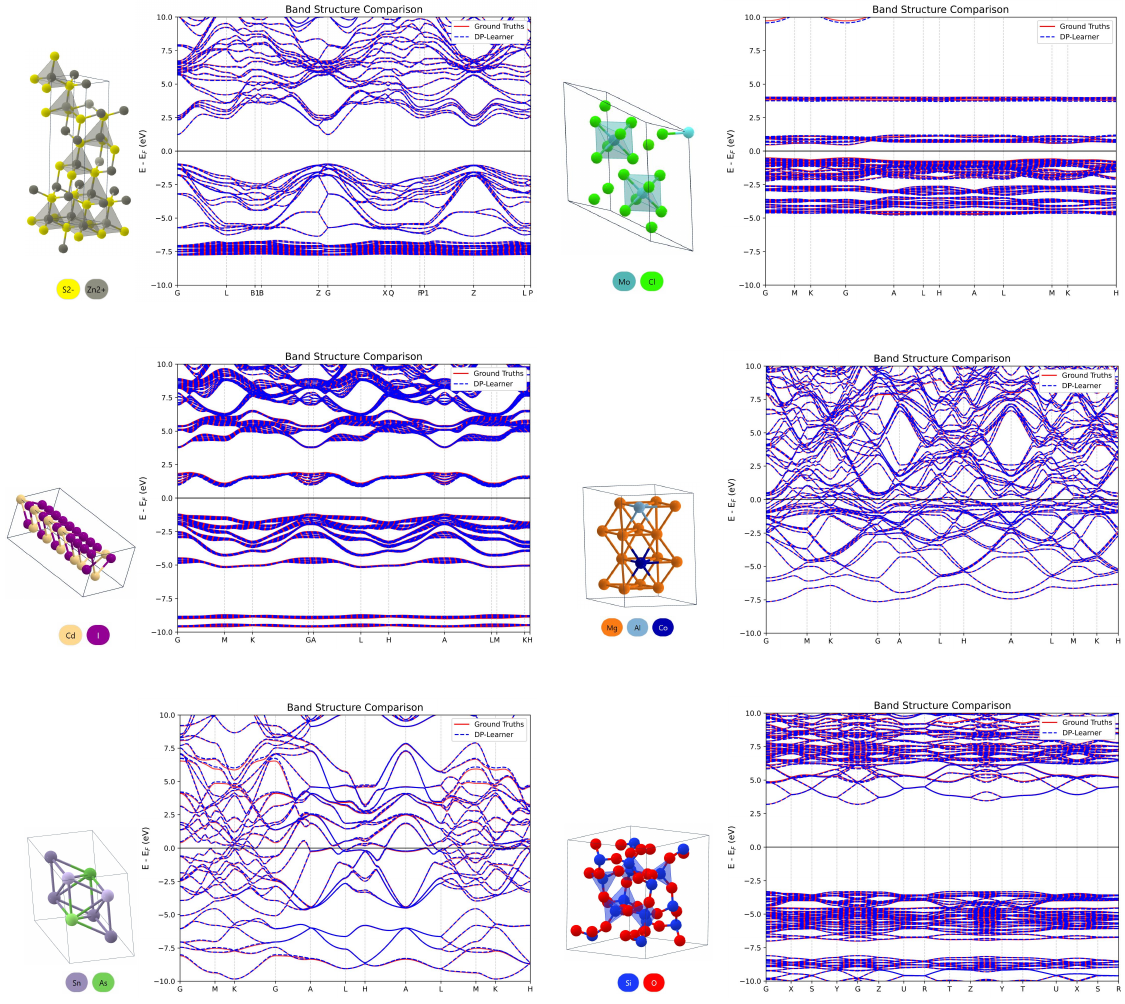}
	\caption{More band structure results for representative Materials-HAM-SOC testing samples. Red solid lines show the DFT reference bands used as ground truths in our dataset; blue dashed lines show DP-Learner predictions. Energies are referenced to the Fermi energy $E_F$.}
	\label{fig:band_res1}
\end{figure}

\clearpage

\begin{figure}[p]
	\centering
	\includegraphics[scale=0.4]{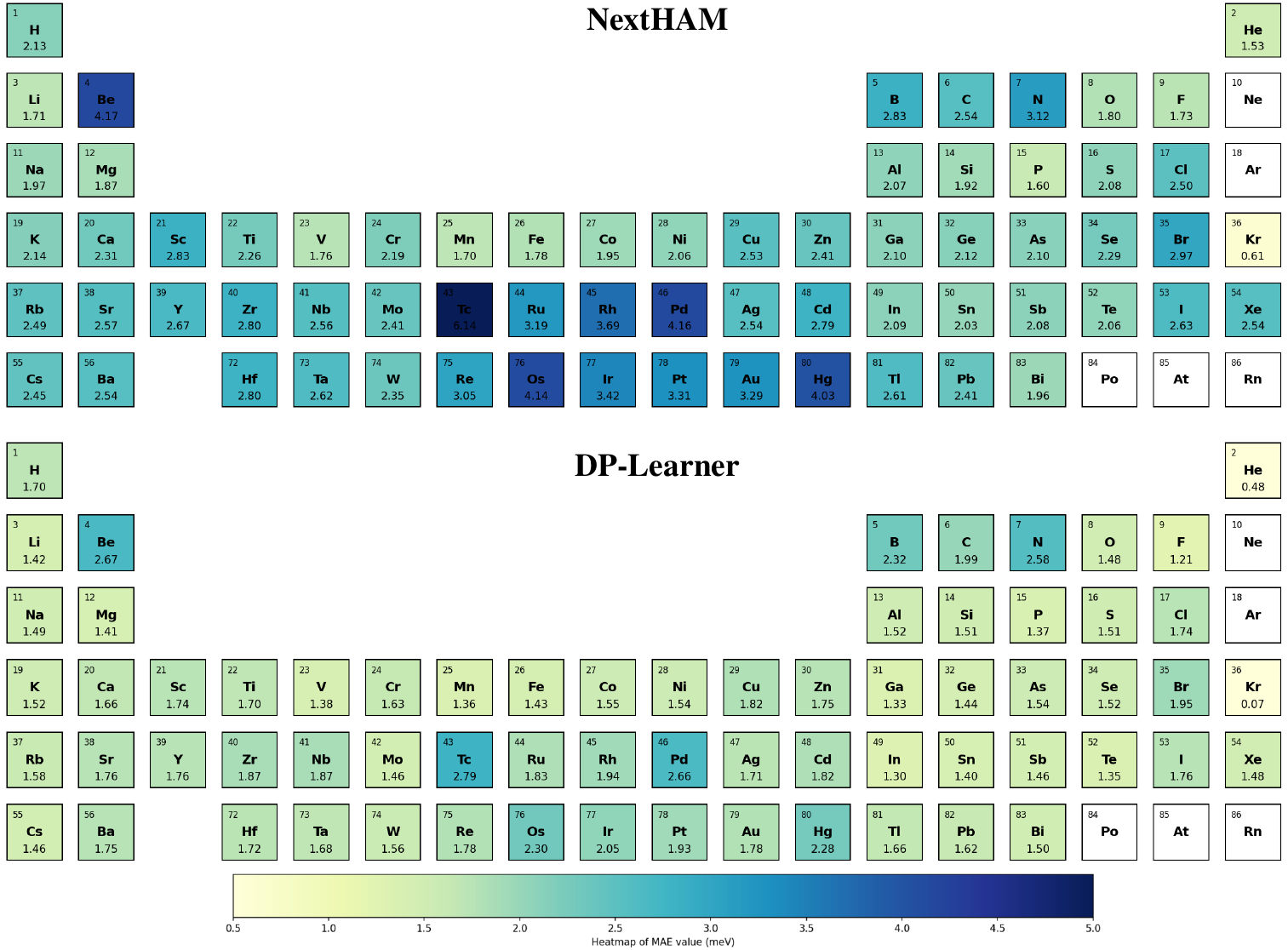}
	\caption{Gauge MAE (meV) values for element-defined subsets of the MP-HAM-110K testing set. Each subset contains all test structures with the indicated element. Colors are clipped to the displayed range; numerical labels give the actual errors.}
	\label{fig:compare_table2}
\end{figure}

\begin{figure}[p]
	\centering
	\includegraphics[scale=0.7]{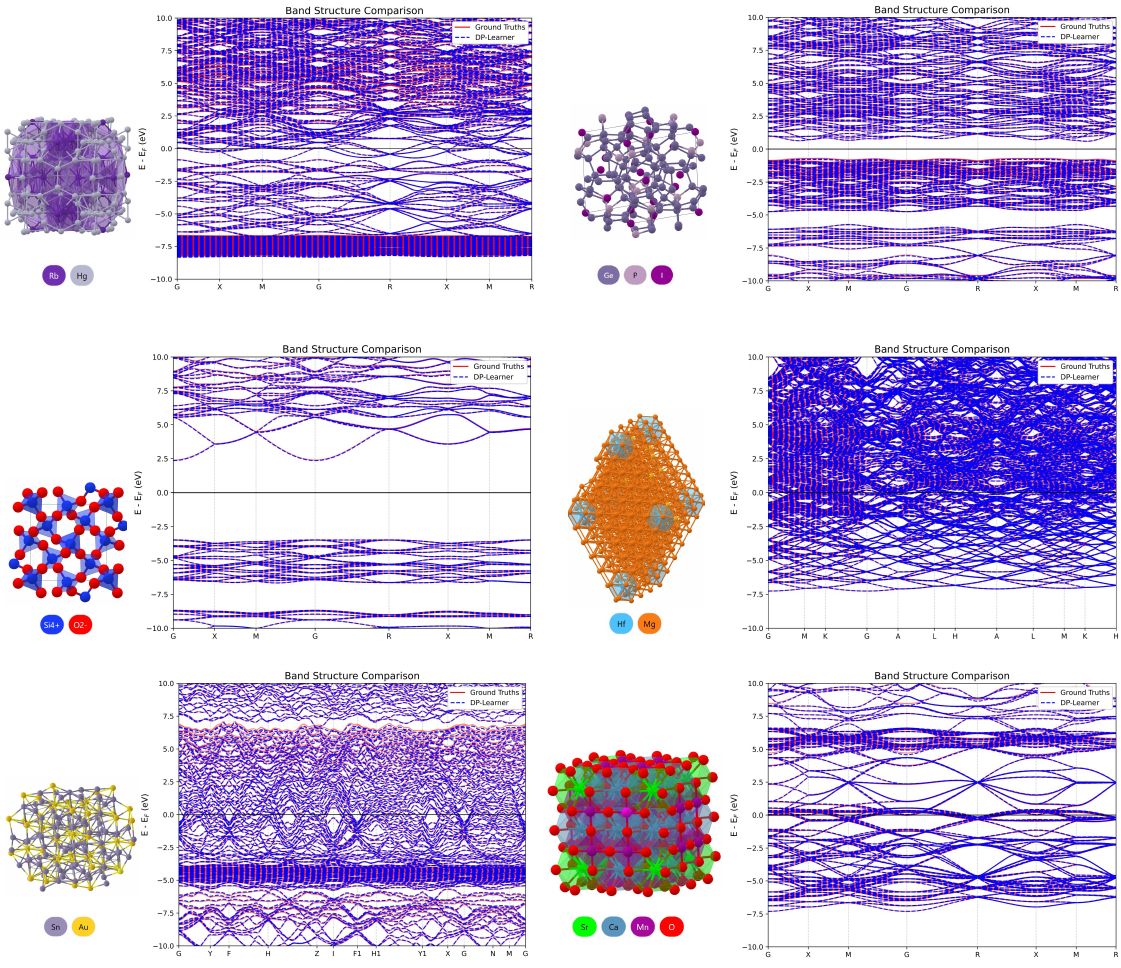}
	\caption{Additional band-structure results for representative test structures from MP-HAM-110K. Red solid lines show the DFT reference bands used as ground truths in our dataset; blue dashed lines show DP-Learner predictions. Energies are referenced to the Fermi energy $E_F$. }
	\label{fig:band_res2}
\end{figure}

\end{document}

%% file: math_commands.tex
\usepackage{amsmath,amsfonts,bm}

\def\eqref#1{equation~\ref{#1}}

\def\1{\bm{1}}

\DeclareMathAlphabet{\mathsfit}{\encodingdefault}{\sfdefault}{m}{sl}
\SetMathAlphabet{\mathsfit}{bold}{\encodingdefault}{\sfdefault}{bx}{n}



%% file: nsoc_pseudopotential_orbital_table_longtable_fullwidth.tex
\begingroup
\setlength{\LTleft}{0pt}
\setlength{\LTright}{0pt}
\setlength{\tabcolsep}{2pt}
\renewcommand{\arraystretch}{1.05}
\scriptsize
\begin{longtable}{@{}p{0.05\textwidth}p{0.09\textwidth}p{0.12\textwidth}p{0.40\textwidth}p{0.19\textwidth}p{0.10\textwidth}@{}}
\caption{Element-resolved scalar-relativistic ONCV-PBE pseudopotentials and numerical atomic-orbital (NAO) basis sets used for the non-SOC dataset.}\label{tab:nsoc_pseudo_orbital}\\
\toprule
$Z$ & Element & UPF & Pseudopotential valence & NAO basis & $r_{\mathrm{c}}$ (a.u.) \\ 
\midrule
\endfirsthead
\multicolumn{6}{c}{\tablename\ \thetable{} -- continued from previous page}\\
\toprule
$Z$ & Element & UPF & Pseudopotential valence & NAO basis & $r_{\mathrm{c}}$ (a.u.) \\ 
\midrule
\endhead
\midrule
\multicolumn{6}{r}{Continued on next page}\\
\endfoot
\bottomrule
\endlastfoot
1 & H & \texttt{H.upf} & $1s^{1}$ & $2s\,1p$ & 7 \\
2 & He & \texttt{He.upf} & $1s^{2}$ & $2s\,1p$ & 7 \\
3 & Li & \texttt{Li.upf} & $1s^{2}\,2s^{1}$ & $4s\,1p$ & 7 \\
4 & Be & \texttt{Be.upf} & $1s^{2}\,2s^{2}$ & $4s\,1p$ & 7 \\
5 & B & \texttt{B.upf} & $2s^{2}\,2p^{1}$ & $2s\,2p\,1d$ & 7 \\
6 & C & \texttt{C.upf} & $2s^{2}\,2p^{2}$ & $2s\,2p\,1d$ & 7 \\
7 & N & \texttt{N.upf} & $2s^{2}\,2p^{3}$ & $2s\,2p\,1d$ & 7 \\
8 & O & \texttt{O.upf} & $2s^{2}\,2p^{4}$ & $2s\,2p\,1d$ & 7 \\
9 & F & \texttt{F.upf} & $2s^{2}\,2p^{5}$ & $2s\,2p\,1d$ & 7 \\
10 & Ne & \texttt{Ne.upf} & $2s^{2}\,2p^{6}$ & $2s\,2p\,1d$ & 7 \\
11 & Na & \texttt{Na.upf} & $2s^{2}\,2p^{6}\,3s^{1}$ & $4s\,2p\,1d$ & 7 \\
12 & Mg & \texttt{Mg.upf} & $2s^{2}\,2p^{6}\,3s^{2}$ & $4s\,2p\,1d$ & 7 \\
13 & Al & \texttt{Al.upf} & $3s^{2}\,3p^{1}$ & $2s\,2p\,1d$ & 7 \\
14 & Si & \texttt{Si.upf} & $3s^{2}\,3p^{2}$ & $2s\,2p\,1d$ & 7 \\
15 & P & \texttt{P.upf} & $3s^{2}\,3p^{3}$ & $2s\,2p\,1d$ & 7 \\
16 & S & \texttt{S.upf} & $3s^{2}\,3p^{4}$ & $2s\,2p\,1d$ & 7 \\
17 & Cl & \texttt{Cl.upf} & $3s^{2}\,3p^{5}$ & $2s\,2p\,1d$ & 7 \\
18 & Ar & \texttt{Ar.upf} & $3s^{2}\,3p^{6}$ & $2s\,2p\,1d$ & 7 \\
19 & K & \texttt{K.upf} & $3s^{2}\,3p^{6}\,4s^{1}$ & $4s\,2p\,1d$ & 7 \\
20 & Ca & \texttt{Ca.upf} & $3s^{2}\,3p^{6}\,4s^{2}$ & $4s\,2p\,1d$ & 8 \\
21 & Sc & \texttt{Sc.upf} & $3s^{2}\,3p^{6}\,3d^{1}\,4s^{2}$ & $4s\,2p\,2d\,1f$ & 7 \\
22 & Ti & \texttt{Ti.upf} & $3s^{2}\,3p^{6}\,3d^{2}\,4s^{2}$ & $4s\,2p\,2d\,1f$ & 8 \\
23 & V & \texttt{V.upf} & $3s^{2}\,3p^{6}\,3d^{3}\,4s^{2}$ & $4s\,2p\,2d\,1f$ & 7 \\
24 & Cr & \texttt{Cr.upf} & $3s^{2}\,3p^{6}\,3d^{5}\,4s^{1}$ & $4s\,2p\,2d\,1f$ & 8 \\
25 & Mn & \texttt{Mn.upf} & $3s^{2}\,3p^{6}\,3d^{5}\,4s^{2}$ & $4s\,2p\,2d\,1f$ & 8 \\
26 & Fe & \texttt{Fe.upf} & $3s^{2}\,3p^{6}\,3d^{6}\,4s^{2}$ & $4s\,2p\,2d\,1f$ & 7 \\
27 & Co & \texttt{Co.upf} & $3s^{2}\,3p^{6}\,3d^{7}\,4s^{2}$ & $4s\,2p\,2d\,1f$ & 7 \\
28 & Ni & \texttt{Ni.upf} & $3s^{2}\,3p^{6}\,3d^{8}\,4s^{2}$ & $4s\,2p\,2d\,1f$ & 7 \\
29 & Cu & \texttt{Cu.upf} & $3s^{2}\,3p^{6}\,3d^{10}\,4s^{1}$ & $4s\,2p\,2d\,1f$ & 7 \\
30 & Zn & \texttt{Zn.upf} & $3s^{2}\,3p^{6}\,3d^{10}\,4s^{2}$ & $4s\,2p\,2d\,1f$ & 7 \\
31 & Ga & \texttt{Ga.upf} & $3d^{10}\,4s^{2}\,4p^{1}$ & $2s\,2p\,2d\,1f$ & 7 \\
32 & Ge & \texttt{Ge.upf} & $3d^{10}\,4s^{2}\,4p^{2}$ & $2s\,2p\,2d\,1f$ & 7 \\
33 & As & \texttt{As.upf} & $4s^{2}\,4p^{3}$ & $2s\,2p\,1d$ & 7 \\
34 & Se & \texttt{Se.upf} & $4s^{2}\,4p^{4}$ & $2s\,2p\,1d$ & 7 \\
35 & Br & \texttt{Br.upf} & $4s^{2}\,4p^{5}$ & $2s\,2p\,1d$ & 7 \\
36 & Kr & \texttt{Kr.upf} & $4s^{2}\,4p^{6}$ & $2s\,2p\,1d$ & 8 \\
37 & Rb & \texttt{Rb.upf} & $4s^{2}\,4p^{6}\,5s^{1}$ & $4s\,2p\,1d$ & 8 \\
38 & Sr & \texttt{Sr.upf} & $4s^{2}\,4p^{6}\,5s^{2}$ & $4s\,2p\,1d$ & 8 \\
39 & Y & \texttt{Y.upf} & $4s^{2}\,4p^{6}\,4d^{1}\,5s^{2}$ & $4s\,2p\,2d\,1f$ & 7 \\
40 & Zr & \texttt{Zr.upf} & $4s^{2}\,4p^{6}\,4d^{2}\,5s^{2}$ & $4s\,2p\,2d\,1f$ & 8 \\
41 & Nb & \texttt{Nb.upf} & $4s^{2}\,4p^{6}\,4d^{4}\,5s^{1}$ & $4s\,2p\,2d\,1f$ & 7 \\
42 & Mo & \texttt{Mo.upf} & $4s^{2}\,4p^{6}\,4d^{5}\,5s^{1}$ & $4s\,2p\,2d\,1f$ & 7 \\
43 & Tc & \texttt{Tc.upf} & $4s^{2}\,4p^{6}\,4d^{5}\,5s^{2}$ & $4s\,2p\,2d\,1f$ & 8 \\
44 & Ru & \texttt{Ru.upf} & $4s^{2}\,4p^{6}\,4d^{7}\,5s^{1}$ & $4s\,2p\,2d\,1f$ & 8 \\
45 & Rh & \texttt{Rh.upf} & $4s^{2}\,4p^{6}\,4d^{8}\,5s^{1}$ & $4s\,2p\,2d\,1f$ & 8 \\
46 & Pd & \texttt{Pd.upf} & $4s^{2}\,4p^{6}\,4d^{10}$ & $2s\,2p\,2d\,1f$ & 7 \\
47 & Ag & \texttt{Ag.upf} & $4s^{2}\,4p^{6}\,4d^{10}\,5s^{1}$ & $4s\,2p\,2d\,1f$ & 7 \\
48 & Cd & \texttt{Cd.upf} & $4s^{2}\,4p^{6}\,4d^{10}\,5s^{2}$ & $4s\,2p\,2d\,1f$ & 7 \\
49 & In & \texttt{In.upf} & $4d^{10}\,5s^{2}\,5p^{1}$ & $2s\,2p\,2d\,1f$ & 7 \\
50 & Sn & \texttt{Sn.upf} & $4d^{10}\,5s^{2}\,5p^{2}$ & $2s\,2p\,2d\,1f$ & 7 \\
51 & Sb & \texttt{Sb.upf} & $4d^{10}\,5s^{2}\,5p^{3}$ & $2s\,2p\,2d\,1f$ & 7 \\
52 & Te & \texttt{Te.upf} & $4d^{10}\,5s^{2}\,5p^{4}$ & $2s\,2p\,2d\,1f$ & 7 \\
53 & I & \texttt{I.upf} & $5s^{2}\,5p^{5}$ & $2s\,2p\,1d$ & 7 \\
54 & Xe & \texttt{Xe.upf} & $5s^{2}\,5p^{6}$ & $2s\,2p\,1d$ & 7 \\
55 & Cs & \texttt{Cs.upf} & $5s^{2}\,5p^{6}\,6s^{1}$ & $4s\,2p\,1d$ & 8 \\
56 & Ba & \texttt{Ba.upf} & $5s^{2}\,5p^{6}\,6s^{2}$ & $4s\,2p\,1d$ & 8 \\
72 & Hf & \texttt{Hf.upf} & $5s^{2}\,5p^{6}\,5d^{2}\,6s^{2}$ & $4s\,2p\,2d\,1f$ & 7 \\
73 & Ta & \texttt{Ta.upf} & $5s^{2}\,5p^{6}\,5d^{3}\,6s^{2}$ & $4s\,2p\,2d\,1f$ & 7 \\
74 & W & \texttt{W.upf} & $5s^{2}\,5p^{6}\,5d^{4}\,6s^{2}$ & $4s\,2p\,2d\,1f$ & 8 \\
75 & Re & \texttt{Re.upf} & $5s^{2}\,5p^{6}\,5d^{5}\,6s^{2}$ & $4s\,2p\,2d\,1f$ & 7 \\
76 & Os & \texttt{Os.upf} & $5s^{2}\,5p^{6}\,5d^{6}\,6s^{2}$ & $4s\,2p\,2d\,1f$ & 8 \\
77 & Ir & \texttt{Ir.upf} & $5s^{2}\,5p^{6}\,5d^{7}\,6s^{2}$ & $4s\,2p\,2d\,1f$ & 8 \\
78 & Pt & \texttt{Pt.upf} & $5s^{2}\,5p^{6}\,5d^{9}\,6s^{1}$ & $4s\,2p\,2d\,1f$ & 7 \\
79 & Au & \texttt{Au.upf} & $5s^{2}\,5p^{6}\,5d^{10}\,6s^{1}$ & $4s\,2p\,2d\,1f$ & 7 \\
80 & Hg & \texttt{Hg.upf} & $5s^{2}\,5p^{6}\,5d^{10}\,6s^{2}$ & $4s\,2p\,2d\,1f$ & 7 \\
81 & Tl & \texttt{Tl.upf} & $5d^{10}\,6s^{2}\,6p^{1}$ & $2s\,2p\,2d\,1f$ & 7 \\
82 & Pb & \texttt{Pb.upf} & $5d^{10}\,6s^{2}\,6p^{2}$ & $2s\,2p\,2d\,1f$ & 7 \\
83 & Bi & \texttt{Bi.upf} & $5d^{10}\,6s^{2}\,6p^{3}$ & $2s\,2p\,2d\,1f$ & 7 \\
\end{longtable}
\endgroup